\pdfoutput=1

\documentclass[11pt]{article}

\usepackage[]{EMNLP2024}

\usepackage{times}
\usepackage{latexsym}

\usepackage{amsmath}
\usepackage{multirow}
\usepackage{bbm}
\usepackage{graphicx}
\usepackage{amssymb}
\usepackage{booktabs}
\usepackage{lscape}
\usepackage{pifont}
\usepackage{algorithm}
\usepackage{graphicx}
\usepackage{caption}
\usepackage{subcaption}

\usepackage{fdsymbol}

\DeclareMathOperator*{\argmax}{arg\,max}

\newcommand\benchmarkname{\textcolor{black}{\textsc{MINERS Benchmark}}}
\newcommand\benchmarknameonly{\textcolor{black}{\textsc{MINERS}}}

\usepackage[T1]{fontenc}

\usepackage[utf8]{inputenc}

\usepackage{microtype}

\usepackage{color,soul}
\usepackage{inconsolata}

\title{MINERS$\vcenter{\hbox{\includegraphics[scale=0.15]{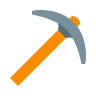}}} $: Multilingual Language Models as Semantic Retrievers}

\author{Genta Indra Winata$^1$\thanks{$\text{ }$ The work was conducted outside Capital One.}$\hspace{1.7mm}$, Ruochen Zhang$^2$, David Ifeoluwa Adelani$^3$ \\
  $^1$Capital One $\quad$ $^2$Brown University \\ $^3$University College London\\
  \texttt{genta.winata@capitalone.com, ruochen\_zhang@brown.edu, d.adelani@ucl.ac.uk}}

\begin{document}
\maketitle
\begin{abstract}
Words have been represented in a high-dimensional vector space that encodes their semantic similarities, enabling downstream applications such as retrieving synonyms, antonyms, and relevant contexts. However, despite recent advances in multilingual language models (LMs), the effectiveness of these models' representations in semantic retrieval contexts has not been comprehensively explored.
To fill this gap, this paper introduces the \benchmarknameonly{}, a benchmark designed to evaluate the ability of multilingual LMs in semantic retrieval tasks, including bitext mining and classification via retrieval-augmented contexts. We create a comprehensive framework to assess the robustness of LMs in retrieving samples across over 200 diverse languages, including extremely low-resource languages in challenging cross-lingual and code-switching settings. Our results demonstrate that by solely retrieving semantically similar embeddings yields performance competitive with state-of-the-art approaches, without requiring any fine-tuning. 
\end{abstract}

\section{Introduction}

Language models (LMs) play a crucial role in learning natural language representations~\cite{cer2018universal,kenton2019bert,reimers2019sentence,gao2021simcse,feng2022language} and have been successfully applied to various natural language processing (NLP) tasks, such as document retrieval~\cite{yang2019simple,wang2023colbert}. Existing benchmarks have systematically evaluated LMs to provide empirical assessments of their performance across a range of embedding tasks. Some notable benchmarks include Big-Bench~\cite{srivastava2023beyond}, MTEB~\cite{muennighoff2023mteb}, SemEval~\cite{cer2017semeval}, and BEIR Benchmark~\cite{thakur2021beir}. MTEB, in particular, has been established as a comprehensive benchmark for evaluating the effectiveness of embeddings in downstream NLP applications. However, their analysis of the multilingual space has been limited to bitext mining, without further exploration of how these embeddings can be utilized in other multilingual downstream tasks.

The advancement of multilingual LMs is remarkable, demonstrating impressive capabilities in adapting to new languages through fine-tuning~\cite{conneau2019cross, alabi2022adapting}, learning from few-shot samples via in-context learning (ICL)~\cite{lin2021few, winata2021language, tanwar2023multilingual, cahyawijaya2024llms,biderman2024lessons}, enabling cross-lingual zero-shot transfer~\cite{ruder2021xtreme}, and incorporating language-specific adapters~\cite{ansell2021mad,yong2023bloom+}. This exploration now includes low-resource and regional languages not part of the pretraining phase, promoting NLP research for underrepresented languages~\cite{adelani2022masakhaner, winata2022cross, song2023globalbench}. However, multilingual LMs face two key challenges: (1) the lack of a comprehensive benchmark for evaluating effectiveness in semantic retrieval, and (2) limited understanding of code-switching (CS) texts common in multilingual communities. 

Current CS evaluations focus on model fine-tuning benchmarks~\cite{aguilar2020lince, khanuja2020gluecos, winata2021multilingual, zhang2023multilingual}, without deeply exploring their potential as multilingual retrievers. Recent studies by \citet{winata2023efficient} have primarily focused on semantic similarity using encoder LMs in zero-shot cross-lingual settings but have not explored their application in generative LMs. This gap presents an opportunity to leverage these models as context providers for multilingual generative LMs~\cite{lewis2020retrieval, bevilacqua2022autoregressive}.

In this paper, we introduce \benchmarknameonly{}, the first benchmark designed to assess the multilingual LMs' ability in semantic retrieval across various tasks. \benchmarknameonly{} evaluates the representation of dense vectors in multiple tasks, including bitext retrieval, retrieval-based classification, and ICL classification. We have developed \benchmarknameonly{} to be a reproducible and reliable benchmark that utilizes high-dimensional multilingual vector representations. Notably, these tasks do not require any fine-tuning. 
The paper's contribution can be summarized as follows:
\begin{itemize}
    \item We introduce~\benchmarknameonly{}, the first comprehensive benchmark designed to systematically evaluate multilingual LMs as semantic retrievers across a vast array of languages. Covering 200+ languages, 11 encoder LMs, and 11 generative LMs, including open-source and commercial APIs. \benchmarknameonly{} offers a robust evaluation framework for assessing the effectiveness of LMs in diverse linguistic contexts.
    \item We show \benchmarknameonly{} is highly adaptable and scalable across various models. By consolidating scores from multiple models, \benchmarknameonly{} facilitates a comprehensive evaluation of task performance, providing insights into different approaches' strengths and weaknesses.
    \item We provide a thorough analysis across different evaluation difficulty levels, including monolingual, cross-lingual, and CS scenarios. We examine performance variations across different numbers of retrieved samples to offer insights into the impact of sample quantity on retrieval effectiveness.
    \item We compare the time efficiency of retrieval methods with conventional fine-tuning approaches. By demonstrating that retrieval methods require no training and offer a comparable performance of leveraging pre-trained models for semantic retrieval tasks.
\end{itemize}

\section{\benchmarkname{}}

\subsection{Motivation}
The \benchmarkname{}\footnote{We release the code to reproduce the benchmark results at~\url{https://github.com/gentaiscool/miners}} is introduced as a significant step forward in assessing the capabilities of multilingual LMs in producing high-dimensional representations for semantic retrieval. This benchmark is constructed with three fundamental aspects: \textbf{(1) Language Diversity}: The benchmark offers insights into the performance of LMs across a wide array of languages. It assesses not only the models' effectiveness in high-resource languages but also their capabilities in low-resource languages from various language families. Additionally, the benchmark includes evaluations of unseen languages to gauge the robustness of the models in predicting languages not encountered during pre-training. CS datasets are also incorporated to simulate realistic scenarios where bilingual or multilingual speakers mix languages, providing a more comprehensive assessment of the models' capabilities. \textbf{(2) Usefulness}: The benchmark includes evaluations across three distinct tasks to systematically measure the performance of multilingual LMs. First, it assesses the models' ability to retrieve semantically similar parallel data in bitext retrieval tasks. Second, it uses the retrieved samples for classification, evaluating the models' accuracy in categorizing text. Third, it employs the retrieved samples as context for generating labels in downstream classification tasks, highlighting the models' capability to incorporate retrieved information into context-aware classification. Additionally, the benchmark demonstrates the potential of using multiple LMs and APIs together to represent text as an ensemble, further emphasizing their utility. \textbf{(3) Efficiency}: The benchmark is crafted with efficiency as a key principle. It is designed to be straightforward and easily extendable, accommodating new datasets to ensure its longevity and continued relevance. Additionally, the benchmark is publicly available, promoting result reproducibility and encouraging collaboration and further research within the field. Importantly, the benchmark does not necessitate any model fine-tuning, as all evaluations are conducted exclusively through model inference, thereby streamlining the assessment process.

\begin{figure*}[!th]
    \centering
    \includegraphics[width=\linewidth]{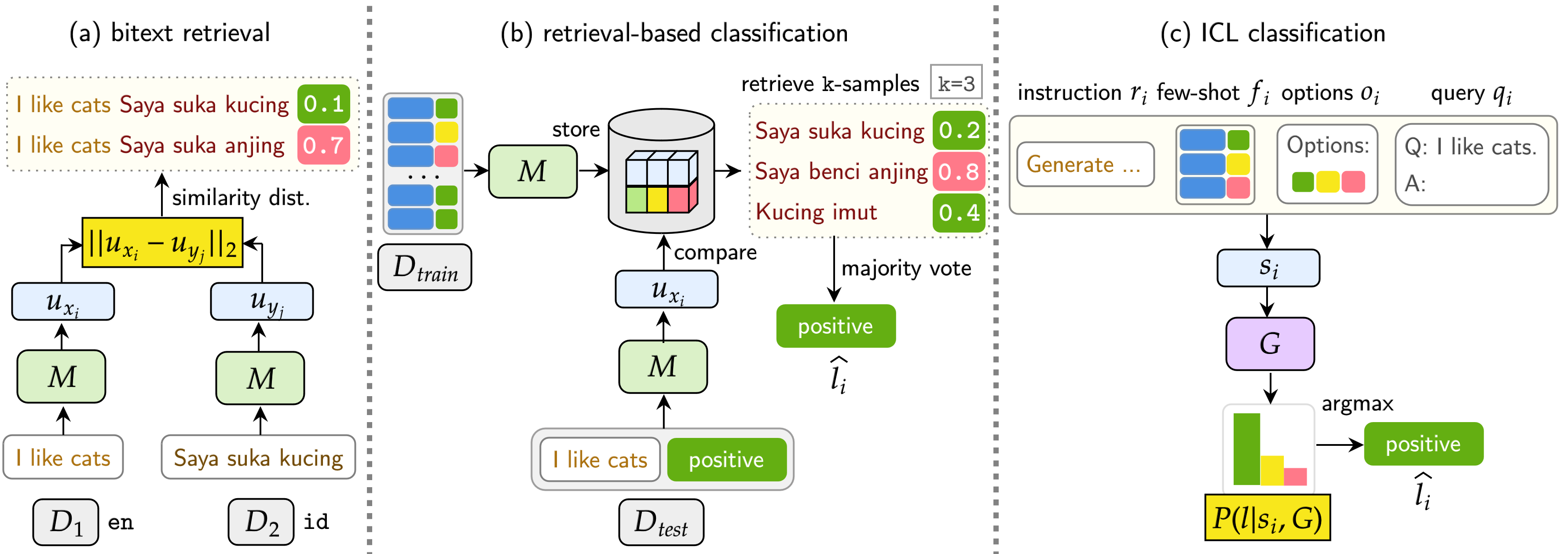} 
    \caption{\benchmarkname{} tasks. In this example, we compare English \texttt{(en)} and Indonesian \texttt{(id)} texts across three tasks: \textbf{(a)} bitext retrieval, \textbf{(b)} retrieval-based classification, and \textbf{(c)} ICL classification. Light blue cubes represent vector representations of samples from the training dataset $\mathcal{D}_{train}$, generated by $\mathcal{M}$, while green, yellow, and red cubes denote raw text labels. The few-shot samples $f_i$ in task (c) are retrieved in the same manner as in task (b). The English translations of the text in the figure are as follows: "Saya suka kucing" ("I like cats"), "Saya suka anjing" ("I like dogs"), "Saya benci anjing" ("I hate dogs"), and "Kucing imut" ("Cute cats").}
    \label{fig:pipeline}
\end{figure*}

\subsection{Tasks}
Our benchmark evaluates LMs on three tasks: bitext retrieval, retrieval-based classification, and ICL classification.  Figure~\ref{fig:pipeline} provides an overview of tasks.
We describe the task details as follows:

\paragraph{Bitext Retrieval}
This task aims to measure the LM's ability to retrieve semantically similar samples from parallel datasets. The task is also useful to understand how the model perform when there are language distribution shifts, especially when some words are code-switched. Formally, given a parallel dataset $\mathcal{D}$ with two language $L_1$ and $L_2$, we can have two different datasets $\mathcal{D}_{L_1}$ and $\mathcal{D}_{L_2}$. For each sample $x_i$ in $\mathcal{D}_{L_1}$, the closest sample $\hat{y}$ is searched through $\mathcal{D}_{L_2}$, by finding the lowest distance score between two samples $x_i$ and $y_j$. The score $s_{i,j}$ is computed by measuring the Euclidean distance of their high-dimensional vector representation which generated by using an LM $\mathcal{M}$. In this case, euclidean distance is used to compute the score $s_{i,j}=||\textbf{u}_{x_i} - \textbf{u}_{y_j}||_2$, where $\textbf{u}_{x_i}$ and $\textbf{u}_{y_j}$ are vector representation of samples $x_i$ and $y_j$, respectively. We can also use other distance measures, but the difference is minimal.

\paragraph{Retrieval-based Classification}
This task involves using the retrieved samples' labels from the training set to predict labels in downstream NLP classification tasks. The goal is to assess the usefulness of our retrieved samples and introduce an efficient prediction method by directly searching for similar samples in the training set. Given the retrieved $k$ pairs of training samples with labels $[(y_1, l_1), \cdots, (y_k, l_k)]$, a label $\hat{l}$ is selected by majority voting and assigned to the corresponding test sample. Increasing $k$ can enhance performance.

\begin{table*}[!ht]
\centering
\resizebox{\textwidth}{!}{
    \begin{tabular}{lccc}
    \toprule
    \textbf{Dataset} & \textbf{|Lang.|} & \textbf{Task} & \textbf{Eval Metric} \\ \midrule
    BUCC~\cite{zweigenbaum2017overview,zweigenbaum2018overview} & 5 & Bitext Retrieval & F1 \\
    MASSIVE~\cite{fitzgerald2023massive} & 51 & Intent Classification$^\diamondsuit$ & Acc. \\
    NollySenti~\cite{shode2023nollysenti} & 5 & Bitext Retrieval & F1 \\
    & & Sentiment Analysis$^\diamondsuit$$^\spadesuit$ & Acc. \\
    NusaX~\cite{winata2023nusax} & 12 & Bitext Retrieval &  F1 \\
    & & Sentiment Analysis$^\diamondsuit$$^\spadesuit$ & F1 \\
    NusaT~\cite{cahyawijaya2023nusawrites} & 12 & Bitext Retrieval &  F1 \\
    SIB-200~\cite{adelani2023sib} & 205 & Topic Classification$^\diamondsuit$$^\spadesuit$ & Acc. \\
    Tatoeba~\cite{tiedemann2020tatoeba} & 113 &  Bitext Retrieval & F1 \\ \midrule
    Code-switching \\ \midrule
    FIRE 2020~\cite{chakravarthi2020sentiment,hegde2022corpus} & 3 & Sentiment Analysis$^\diamondsuit$$^\spadesuit$ & Acc. \\ 
    LinCE MT~\cite{aguilar2020lince} & 2 & Bitext Retrieval & F1 \\
    LinCE SA~\cite{patwa2020semeval} & 2 & Sentiment Analysis$^\diamondsuit$$^\spadesuit$ & Acc. \\
    PHINC~\cite{srivastava2020phinc} & 2 & Bitext Retrieval & F1
    \\ \bottomrule
    \end{tabular}
}
\caption{Dataset list of \benchmarkname{}. The symbols indicate the tasks run on datasets.  $^\diamondsuit$Retrieval-based classification task. $^\spadesuit$ICL classification task.}
\label{dataset-list}
\vspace{-2mm}
\end{table*}

\paragraph{ICL Classification}
We aim to further utilize the retrieved training samples for natural generation tasks by using them as few-shot context, combined with task-specific instructions and a query. Formally, given a generative LLM $G$, we input a text sequence $s_i=(r_i;f_i;o_i;q_i)$, which includes a text instruction $r_i$, few-shot samples $f_i=[(y_1, l_1), \cdots, (y_k, l_k)]$, a list of label options $o_i$, and a query $q_i$, to generate an output text sequence. To generate the prediction, we use one of two methods based on the model's capabilities: (a) computing label probabilities, which offers precise predictions by reducing issues like typos, and (b) directly predicting labels through instructions, which is more efficient as responses match desired labels, eliminating the need to evaluate all options. We use method (a) when we can calculate the log-likelihood of the next token prediction; otherwise, we resort to method (b). For method (a), we compute the probability of each output class, normalize it by the token length, and select the label with the highest probability from the distribution as follows:
\begin{align}
    \hat{l}_i = \argmax_{l \in L} P(l|s_i,G),
\end{align}
where $L$ denotes the number of possible classes. For more details on model inference, please refer to Appendix~\ref{lm-inference}.

\subsection{Settings}
We gauge LMs' robustness to various text inputs with three different evaluation settings:
\begin{itemize}
    \item \textbf{Monolingual (Mono)}: We measure the individual language performance using the same language as train and test sets.
    \item \textbf{Code-switching (CS)}: We measure the performance of mixed language datasets. For bitext retrieval, we find a corresponding CS text translation from a monolingual text, or vice versa, and for retrieval-based classification and ICL classification, we take CS texts as input and predict their labels.
    \item \textbf{Cross-lingual (XL)}: We measure the performance of multilingual datasets with one language as the source language and the rest as target languages. For detailed information, please refer to Table~\ref{cross-lingual-settings} in the Appendix.
    \item \textbf{Cross-lingual Code-switching (XL CS)}: We tackle a more challenging scenario by evaluating CS data within a cross-lingual context. 
\end{itemize}

\begin{table*}[!th]
\centering
\resizebox{.93\textwidth}{!}{
    \begin{tabular}{lccc|ccccc}
    \toprule
    \textbf{Model} & \multicolumn{3}{c|}{\textbf{Bitext Retrieval}} & \multicolumn{5}{c}{\textbf{Retrieval-based Classification}} \\ 
    & XL & CS & avg. & Mono & XL & CS & XL CS & avg. \\ \midrule 
    Fine-tune (XLM-R$_\text{BASE}$) & N/A & N/A & N/A & \textbf{79.55} & 65.92 & 62.28 & 34.64 & 60.60 \\ \midrule
    LaBSE & 83.90 &  52.03 &67.97 & 73.46 & \textbf{72.73} & 60.64  & 41.10 & 61.98\\
    CMLM & 70.77 &  42.62 & 56.70&73.05 & 70.31 & 59.27 & 40.88 & 60.88\\
    E5$_\text{BASE}$ & 72.26 & 43.29 &57.78 &  75.08 & 65.51 & 61.16 & \textbf{42.73} & 61.12\\
    E5$_\text{LARGE}$ & 76.35 & 49.97 & 63.16& 77.52 & 71.08 & 61.91 & 41.99 & 63.13\\
    MPNet$_\text{BASE}$v2 &52.25 & 25.87 & 39.06&  66.17 & 59.69 & 58.33  & 41.25 & 56.36\\ 
    MiniLM$_\text{L12-E384}$ & 24.82 &  9.90 & 17.36& 63.18 & 51.16 & 57.28  & 39.61 & 52.81\\ 
    Glot-500 & 14.68 &  16.64 & 15.66& 65.66 & 51.75 & 58.11  & 40.06 & 53.90\\ 
    XLM-R$_\text{BASE}$ & 17.79 & 10.61 & 14.20& 63.62 & 47.59 & 58.25 & 41.02 & 52.62\\
    XLM-R$_\text{LARGE}$ & 12.45 & 6.04 & 9.25& 61.76 & 43.88 & 57.30 & 39.47 & 50.60\\ \midrule
    Cohere-Embedv3 & 76.39 &  53.25 & 64.82&  78.56 &  \underline{72.67} & 62.12 & \underline{42.36} & \textbf{63.93}\\ 
    OpenAI-Embedv3 & 69.02 & \textbf{68.73} & 68.88& 73.97  & 67.13 & \textbf{62.77} & 40.50 & 61.09\\ \midrule
    DistFuse (2)$^\dagger$ & \textbf{84.72} & 56.47 & \textbf{70.60} & 78.34 & 70.87 & 62.13  & 40.73 & 63.02 \\
    DistFuse (3)$^\dagger$ & \underline{83.28} & \underline{56.83} & \underline{70.06} & \underline{78.80} & 70.19 & \underline{62.31}  & 41.77 & \underline{63.27} \\ \bottomrule
    \end{tabular}
}
\caption{Results for bitext retrieval task ($k=1$) and retrieval-based classification ($k=10$). \textbf{Mono}, \textbf{XL} and \textbf{CS} denote monolingual, cross-lingual and code-switching, respectively. \textbf{Bold} and \underline{underlined} numbers present the best and second-best models. $^\dagger$For DistFuse (2), we use $\alpha=1, \beta=3$ and for DistFuse (3), we use $\alpha=1, \beta=2, \gamma=3$. The reported weights represent the best-performing configurations identified during our tuning process.} 
\label{results-bitext-retrieval-class-overall}
\end{table*}

\subsection{Datasets}
Table~\ref{dataset-list} presents 11 datasets: 7 multilingual and 4 CS datasets, covering both parallel and classification types. Parallel datasets are ideal for bitext retrieval due to their aligned multilingual content, enabling bitext mining and machine translation tasks. Classification datasets include intent classification, sentiment analysis, and topic classification, which we evaluate for retrieval-based and ICL classification tasks.
For ICL, we construct prompts using a unified English template across all generative language models to ensure simplicity and consistency. Detailed instructions for each task are provided in Tables~\ref{icl-prompt-templates-labse} and~\ref{icl-prompt-templates-e5} in the Appendix.

\subsection{Models}
\paragraph{Encoder LMs and APIs} We use 9 open-source LMs: LaBSE~\cite{feng2022language}, CMLM~\cite{cer2018universal}, multilingual E5$_\text{BASE}$, multilingual E5$_\text{LARGE}$~\cite{wang2024multilingual}, multilingual MPNet$_\text{BASE}$v2~\cite{song2020mpnet}, multilingual MiniLM$_\text{L12-E384}$~\cite{wang2020minilm}, Glot-500~\cite{imanigooghari2023glot500}, XLM-R$_\text{BASE}$, XLM-R$_\text{LARGE}$~\cite{conneau2019cross}, and two commercial embedding APIs: Cohere-Embedv3 \texttt{(embed-multilingual-v3.0)} and OpenAI-Embedv3 \texttt{(text-embedding-3-large)}.\footnote{The APIs were accessed on May 2024.}

\paragraph{Generative LMs} We opt for 8 different open-source LMs: (1) BLOOMZ~\cite{muennighoff2023crosslingual}, an instruction tuned BLOOM~\cite{le2023bloom} with three different sizes (\texttt{560m}, \texttt{1B}, \texttt{3B}) to further analyze the performance trend when increasing the model size, (2) mT0 3B \texttt{(xl)}~\cite{muennighoff2023crosslingual}, an instruction tuned mT5~\cite{xue2021mt5}, (3) XGLM~\cite{lin2021few} with two different sizes \texttt{(564m and 2.9B)}, (4) Aya-23 8B~\cite{aryabumi2024aya}, (5) Aya-101 13B~\cite{ustun2024aya}, (6) Gemma 1.1 Instruct~\cite{team2024gemma}, (7) Llama 3 8B Instruct, and (8) Llama 3.1 8B Instruct~\cite{dubey2024llama}, and three commercial APIs:  (1) Command-R, (2) GPT-3.5 Turbo \texttt{(gpt-3.5-turbo-0125)} and (3) GPT-4o \texttt{(gpt-4o-2024-05-13)}. All open-source models can be found on Hugging Face. Please check the Appendix on Table~\ref{hf-models} for details.

\paragraph{Ensemble Models}
To enhance scalability and effectiveness, we can use multiple models with DistFuse~\cite{winata2023efficient} to improve retrieval results. DistFuse combines models by calculating distance scores of label distributions and merging them through a linear combination. We report two DistFuse settings for bitext retrieval and retrieval-based classification tasks:
\begin{itemize}
    \item \textbf{DistFuse (2)} utilizes two models: \text{LaBSE} and \text{E5}$_\text{LARGE}$;
    \item \textbf{DistFuse (3)} utilizes three models: \text{LaBSE}, \text{E5}$_\text{LARGE}$, and \text{Cohere-Embedv3}.
\end{itemize}
To maintain conciseness, we denote the weights assigned to distances computed by \text{LaBSE}, \text{E5}$_\text{LARGE}$, and \text{Cohere-Embedv3} as $\alpha$, $\beta, \gamma$, respectively.   

\begin{table*}[!t]
\centering
\resizebox{\textwidth}{!}{
    \begin{tabular}{lccccc|ccccc}
    \toprule
    \textbf{Model} & \multicolumn{5}{c|}{\textbf{Zero-shot ICL}} & \multicolumn{5}{c}{\textbf{One-shot ICL}}\\
    & \multicolumn{1}{c}{Mono} & \multicolumn{1}{c}{XL} & \multicolumn{1}{c}{CS}& \multicolumn{1}{c}{XL CS} & avg. & \multicolumn{1}{c}{Mono} & \multicolumn{1}{c}{XL} & \multicolumn{1}{c}{CS}& \multicolumn{1}{c}{XL CS} & avg. \\ \midrule
    BLOOMZ 560M & 45.88 & 43.36 & 35.83 & 12.09 & 34.29 & 72.37 &	71.98 & 	54.25	& 36.35	& 58.74\\ 
    BLOOMZ 1.7B & 54.10 &	52.86 & 	35.70	& 11.80 & 38.62 & 71.38 & 70.65 & 	\textbf{58.04} & 	38.50	& 59.64\\
    BLOOMZ 3B & 53.20 &	51.78 & 36.32 & 9.50 & 	37.70 &  74.08 & 73.19 & \underline{57.44} & 	39.09	& 60.95\\ 
    mT0 3B & 53.29& 53.64 &  40.11   & 42.51 & 47.39 &59.02 & 57.86 &  46.66 &  42.36 & 51.48\\ 
    XGLM 564m & 39.25 &37.19    & 29.92 & 10.46& 29.21 & 37.26    & 40.12   & 22.64& 12.83 & 28.21 \\ 
    XGLM 2.9B & 42.41   &40.16    & 34.71 & 10.39& 31.92 &42.57 & 48.76  &  27.45 &  10.39 & 32.29 \\
    Aya-23 8B & 39.88 & 	36.88 & 	53.72 & 	43.18	& 43.42 & 63.66 & 63.53 & 53.12	& 38.50 & 54.70 \\
    Aya-101 13B & \underline{78.65} & \underline{77.72} & 42.29 & 26.26 & 56.23 & \underline{81.00} & \underline{80.20} & 50.90  & 36.20 & \underline{62.08} \\
    Gemma 1.1 7B Instruct & 55.51 & 53.36    & 51.62 & 37.24 & 49.43 & 65.82  &   64.49   & 53.12 & 35.68 & 54.78 \\
    Llama 3 8B Instruct &   62.40   & 60.41  & 52.72& 36.05& 52.90 & 74.85 &69.61   & 54.12 & 35.68 & 58.57 \\ 
    Llama 3.1 8B Instruct & 60.59 & 58.86  & 47.99 & 26.56 & 48.50 & 72.68 & 59.00 & 54.11 & 35.16 & 55.24 \\ \midrule
    Command-R & 47.98 & 46.02 & \textbf{54.84} & 44.44 & 48.32 & 58.36 & 56.89   &  56.84 & 41.99 & 53.52 \\
    GPT-3.5 Turbo & 67.10 & 65.13& \underline{54.32} & \underline{45.18} & \underline{57.93} & 71.01 & 71.56 & 	57.13 & \underline{42.73} & 60.61 \\ 
    GPT-4o & \textbf{79.92} & \textbf{79.15} & 	53.48	& \textbf{53.04}	 & \textbf{66.40} & \textbf{82.24} & \textbf{80.95} & 57.14 & 	\textbf{49.26}	& \textbf{67.40} \\ \bottomrule
    \end{tabular}
}
\caption{Results on ICL classification with E5$_\text{LARGE}$ retriever. \textbf{Bold} and \underline{underlined} numbers present the best and second-best models.} 
\label{results-icl}
\end{table*}

\begin{figure*}[!th]
    \centering
    \begin{subfigure}[t]{0.19\textwidth}
        \centering
        \caption{}
        \includegraphics[width=1.018\linewidth]{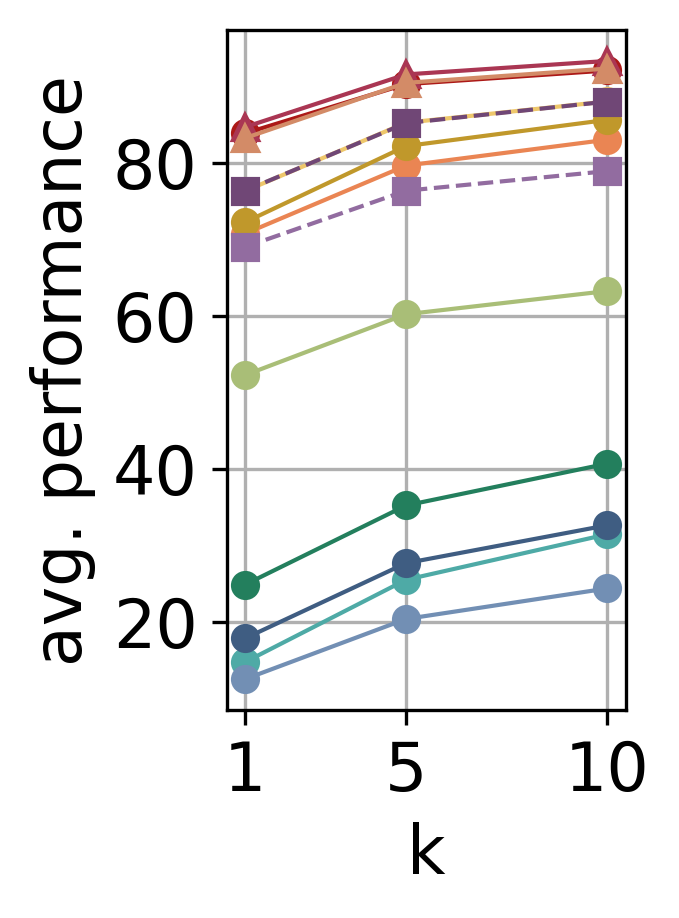} 
        
    \end{subfigure}
    \begin{subfigure}[t]{0.19\textwidth}
        \centering
        \caption{}
        \includegraphics[width=\linewidth]{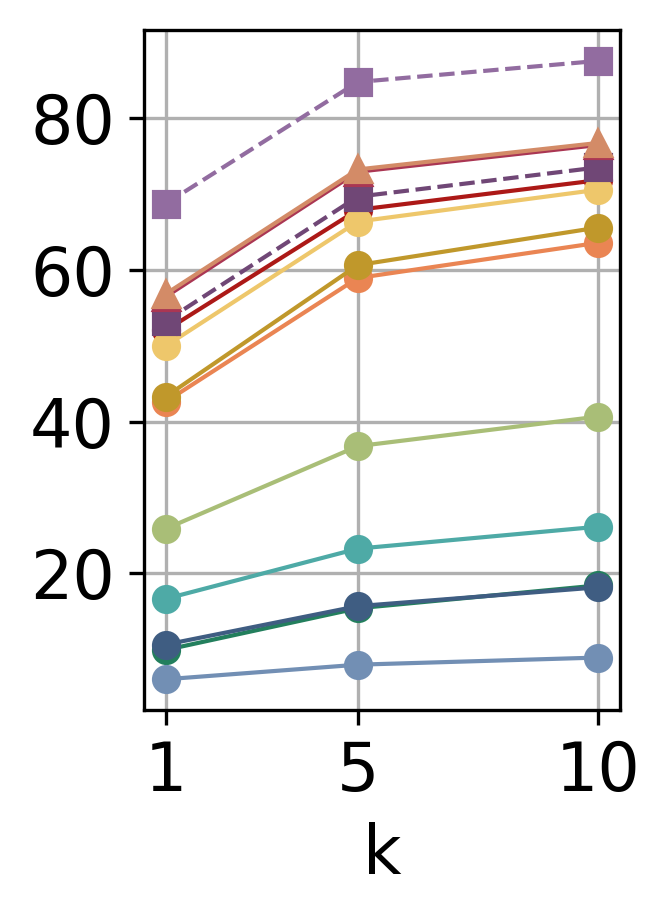} 
        
    \end{subfigure}
    \begin{subfigure}[t]{0.19\textwidth}
        \centering
        \caption{}
        \includegraphics[width=\linewidth]{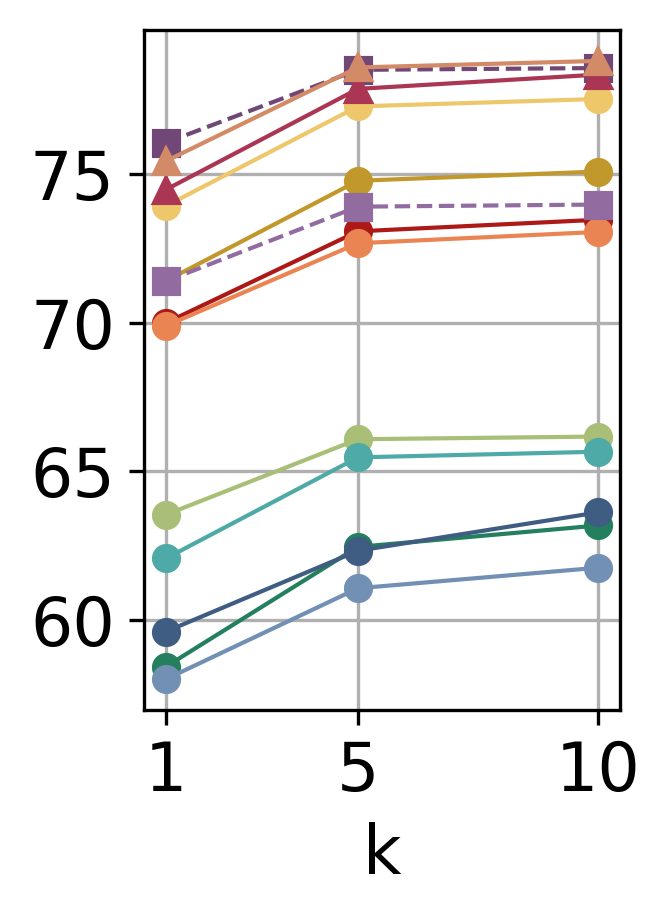} 
        
    \end{subfigure}
    \begin{subfigure}[t]{0.19\textwidth}
        \centering
        \caption{}
        \includegraphics[width=1.0\linewidth]{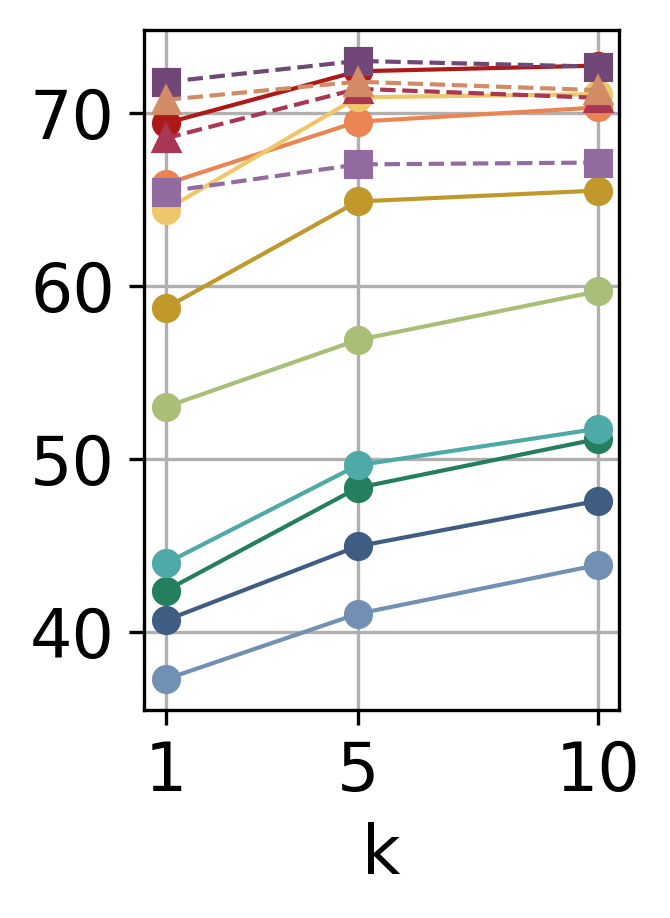} 
        
    \end{subfigure}
    \begin{subfigure}[t]
    {0.19\textwidth}
        \centering
        \caption{}
        \includegraphics[width=1.0\linewidth]{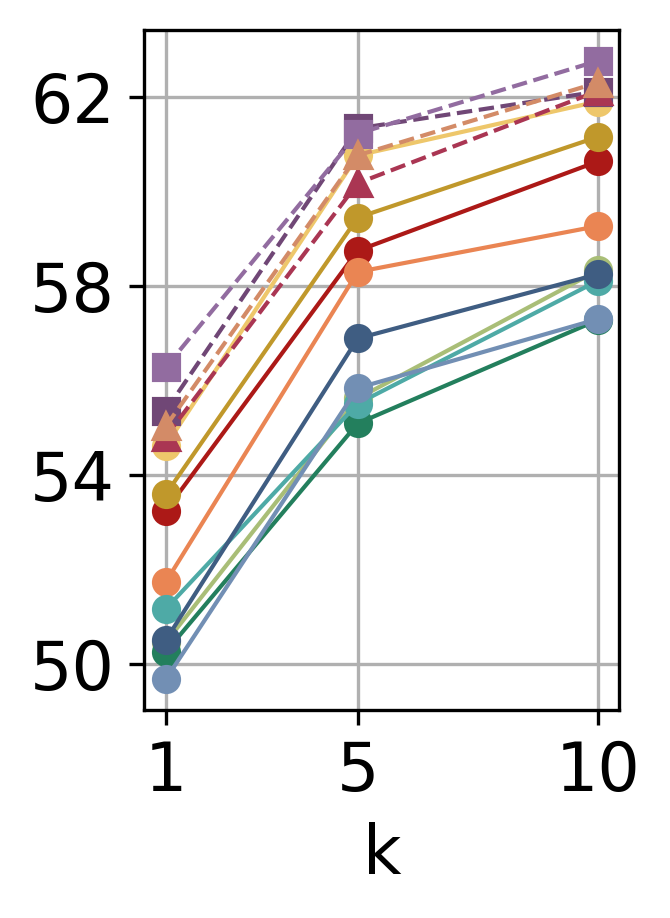} 
        
    \end{subfigure}
    \begin{subfigure}[t]{\textwidth}
        \centering
        \includegraphics[width=\linewidth]{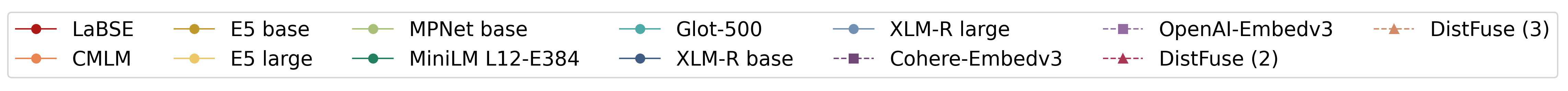} 
    \end{subfigure}
    \caption{Results with different $k=[1,5,10]$ on bitext retrieval: \textbf{(a)} cross-lingual and \textbf{(b)} code-switching, retrieval-based classification: \textbf{(c)} monolingual, \textbf{(d)} cross-lingual, and \textbf{(e)} code-switching.}
    \label{dynamics-k}
\end{figure*} 

\section{Results}

\subsection{Bitext Retrieval} 
Table~\ref{results-bitext-retrieval-class-overall} highlights DistFuse (2) and OpenAI-Embedv3-large as top performers in XS and CS tasks, respectively, with LaBSE ranking highest among open-source models. DistFuse (2) demonstrates superior performance across various settings. While XLM-R and Glot-500 struggle in bitext retrieval, they perform better in retrieval-based classification. Most models face challenges in CS tasks for both bitext retrieval and retrieval-based classification, where APIs generally perform slightly better. OpenAI-Embedv3 outperforms Cohere-Embedv3 on CS datasets. The specifics of CS training data remain unclear, potentially explaining the APIs' edge over open-source models. Combining model scores significantly boosts performance, with up to a 2.63\% improvement in bitext retrieval over LaBSE and a 1.72\% improvement over OpenAI-Embedv3. Similar gains are observed in retrieval-based classification, where the leading DistFuse model, though slightly behind Cohere-Embedv3, notably surpasses OpenAI-Embedv3.

\subsection{Retrieval-based Classification Results}
Table~\ref{results-bitext-retrieval-class-overall} illustrates that the Cohere-Embedv3 API outperforms all models by an average of 1.95\%, with LaBSE closely behind at 1.15\%. XLM-R and Glot-500 excel in classification tasks. Despite this, they lag behind models trained with contrastive learning or alignment objectives like LaBSE, CMLM, or E5 models, emphasizing the significance of text alignment in NLP tasks. Merging model scores notably boosts prediction accuracy, especially in Mono and XL settings. However, performance in CS and XL CS settings remains lower compared to API models. Additionally, our model outperforms fine-tuned models, requiring no fine-tuning in XL and CS tasks.

\subsection{ICL Classification Results}
Based on Table~\ref{results-icl}, we present the ICL classification results using E5$_\text{LARGE}$ as the retriever. Please see Appendix Table~\ref{results-icl-full} for results from alternate retrievers. The inclusion of few-shot context significantly improves the generative LM's precision in predicting class labels, leading to enhancements. There is a positive scaling law with increased model size in the one-shot setup. For instance, using a model with 6$\times$ more parameters (BLOOMZ 3B) boosts performance by 2.21\% compared to the top BLOOMZ 560m model. However, performance decreases for CS and XL CS tasks with increasing complexity. Despite focusing on English, Llama 3 and Llama 3.1 models generally outperform multilingual open-source models like BLOOMZ, mT0, XGLM, and Aya-23. BLOOMZ excels in the one-shot scenario, outperforming both Llama models. Notably, mT0 outperforms XGLM and Aya-23 in zero-shot settings, despite Aya-23's larger size. Aya-101 is the top open-source LM in both zero-shot and one-shot tasks, bridging the gap with commercial APIs like GPT-4o. Commercial generative LM APIs, such as GPT-3.5 Turbo and GPT-4o outperform all other models, particularly in CS and XL CS contexts. However, their superior performance may be attributed to prior exposure to these datasets, though this aspect remains unclear.

\begin{figure}[!t]
    \centering
    \begin{subfigure}[t]{0.235\textwidth}
        \centering
        \includegraphics[width=\linewidth]{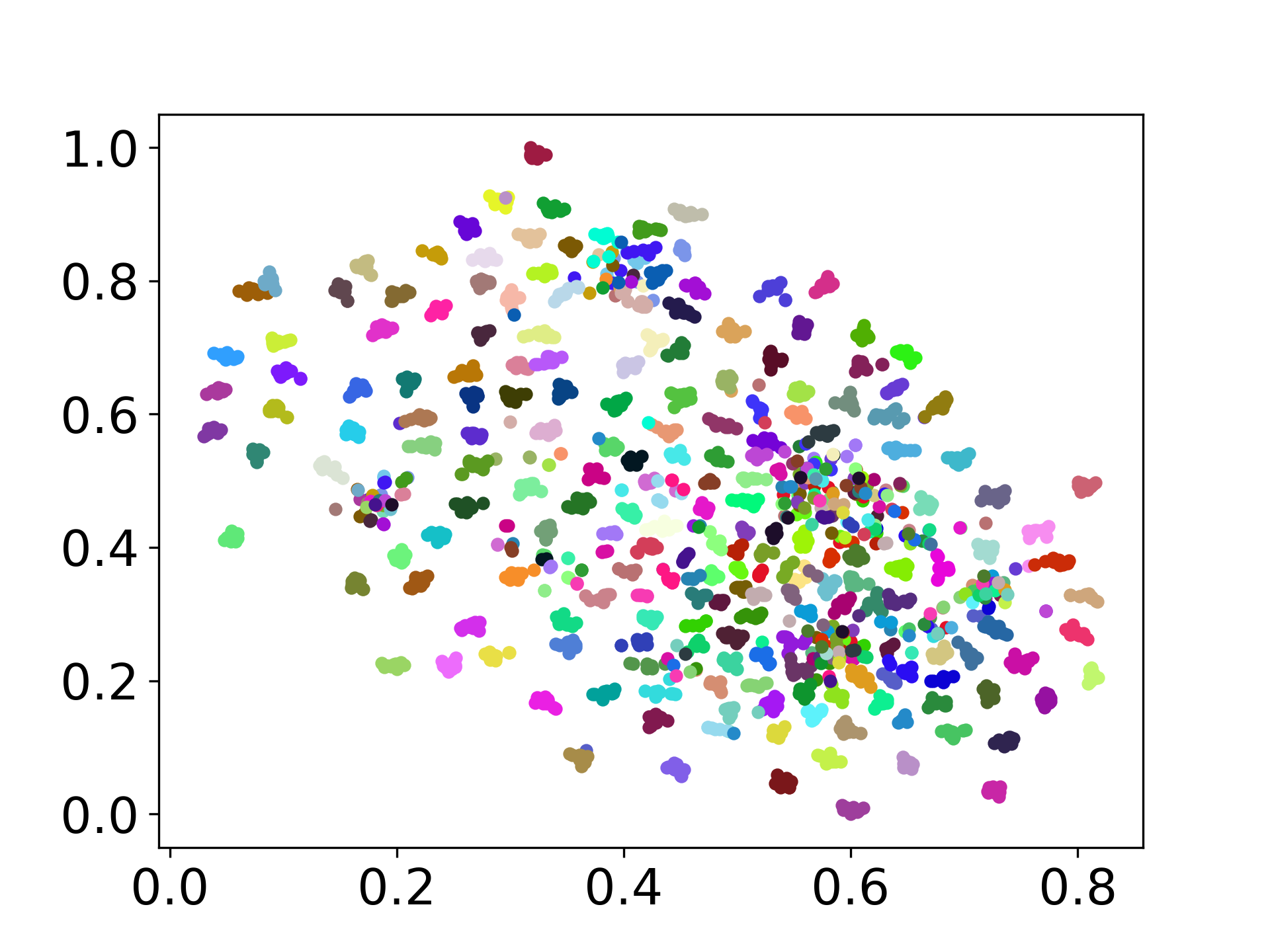} 
        \caption{E5$_\text{LARGE}$ (sample ID)}
    \end{subfigure}
    \begin{subfigure}[t]{0.235\textwidth}
        \centering
        \includegraphics[width=\linewidth]{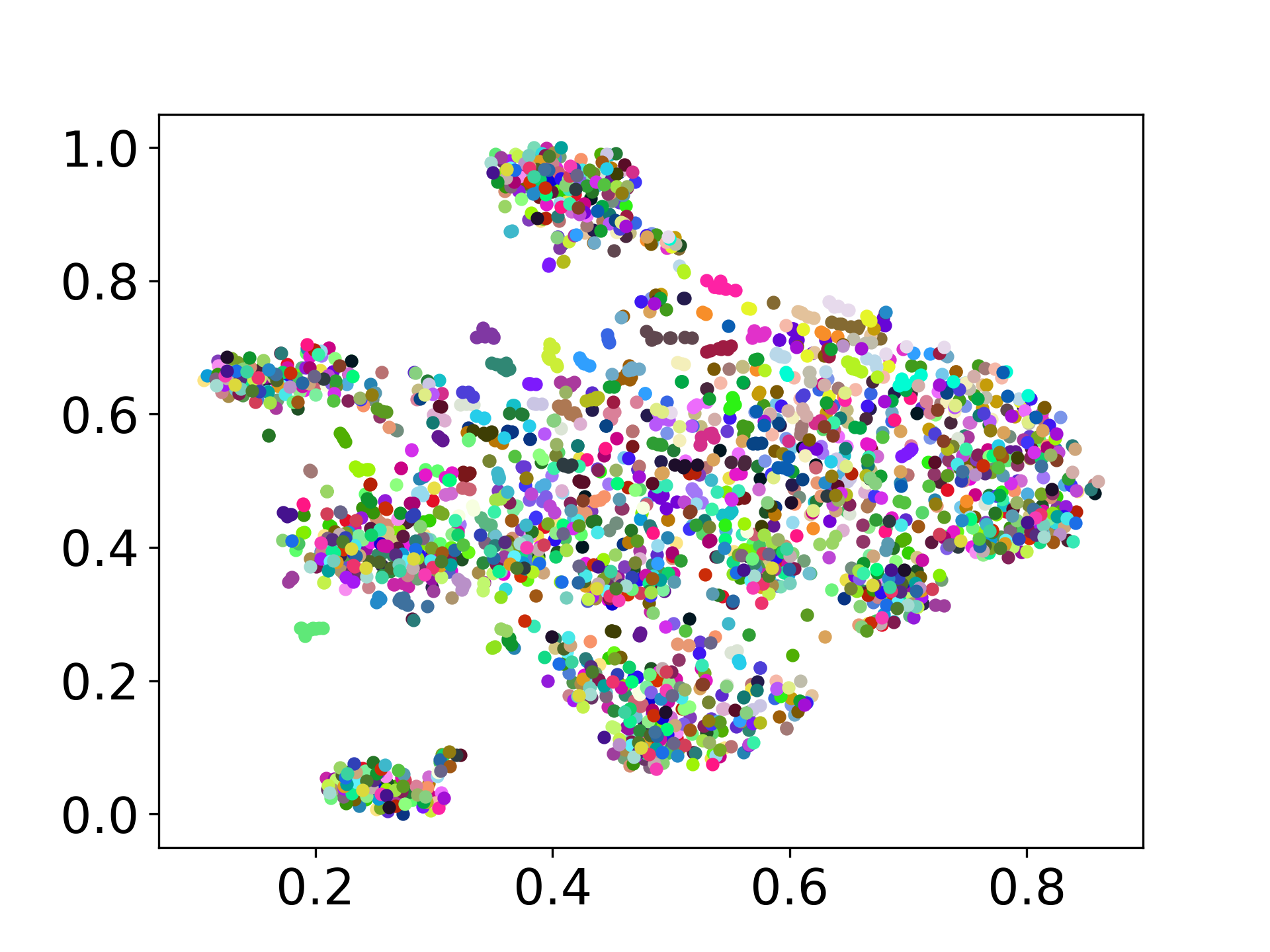} 
        \caption{XLM-R$_\text{BASE}$ (sample ID)}
    \end{subfigure}
    \begin{subfigure}[t]{0.235\textwidth}
        \centering
        \includegraphics[width=\linewidth]{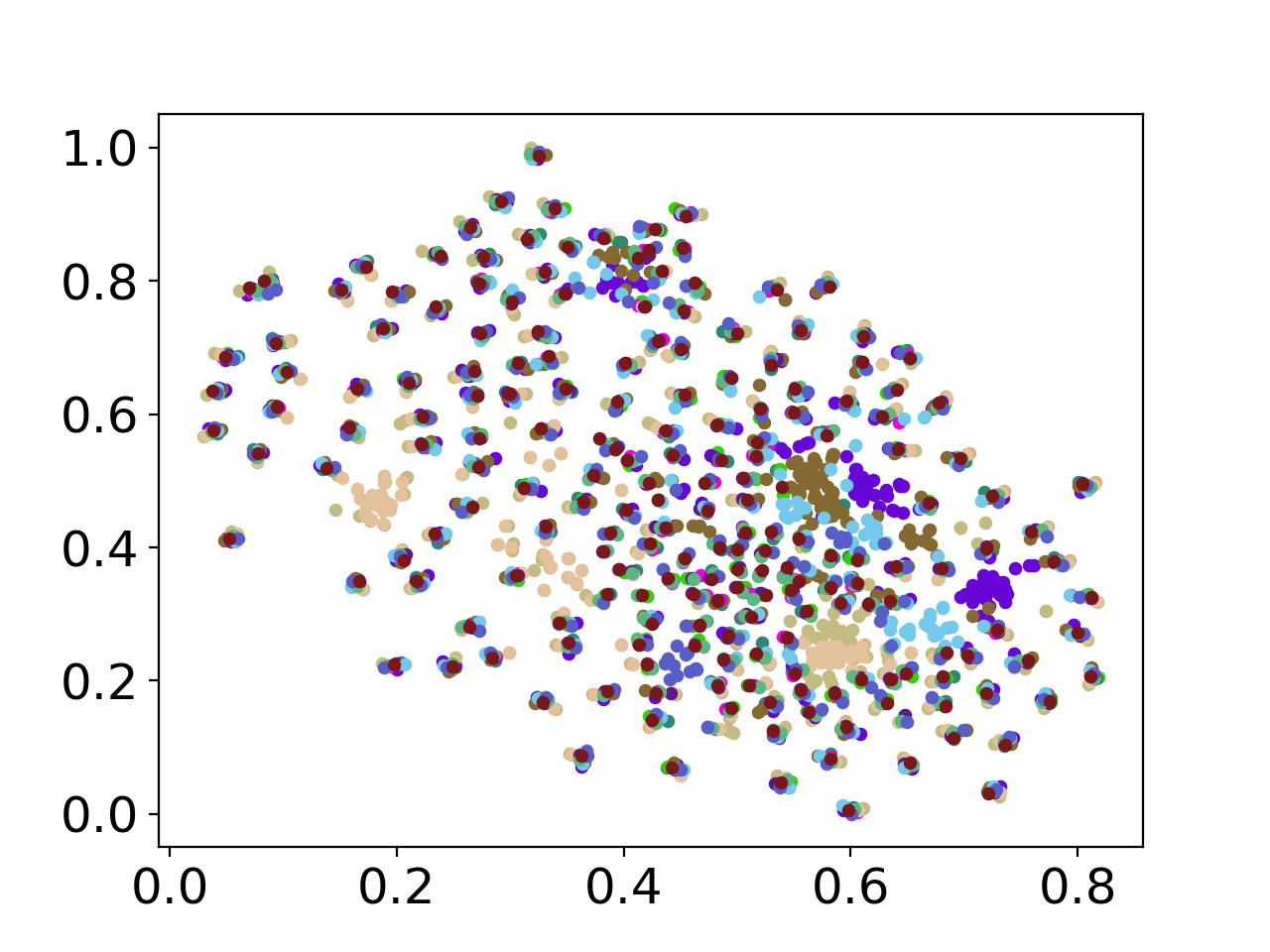} 
        \caption{E5$_\text{LARGE}$ (language)}
    \end{subfigure}
    \begin{subfigure}[t]{0.235\textwidth}
        \centering
        \includegraphics[width=\linewidth]{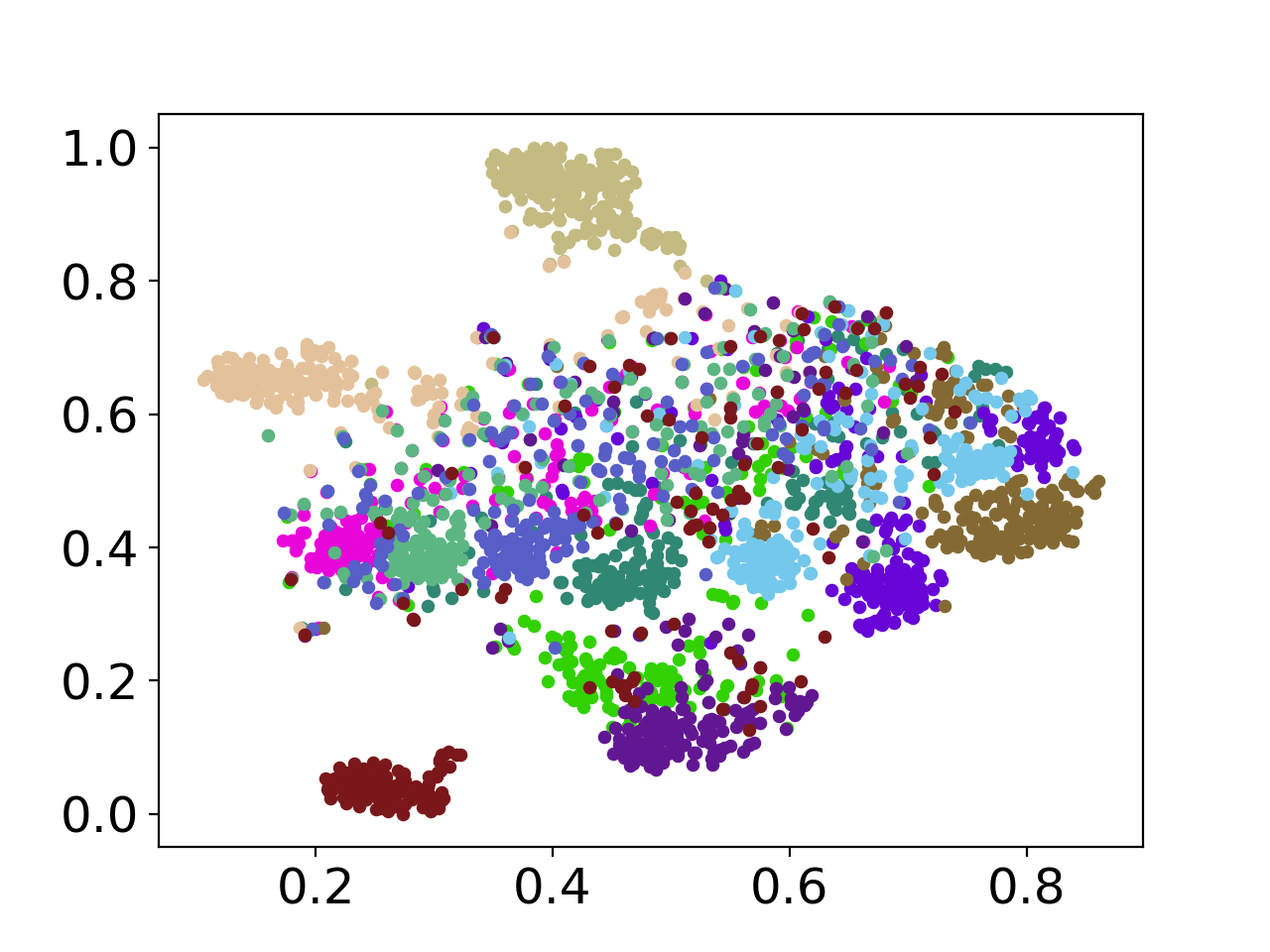} 
        \caption{XLM-R$_\text{BASE}$ (language)}
    \end{subfigure}
    \begin{subfigure}[t]{0.235\textwidth}
        \centering
        \includegraphics[width=\linewidth]{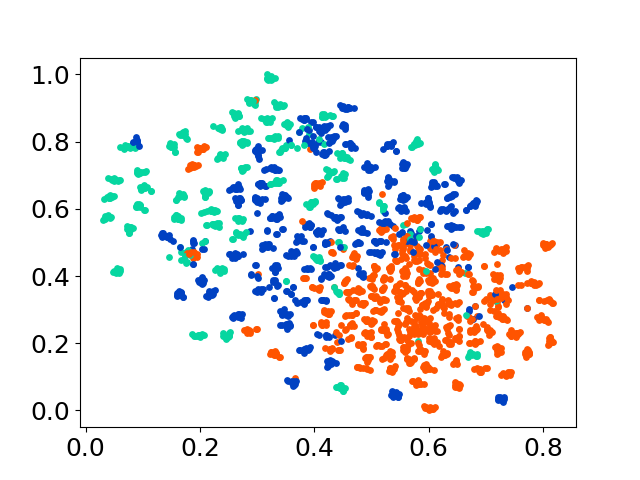} 
        \caption{E5$_\text{LARGE}$ (class label)}
    \end{subfigure}
    \begin{subfigure}[t]{0.235\textwidth}
        \centering
        \includegraphics[width=\linewidth]{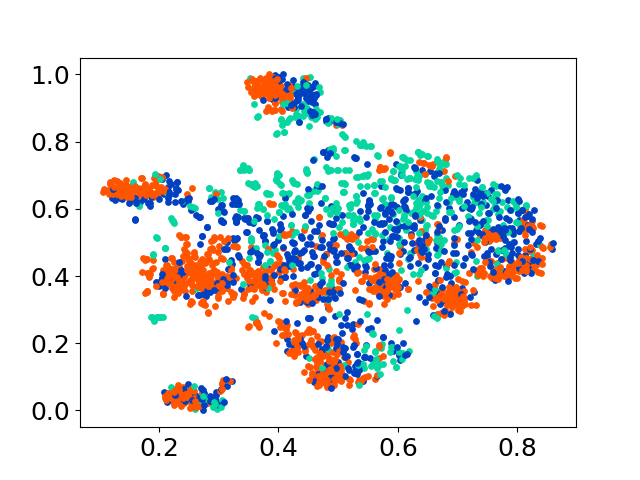} 
        \caption{XLM-R$_\text{BASE}$ (class label)}
    \end{subfigure}
    \caption{t-SNE representation of 200 randomly training samples from the NusaX dataset. The color on the figures show the sample ID for \textbf{(a)} and \textbf{(b)}, language for \textbf{(c)} and \textbf{(d)}, and class for \textbf{(e)} and \textbf{(f)}.}
\label{visualization-xlmr-e5-large}
\vspace{-3mm}
\end{figure}

\subsection{Performance Dynamics Over $k$}
Figure~\ref{dynamics-k} shows a consistent positive trend as the retrieved sample size increases for both bitext retrieval and retrieval-based classification tasks. This indicates that model performance improves with more retrieved samples. In bitext retrieval, a larger $k$ provides a richer set of bilingual text pairs, enhancing retrieval. Similarly, in retrieval-based classification, a larger $k$ offers more contextual examples, leading to more precise label predictions through majority voting.

\section{Further Analysis}

\begin{table}[!t]
\centering
\resizebox{.49\textwidth}{!}{
    \begin{tabular}{ll}
    \toprule
    \multicolumn{2}{l}{\textbf{Fine-tune}}
    \\ \midrule
    (1) Train & $n_{\text{epoch}} \times (|\mathcal{D}_{\text{train}}| \times{} (f_\mathcal{M} + b_\mathcal{M}) + |\mathcal{D}_{\text{dev}}| \times f_\mathcal{M}$) \\
    (2) Evaluate & $|\mathcal{D}_{\text{test}}| \times f_\mathcal{M}$
    \\ \midrule
    \multicolumn{2}{l}{\textbf{Retrieval-based Classification}}
    \\ \midrule
    (1) Generate vectors & $(|\mathcal{D}_{\text{train}}| + |\mathcal{D}_{\text{test}}|) \times f_\mathcal{M}$ \\
    (2) Retrieve samples & $|\mathcal{D}_{\text{train}}| \times |\mathcal{D}_{\text{test}}| \times (n_{dim} \times (p_{+} + p_{-} + p_\text{sq}) + p_{\sqrt{}})$ \\
    \midrule
    \multicolumn{2}{l}{\textbf{ICL Classification}}
    \\ \midrule
    (1) Generate vectors & $(|\mathcal{D}_{\text{train}}| + |\mathcal{D}_{\text{test}}|) \times f_\mathcal{M}$ \\
    (2) Retrieve samples & $|\mathcal{D}_{\text{train}}| \times |\mathcal{D}_{\text{test}}| \times (n_{dim} \times (p_{+} + p_{-} + p_\text{sq}) + p_{\sqrt{}})$ \\
    (3a) Generate probability & $|\mathcal{D}_{\text{test}}| \times f_\mathcal{G} \times |L| \times \bar{|L|}$ \\
    (3b) Generate responses & $|\mathcal{D}_{\text{test}}| \times f_\mathcal{G} \times \bar{|L|}$
    \\ \bottomrule
    \end{tabular}
}
\caption{FLOPs computation formulae. Here, \(n_\text{epoch}\) and \(n_\text{dim}\) denote the number of epochs and vector dimension, respectively. \(f_\mathcal{M}\) and \(b_\mathcal{M}\) represent the forward and backward FLOPs of model \(\mathcal{M}\), respectively. \(f_\mathcal{G}\) denotes the forward FLOPs of model \(\mathcal{G}\). The symbols \(p_{+}\), \(p_{-}\), \(p_\text{sq}\), and \(p_{\sqrt{}}\) indicate the FLOPs required to perform the operations of addition, subtraction, squaring, and square root, respectively. Additionally, \(|L|\) and \(\bar{|L|}\) denote the number of labels and the average sequence length of the labels, respectively. The variables \(|\mathcal{D}_\text{train}|\), \(|\mathcal{D}_\text{dev}|\), and \(|\mathcal{D}_\text{test}|\) represent the sizes of the train, development, and test data splits, respectively.}
\label{flops-formulae}
\vspace{-5mm}
\end{table}

\subsection{Model Representation}
Figure \ref{visualization-xlmr-e5-large} shows 2D scatter plots of the vector representation generated using t-SNE~\cite{van2008visualizing}. We take 200 random training samples from the NusaX dataset, reduce the high-dimensional vectors into 2D and color the scatter plots in three ways. \textbf{(1) By sample ID}. We assign the same color for parallel samples. \textbf{(2) By language}. We assign a color for each language. \textbf{(3) By class label}. We assign a color for each class label. We observe that the E5$_\text{LARGE}$ model forms small, color-coded clusters based on sample ID, indicating its proficiency in aligning text across different languages. In contrast, the $\text{XLM-R}_\text{BASE}$ model forms larger clusters where samples of the same language group closely together, suggesting it is more effective at identifying same-language data, even for unseen languages in NusaX. However, $\text{XLM-R}_\text{BASE}$ displays a sparse distribution when classifying samples by sample ID, aligning with our bitext retrieval task results. Both models effectively distinguish label classes, with E5$_\text{LARGE}$ achieving better color separation than $\text{XLM-R}_\text{BASE}$, as shown in Figures~\ref{visualization-xlmr-e5-large} (e) and (f). Similar findings are observed for other models. For more details, refer to Appendix~\ref{sec:more-visualization}.

\subsection{Samples Relevance}
\begin{figure}[!t]
    \centering
    \includegraphics[width=\linewidth]{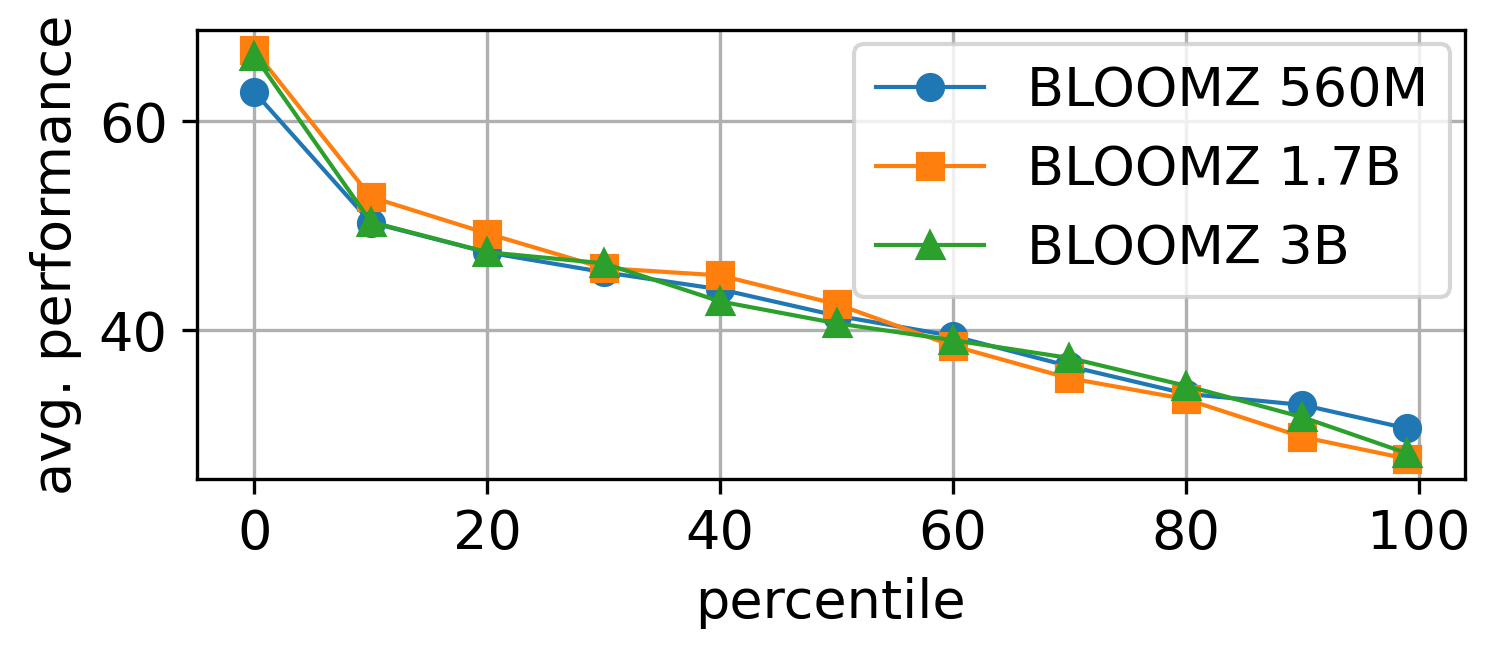} 
    \caption{ICL performance dynamics of BLOOMZ models on the NusaX dataset using context retrieved from various percentiles with E5$_\text{LARGE}$. Lower percentiles correspond to more semantically relevant samples.}
    \label{fig:percentile}
    \vspace{-5mm}
\end{figure}

Figure~\ref{fig:percentile} illustrates the performance dynamics of BLOOMZ models on the NusaX dataset when retrieving samples from various training data percentiles. Lower percentiles correspond to samples that are more semantically similar to the query. The results indicate that as the percentile decreases, performance improves consistently across all three models. This trend highlights the critical importance of retrieving highly relevant samples for in-context learning (ICL) tasks. By focusing on semantically aligned samples, the models are able to enhance the contextual understanding, which in turn leads to more accurate and reliable predictions. These findings highlight the potential benefits of optimizing sample retrieval strategies to improve model performance in various ICL applications.

\subsection{Compute Efficiency}
We aim to measure the theoretical time complexity by evaluating computation in terms of FLOPs (Floating Point Operations), irrespective of the machine configuration. Table~\ref{flops-formulae} details the components contributing to this calculation. The time complexity for fine-tuning a model scales with the number of training epochs, with more epochs significantly increasing complexity. The backward pass FLOPs, which are substantially higher than forward pass FLOPs, are a major factor. Retrieval-based classification is much more efficient, relying primarily on generating vector representations through forward passes. The retrieval process itself is efficient, with complexity influenced mainly by the sizes of the training and test datasets—factors typically smaller than the computational demands of fine-tuning. In contrast, ICL classification incurs higher inference costs due to the increased forward FLOPs of generative models. With very large LMs, the inference cost can even exceed that of fine-tuning. However, as the training data size increases, the complexity of fine-tuning eventually surpasses ICL model inference. For ICL classification, we have two methods: (a) computing label probabilities, which offers precise predictions, and (b) directly predicting labels through instructions, which is more efficient as responses match desired labels, eliminating the need to evaluate all options. While direct prediction may generate extraneous tokens, this can be mitigated with additional instructions to output only the label.

\subsection{Bitext Retrieval is Unsymmetrical}
We evaluate the bitext retrieval performance with different source and target language(s) directions. Based on the results presented in Table~\ref{bitext-retrieval-directions-results}, it is evident that the bitext retrieval performance is asymmetrical. Specifically, we observe that using non-English data to retrieve English data tends to be more effective than the reverse scenario.

\begin{table}[!t]
\centering
\resizebox{.49\textwidth}{!}{
    \begin{tabular}{lcc|cc}
    \toprule
    \textbf{Model} & \multicolumn{2}{c|}{\textbf{x$\rightarrow$eng}} & \multicolumn{2}{c}{\textbf{eng$\rightarrow$x}} \\ 
    & BUCC & Tatoeba & BUCC & Tatoeba \\ \midrule
    LaBSE & \textbf{98.77} & \textbf{83.76} & \textbf{98.93} & \textbf{80.31} \\
    E5$_\text{LARGE}$ & 98.66 & 75.73 & 98.90 & 75.98 \\
    Glot-500 & 17.90 & 10.58 & 16.39 & 14.07\\
    XLM-R$_\text{BASE}$ & 39.70 & 12.62 & 24.70 & 8.61 \\ 
    XLM-R$_\text{LARGE}$ & 26.51 & 6.57 & 11.95 & 3.30 \\
    \midrule
    Cohere-Embedv3 & 98.76 & 74.66 & 98.89 & 76.43
    \\ \bottomrule
    \end{tabular}
}
\caption{Bitext retrieval F1@1 performance on two different source-to-target language(s) directions. \textbf{Bold} and \underline{underlined} numbers present the best and second-best models.} 
\label{bitext-retrieval-directions-results}
\end{table}

\section{Related Work}

\paragraph{Dense Retrieval via LM} Dense retrieval has marked a significant advancement in information retrieval, enabling rapid sample searches across vast document collections. Research has focused on training objectives and architectures that produce similarity scores between text samples. \citet{reimers2019sentence} introduce a Siamese network architecture trained with contrastive learning, enhancing retrieval by enabling vector representation comparison using similarity measures, applied to BERT~\cite{kenton2019bert}. Efforts to improve alignment include incorporating annotated pairs from natural language inference datasets using SimCSE loss~\cite{gao2021simcse}. Furthermore, \citet{feng2022language} propose combining monolingual and translation alignment losses to enhance performance, such as masked language modeling (MLM)~\cite{devlin2019bert} and translation language modeling (TLM) objectives~\cite{conneau2019cross}, dual encoder translation ranking~\cite{guo2018effective}, and additive margin softmax~\cite{yang2019improving}.
\citet{khattab2020colbert} introduce a late interaction paradigm, comparing embedding representations via vector similarity indexes for relevance estimation in ranking tasks. \citet{wang2024multilingual} further innovate by using in-batch negatives to leverage weakly supervised data from diverse, heterogeneous sources.

\paragraph{Semantic Retrieval for NLP Tasks}
Retrieving labels using semantic retrieval has proven beneficial for classification. \citet{bari2021nearest} enhance accuracy with cross-lingual few-shot nearest neighbor adaptation. \citet{winata2023efficient} predict test data labels efficiently using English training data without prior adaptation via ICL. \citet{li2023unified} introduce a ranking framework to retrieve high-quality demonstrations for various tasks. Building on these methods, we adopt a straightforward and efficient retrieval approach similar to \citet{winata2023efficient}, supporting multiple retrieval models for open-source tools and APIs. We extend this approach to the ICL setting, enhancing its utility and accessibility across diverse scenarios.

\section{Conclusion}
This paper introduces \benchmarknameonly{}, a benchmark for evaluating the efficacy of multilingual LMs in semantic retrieval tasks, including bitext retrieval and classification through semantic search and retrieval-augmented contexts. Our framework rigorously assesses LMs' robustness in retrieving samples from over 200 languages. Empirical results demonstrate that our method, which focuses on retrieving semantically similar vector representations, achieves performance comparable to state-of-the-art fine-tuned approaches, without requiring fine-tuning across multiple datasets and languages. We also explore the mechanisms behind these representations, offering insights to improve the efficiency and accuracy of label retrieval methods. Our research aims to pave the way for future exploration and optimization in semantic retrieval and classification, ultimately contributing to more robust and adaptable NLP systems.

\section*{Limitations}
We have identified potential avenues for enhancing the performance of the ICL classification task through the application of ensemble techniques such as DistFuse and using the target language prompts instead of English. Additionally, while we have primarily focused on evaluating the BLOOMZ, mT0, XGLM, Gemma, Llama 3, Llama 3.1, Aya-23, Aya-101, Command-R, GPT-3.5 Turbo, and GPT-4o models within the benchmark, we acknowledge that there may be other models that could also yield promising results. These aspects represent areas for future exploration and expansion of our research efforts. Due to resource limitations and simplicity, we only test a single prompt template. Running with various prompts could yield different results, but we defer this exploration to future research.

In the future, we plan to explore deeper into the capabilities of ensemble techniques like DistFuse to further improve the performance of the ICL classification task. By combining the strengths of multiple models, we aim to enhance the robustness and accuracy of our classification outcomes, ultimately achieving better results in real-world applications. Furthermore, our current evaluation has been limited to a select few models and datasets as part of our initial assessment phase. However, we recognize the importance of conducting a more comprehensive evaluation by considering a wider range of models and datasets. This will allow us to gain a more comprehensive understanding of the strengths and weaknesses of different approaches, enabling us to make more informed decisions about model selection and optimization strategies.

\section*{Ethical Considerations}
Our research aims to evaluate LMs in the context of multilingual semantic retrieval, a field with significant implications for diverse multilingual communities. We strive to ensure that our evaluation is conducted with the utmost transparency and fairness.

\section*{Acknowledgments}
We would like to extend our gratitude to Zheng-Xin Yong for his insightful comments and suggestions on this work. We also appreciate the insightful reviews from the anonymous reviewers, which have contributed to improving the quality of this work.


\bibliography{anthology,custom}

\begin{thebibliography}{62}
\expandafter\ifx\csname natexlab\endcsname\relax\def\natexlab#1{#1}\fi

\bibitem[{Adelani et~al.(2022)Adelani, Neubig, Ruder, Rijhwani, Beukman, Palen-Michel, Lignos, Alabi, Muhammad, Nabende et~al.}]{adelani2022masakhaner}
David Adelani, Graham Neubig, Sebastian Ruder, Shruti Rijhwani, Michael Beukman, Chester Palen-Michel, Constantine Lignos, Jesujoba Alabi, Shamsuddeen Muhammad, Peter Nabende, et~al. 2022.
\newblock Masakhaner 2.0: Africa-centric transfer learning for named entity recognition.
\newblock In \emph{Proceedings of the 2022 Conference on Empirical Methods in Natural Language Processing}, pages 4488--4508.

\bibitem[{Adelani et~al.(2023)Adelani, Liu, Shen, Vassilyev, Alabi, Mao, Gao, and Lee}]{adelani2023sib}
David~Ifeoluwa Adelani, Hannah Liu, Xiaoyu Shen, Nikita Vassilyev, Jesujoba~O Alabi, Yanke Mao, Haonan Gao, and Annie En-Shiun Lee. 2023.
\newblock Sib-200: A simple, inclusive, and big evaluation dataset for topic classification in 200+ languages and dialects.
\newblock \emph{arXiv preprint arXiv:2309.07445}.

\bibitem[{Aguilar et~al.(2020)Aguilar, Kar, and Solorio}]{aguilar2020lince}
Gustavo Aguilar, Sudipta Kar, and Thamar Solorio. 2020.
\newblock Lince: A centralized benchmark for linguistic code-switching evaluation.
\newblock In \emph{Proceedings of the Twelfth Language Resources and Evaluation Conference}, pages 1803--1813.

\bibitem[{Alabi et~al.(2022)Alabi, Adelani, Mosbach, and Klakow}]{alabi2022adapting}
Jesujoba~O Alabi, David~Ifeoluwa Adelani, Marius Mosbach, and Dietrich Klakow. 2022.
\newblock Adapting pre-trained language models to african languages via multilingual adaptive fine-tuning.
\newblock In \emph{Proceedings of the 29th International Conference on Computational Linguistics}, pages 4336--4349.

\bibitem[{Ansell et~al.(2021)Ansell, Ponti, Pfeiffer, Ruder, Glava{\v{s}}, Vuli{\'c}, and Korhonen}]{ansell2021mad}
Alan Ansell, Edoardo~Maria Ponti, Jonas Pfeiffer, Sebastian Ruder, Goran Glava{\v{s}}, Ivan Vuli{\'c}, and Anna Korhonen. 2021.
\newblock Mad-g: Multilingual adapter generation for efficient cross-lingual transfer.
\newblock In \emph{Findings of the Association for Computational Linguistics: EMNLP 2021}, pages 4762--4781.

\bibitem[{Aryabumi et~al.(2024)Aryabumi, Dang, Talupuru, Dash, Cairuz, Lin, Venkitesh, Smith, Marchisio, Ruder et~al.}]{aryabumi2024aya}
Viraat Aryabumi, John Dang, Dwarak Talupuru, Saurabh Dash, David Cairuz, Hangyu Lin, Bharat Venkitesh, Madeline Smith, Kelly Marchisio, Sebastian Ruder, et~al. 2024.
\newblock Aya 23: Open weight releases to further multilingual progress.
\newblock \emph{arXiv preprint arXiv:2405.15032}.

\bibitem[{Bari et~al.(2021)Bari, Haider, and Mansour}]{bari2021nearest}
M~Saiful Bari, Batool Haider, and Saab Mansour. 2021.
\newblock Nearest neighbour few-shot learning for cross-lingual classification.
\newblock In \emph{Proceedings of the 2021 Conference on Empirical Methods in Natural Language Processing}, pages 1745--1753.

\bibitem[{Bevilacqua et~al.(2022)Bevilacqua, Ottaviano, Lewis, Yih, Riedel, and Petroni}]{bevilacqua2022autoregressive}
Michele Bevilacqua, Giuseppe Ottaviano, Patrick Lewis, Scott Yih, Sebastian Riedel, and Fabio Petroni. 2022.
\newblock Autoregressive search engines: Generating substrings as document identifiers.
\newblock \emph{Advances in Neural Information Processing Systems}, 35:31668--31683.

\bibitem[{Biderman et~al.(2024)Biderman, Schoelkopf, Sutawika, Gao, Tow, Abbasi, Aji, Ammanamanchi, Black, Clive et~al.}]{biderman2024lessons}
Stella Biderman, Hailey Schoelkopf, Lintang Sutawika, Leo Gao, Jonathan Tow, Baber Abbasi, Alham~Fikri Aji, Pawan~Sasanka Ammanamanchi, Sidney Black, Jordan Clive, et~al. 2024.
\newblock Lessons from the trenches on reproducible evaluation of language models.
\newblock \emph{arXiv preprint arXiv:2405.14782}.

\bibitem[{Cahyawijaya et~al.(2024)Cahyawijaya, Lovenia, and Fung}]{cahyawijaya2024llms}
Samuel Cahyawijaya, Holy Lovenia, and Pascale Fung. 2024.
\newblock Llms are few-shot in-context low-resource language learners.
\newblock \emph{arXiv preprint arXiv:2403.16512}.

\bibitem[{Cahyawijaya et~al.(2023)Cahyawijaya, Lovenia, Koto, Adhista, Dave, Oktavianti, Akbar, Lee, Shadieq, Cenggoro et~al.}]{cahyawijaya2023nusawrites}
Samuel Cahyawijaya, Holy Lovenia, Fajri Koto, Dea Adhista, Emmanuel Dave, Sarah Oktavianti, Salsabil Akbar, Jhonson Lee, Nuur Shadieq, Tjeng~Wawan Cenggoro, et~al. 2023.
\newblock Nusawrites: Constructing high-quality corpora for underrepresented and extremely low-resource languages.
\newblock In \emph{Proceedings of the 13th International Joint Conference on Natural Language Processing and the 3rd Conference of the Asia-Pacific Chapter of the Association for Computational Linguistics (Volume 1: Long Papers)}, pages 921--945.

\bibitem[{Cer et~al.(2017)Cer, Diab, Agirre, Lopez-Gazpio, and Specia}]{cer2017semeval}
Daniel Cer, Mona Diab, Eneko Agirre, I{\~n}igo Lopez-Gazpio, and Lucia Specia. 2017.
\newblock Semeval-2017 task 1: Semantic textual similarity multilingual and crosslingual focused evaluation.
\newblock In \emph{Proceedings of the 11th International Workshop on Semantic Evaluation (SemEval-2017)}, pages 1--14.

\bibitem[{Cer et~al.(2018)Cer, Yang, Kong, Hua, Limtiaco, John, Constant, Guajardo-Cespedes, Yuan, Tar et~al.}]{cer2018universal}
Daniel Cer, Yinfei Yang, Sheng-yi Kong, Nan Hua, Nicole Limtiaco, Rhomni~St John, Noah Constant, Mario Guajardo-Cespedes, Steve Yuan, Chris Tar, et~al. 2018.
\newblock Universal sentence encoder for english.
\newblock In \emph{Proceedings of the 2018 conference on empirical methods in natural language processing: system demonstrations}, pages 169--174.

\bibitem[{Chakravarthi et~al.(2020)Chakravarthi, Jose, Suryawanshi, Sherly, and McCrae}]{chakravarthi2020sentiment}
Bharathi~Raja Chakravarthi, Navya Jose, Shardul Suryawanshi, Elizabeth Sherly, and John~Philip McCrae. 2020.
\newblock A sentiment analysis dataset for code-mixed malayalam-english.
\newblock In \emph{Proceedings of the 1st Joint Workshop on Spoken Language Technologies for Under-resourced languages (SLTU) and Collaboration and Computing for Under-Resourced Languages (CCURL)}, pages 177--184.

\bibitem[{Conneau and Lample(2019)}]{conneau2019cross}
Alexis Conneau and Guillaume Lample. 2019.
\newblock Cross-lingual language model pretraining.
\newblock \emph{Advances in neural information processing systems}, 32.

\bibitem[{Dettmers et~al.(2022)Dettmers, Lewis, Belkada, and Zettlemoyer}]{dettmers2022gpt3}
Tim Dettmers, Mike Lewis, Younes Belkada, and Luke Zettlemoyer. 2022.
\newblock Gpt3. int8 (): 8-bit matrix multiplication for transformers at scale.
\newblock \emph{Advances in Neural Information Processing Systems}, 35:30318--30332.

\bibitem[{Devlin et~al.(2019)Devlin, Chang, Lee, and Toutanova}]{devlin2019bert}
Jacob Devlin, Ming-Wei Chang, Kenton Lee, and Kristina Toutanova. 2019.
\newblock Bert: Pre-training of deep bidirectional transformers for language understanding.
\newblock In \emph{Proceedings of the 2019 Conference of the North American Chapter of the Association for Computational Linguistics: Human Language Technologies, Volume 1 (Long and Short Papers)}, pages 4171--4186.

\bibitem[{Dubey et~al.(2024)Dubey, Jauhri, Pandey, Kadian, Al-Dahle, Letman, Mathur, Schelten, Yang, Fan et~al.}]{dubey2024llama}
Abhimanyu Dubey, Abhinav Jauhri, Abhinav Pandey, Abhishek Kadian, Ahmad Al-Dahle, Aiesha Letman, Akhil Mathur, Alan Schelten, Amy Yang, Angela Fan, et~al. 2024.
\newblock The llama 3 herd of models.
\newblock \emph{arXiv preprint arXiv:2407.21783}.

\bibitem[{Feng et~al.(2022)Feng, Yang, Cer, Arivazhagan, and Wang}]{feng2022language}
Fangxiaoyu Feng, Yinfei Yang, Daniel Cer, Naveen Arivazhagan, and Wei Wang. 2022.
\newblock Language-agnostic bert sentence embedding.
\newblock In \emph{Proceedings of the 60th Annual Meeting of the Association for Computational Linguistics (Volume 1: Long Papers)}, pages 878--891.

\bibitem[{FitzGerald et~al.(2023)FitzGerald, Hench, Peris, Mackie, Rottmann, Sanchez, Nash, Urbach, Kakarala, Singh et~al.}]{fitzgerald2023massive}
Jack FitzGerald, Christopher Hench, Charith Peris, Scott Mackie, Kay Rottmann, Ana Sanchez, Aaron Nash, Liam Urbach, Vishesh Kakarala, Richa Singh, et~al. 2023.
\newblock Massive: A 1m-example multilingual natural language understanding dataset with 51 typologically-diverse languages.
\newblock In \emph{Proceedings of the 61st Annual Meeting of the Association for Computational Linguistics (Volume 1: Long Papers)}, pages 4277--4302.

\bibitem[{Gao et~al.(2021)Gao, Yao, and Chen}]{gao2021simcse}
Tianyu Gao, Xingcheng Yao, and Danqi Chen. 2021.
\newblock Simcse: Simple contrastive learning of sentence embeddings.
\newblock In \emph{Proceedings of the 2021 Conference on Empirical Methods in Natural Language Processing}, pages 6894--6910.

\bibitem[{Guo et~al.(2018)Guo, Shen, Yang, Ge, Cer, Abrego, Stevens, Constant, Sung, Strope et~al.}]{guo2018effective}
Mandy Guo, Qinlan Shen, Yinfei Yang, Heming Ge, Daniel Cer, Gustavo~Hernandez Abrego, Keith Stevens, Noah Constant, Yun-Hsuan Sung, Brian Strope, et~al. 2018.
\newblock Effective parallel corpus mining using bilingual sentence embeddings.
\newblock In \emph{Proceedings of the Third Conference on Machine Translation: Research Papers}, pages 165--176.

\bibitem[{Hegde et~al.(2022)Hegde, Anusha, Coelho, Shashirekha, and Chakravarthi}]{hegde2022corpus}
Asha Hegde, Mudoor~Devadas Anusha, Sharal Coelho, Hosahalli~Lakshmaiah Shashirekha, and Bharathi~Raja Chakravarthi. 2022.
\newblock Corpus creation for sentiment analysis in code-mixed tulu text.
\newblock In \emph{Proceedings of the 1st Annual Meeting of the ELRA/ISCA Special Interest Group on Under-Resourced Languages}, pages 33--40.

\bibitem[{ImaniGooghari et~al.(2023)ImaniGooghari, Lin, Kargaran, Severini, Sabet, Kassner, Ma, Schmid, Martins, Yvon et~al.}]{imanigooghari2023glot500}
Ayyoob ImaniGooghari, Peiqin Lin, Amir~Hossein Kargaran, Silvia Severini, Masoud~Jalili Sabet, Nora Kassner, Chunlan Ma, Helmut Schmid, Andr{\'e}~FT Martins, Fran{\c{c}}ois Yvon, et~al. 2023.
\newblock Glot500: Scaling multilingual corpora and language models to 500 languages.
\newblock In \emph{Proceedings of the 61st Annual Meeting of the Association for Computational Linguistics (Volume 1: Long Papers)}, pages 1082--1117.

\bibitem[{Kenton and Toutanova(2019)}]{kenton2019bert}
Jacob Devlin Ming-Wei~Chang Kenton and Lee~Kristina Toutanova. 2019.
\newblock Bert: Pre-training of deep bidirectional transformers for language understanding.
\newblock In \emph{Proceedings of NAACL-HLT}, pages 4171--4186.

\bibitem[{Khanuja et~al.(2020)Khanuja, Dandapat, Srinivasan, Sitaram, and Choudhury}]{khanuja2020gluecos}
Simran Khanuja, Sandipan Dandapat, Anirudh Srinivasan, Sunayana Sitaram, and Monojit Choudhury. 2020.
\newblock Gluecos: An evaluation benchmark for code-switched nlp.
\newblock In \emph{Proceedings of the 58th Annual Meeting of the Association for Computational Linguistics}, pages 3575--3585.

\bibitem[{Khattab and Zaharia(2020)}]{khattab2020colbert}
Omar Khattab and Matei Zaharia. 2020.
\newblock Colbert: Efficient and effective passage search via contextualized late interaction over bert.
\newblock In \emph{Proceedings of the 43rd International ACM SIGIR conference on research and development in Information Retrieval}, pages 39--48.

\bibitem[{Le~Scao et~al.(2023)Le~Scao, Fan, Akiki, Pavlick, Ili{\'c}, Hesslow, Castagn{\'e}, Luccioni, Yvon, Gall{\'e} et~al.}]{le2023bloom}
Teven Le~Scao, Angela Fan, Christopher Akiki, Ellie Pavlick, Suzana Ili{\'c}, Daniel Hesslow, Roman Castagn{\'e}, Alexandra~Sasha Luccioni, Fran{\c{c}}ois Yvon, Matthias Gall{\'e}, et~al. 2023.
\newblock Bloom: A 176b-parameter open-access multilingual language model.
\newblock \emph{arXiv preprint arXiv:2211.05100}.

\bibitem[{Lewis et~al.(2020)Lewis, Perez, Piktus, Petroni, Karpukhin, Goyal, K{\"u}ttler, Lewis, Yih, Rockt{\"a}schel et~al.}]{lewis2020retrieval}
Patrick Lewis, Ethan Perez, Aleksandra Piktus, Fabio Petroni, Vladimir Karpukhin, Naman Goyal, Heinrich K{\"u}ttler, Mike Lewis, Wen-tau Yih, Tim Rockt{\"a}schel, et~al. 2020.
\newblock Retrieval-augmented generation for knowledge-intensive nlp tasks.
\newblock \emph{Advances in Neural Information Processing Systems}, 33:9459--9474.

\bibitem[{Li et~al.(2023)Li, Lv, Yan, Lin, Zhu, Ni, Xie, Wang, and Qiu}]{li2023unified}
Xiaonan Li, Kai Lv, Hang Yan, Tianyang Lin, Wei Zhu, Yuan Ni, Guotong Xie, Xiaoling Wang, and Xipeng Qiu. 2023.
\newblock Unified demonstration retriever for in-context learning.
\newblock In \emph{Proceedings of the 61st Annual Meeting of the Association for Computational Linguistics (Volume 1: Long Papers)}, pages 4644--4668.

\bibitem[{Lin et~al.(2021)Lin, Mihaylov, Artetxe, Wang, Chen, Simig, Ott, Goyal, Bhosale, Du et~al.}]{lin2021few}
Xi~Victoria Lin, Todor Mihaylov, Mikel Artetxe, Tianlu Wang, Shuohui Chen, Daniel Simig, Myle Ott, Naman Goyal, Shruti Bhosale, Jingfei Du, et~al. 2021.
\newblock Few-shot learning with multilingual language models.
\newblock \emph{arXiv preprint arXiv:2112.10668}.

\bibitem[{Muennighoff et~al.(2023{\natexlab{a}})Muennighoff, Tazi, Magne, and Reimers}]{muennighoff2023mteb}
Niklas Muennighoff, Nouamane Tazi, Loic Magne, and Nils Reimers. 2023{\natexlab{a}}.
\newblock Mteb: Massive text embedding benchmark.
\newblock In \emph{Proceedings of the 17th Conference of the European Chapter of the Association for Computational Linguistics}, pages 2014--2037.

\bibitem[{Muennighoff et~al.(2023{\natexlab{b}})Muennighoff, Wang, Sutawika, Roberts, Biderman, Le~Scao, Bari, Shen, Yong, Schoelkopf et~al.}]{muennighoff2023crosslingual}
Niklas Muennighoff, Thomas Wang, Lintang Sutawika, Adam Roberts, Stella Biderman, Teven Le~Scao, M~Saiful Bari, Sheng Shen, Zheng~Xin Yong, Hailey Schoelkopf, et~al. 2023{\natexlab{b}}.
\newblock Crosslingual generalization through multitask finetuning.
\newblock In \emph{Proceedings of the 61st Annual Meeting of the Association for Computational Linguistics (Volume 1: Long Papers)}, pages 15991--16111.

\bibitem[{Patwa et~al.(2020)Patwa, Aguilar, Kar, Pandey, Pykl, Gamb{\"a}ck, Chakraborty, Solorio, and Das}]{patwa2020semeval}
Parth Patwa, Gustavo Aguilar, Sudipta Kar, Suraj Pandey, Srinivas Pykl, Bj{\"o}rn Gamb{\"a}ck, Tanmoy Chakraborty, Thamar Solorio, and Amitava Das. 2020.
\newblock Semeval-2020 task 9: Overview of sentiment analysis of code-mixed tweets.
\newblock In \emph{Proceedings of the Fourteenth Workshop on Semantic Evaluation}, pages 774--790.

\bibitem[{Reimers and Gurevych(2019)}]{reimers2019sentence}
Nils Reimers and Iryna Gurevych. 2019.
\newblock Sentence-bert: Sentence embeddings using siamese bert-networks.
\newblock In \emph{Proceedings of the 2019 Conference on Empirical Methods in Natural Language Processing and the 9th International Joint Conference on Natural Language Processing (EMNLP-IJCNLP)}, pages 3982--3992.

\bibitem[{Ruder et~al.(2021)Ruder, Constant, Botha, Siddhant, Firat, Fu, Liu, Hu, Garrette, Neubig et~al.}]{ruder2021xtreme}
Sebastian Ruder, Noah Constant, Jan Botha, Aditya Siddhant, Orhan Firat, Jinlan Fu, Pengfei Liu, Junjie Hu, Dan Garrette, Graham Neubig, et~al. 2021.
\newblock Xtreme-r: Towards more challenging and nuanced multilingual evaluation.
\newblock In \emph{Proceedings of the 2021 Conference on Empirical Methods in Natural Language Processing}, pages 10215--10245.

\bibitem[{Shode et~al.(2023)Shode, Adelani, Peng, and Feldman}]{shode2023nollysenti}
Iyanuoluwa Shode, David~Ifeoluwa Adelani, Jing Peng, and Anna Feldman. 2023.
\newblock Nollysenti: Leveraging transfer learning and machine translation for nigerian movie sentiment classification.
\newblock In \emph{Proceedings of the 61st Annual Meeting of the Association for Computational Linguistics (Volume 2: Short Papers)}, pages 986--998.

\bibitem[{Song et~al.(2020)Song, Tan, Qin, Lu, and Liu}]{song2020mpnet}
Kaitao Song, Xu~Tan, Tao Qin, Jianfeng Lu, and Tie-Yan Liu. 2020.
\newblock Mpnet: Masked and permuted pre-training for language understanding.
\newblock \emph{Advances in neural information processing systems}, 33:16857--16867.

\bibitem[{Song et~al.(2023)Song, Khanuja, Liu, Faisal, Ostapenko, Winata, Aji, Cahyawijaya, Tsvetkov, Anastasopoulos et~al.}]{song2023globalbench}
Yueqi Song, Simran Khanuja, Pengfei Liu, Fahim Faisal, Alissa Ostapenko, Genta Winata, Alham Aji, Samuel Cahyawijaya, Yulia Tsvetkov, Antonios Anastasopoulos, et~al. 2023.
\newblock Globalbench: A benchmark for global progress in natural language processing.
\newblock In \emph{Proceedings of the 2023 Conference on Empirical Methods in Natural Language Processing}, pages 14157--14171.

\bibitem[{Srivastava et~al.(2023)Srivastava, Rastogi, Rao, Shoeb, Abid, Fisch, Brown, Santoro, Gupta, Garriga-Alonso et~al.}]{srivastava2023beyond}
Aarohi Srivastava, Abhinav Rastogi, Abhishek Rao, Abu Awal~Md Shoeb, Abubakar Abid, Adam Fisch, Adam~R Brown, Adam Santoro, Aditya Gupta, Adri{\`a} Garriga-Alonso, et~al. 2023.
\newblock Beyond the imitation game: Quantifying and extrapolating the capabilities of language models.
\newblock \emph{Transactions on Machine Learning Research}.

\bibitem[{Srivastava and Singh(2020)}]{srivastava2020phinc}
Vivek Srivastava and Mayank Singh. 2020.
\newblock Phinc: A parallel hinglish social media code-mixed corpus for machine translation.
\newblock In \emph{Proceedings of the Sixth Workshop on Noisy User-generated Text (W-NUT 2020)}, pages 41--49.

\bibitem[{Tanwar et~al.(2023)Tanwar, Dutta, Borthakur, and Chakraborty}]{tanwar2023multilingual}
Eshaan Tanwar, Subhabrata Dutta, Manish Borthakur, and Tanmoy Chakraborty. 2023.
\newblock Multilingual llms are better cross-lingual in-context learners with alignment.
\newblock In \emph{Proceedings of the 61st Annual Meeting of the Association for Computational Linguistics (Volume 1: Long Papers)}, pages 6292--6307.

\bibitem[{Team et~al.(2024)Team, Mesnard, Hardin, Dadashi, Bhupatiraju, Pathak, Sifre, Rivi{\`e}re, Kale, Love et~al.}]{team2024gemma}
Gemma Team, Thomas Mesnard, Cassidy Hardin, Robert Dadashi, Surya Bhupatiraju, Shreya Pathak, Laurent Sifre, Morgane Rivi{\`e}re, Mihir~Sanjay Kale, Juliette Love, et~al. 2024.
\newblock Gemma: Open models based on gemini research and technology.
\newblock \emph{arXiv preprint arXiv:2403.08295}.

\bibitem[{Thakur et~al.(2021)Thakur, Reimers, R{\"u}ckl{\'e}, Srivastava, and Gurevych}]{thakur2021beir}
Nandan Thakur, Nils Reimers, Andreas R{\"u}ckl{\'e}, Abhishek Srivastava, and Iryna Gurevych. 2021.
\newblock Beir: A heterogeneous benchmark for zero-shot evaluation of information retrieval models.
\newblock In \emph{Thirty-fifth Conference on Neural Information Processing Systems Datasets and Benchmarks Track (Round 2)}.

\bibitem[{Tiedemann(2020)}]{tiedemann2020tatoeba}
J{\"o}rg Tiedemann. 2020.
\newblock The tatoeba translation challenge--realistic data sets for low resource and multilingual mt.
\newblock In \emph{Proceedings of the Fifth Conference on Machine Translation}, pages 1174--1182.

\bibitem[{{\"U}st{\"u}n et~al.(2024){\"U}st{\"u}n, Aryabumi, Yong, Ko, D'souza, Onilude, Bhandari, Singh, Ooi, Kayid et~al.}]{ustun2024aya}
Ahmet {\"U}st{\"u}n, Viraat Aryabumi, Zheng-Xin Yong, Wei-Yin Ko, Daniel D'souza, Gbemileke Onilude, Neel Bhandari, Shivalika Singh, Hui-Lee Ooi, Amr Kayid, et~al. 2024.
\newblock Aya model: An instruction finetuned open-access multilingual language model.
\newblock \emph{arXiv preprint arXiv:2402.07827}.

\bibitem[{Van~der Maaten and Hinton(2008)}]{van2008visualizing}
Laurens Van~der Maaten and Geoffrey Hinton. 2008.
\newblock Visualizing data using t-sne.
\newblock \emph{Journal of machine learning research}, 9(11).

\bibitem[{Wang et~al.(2024)Wang, Yang, Huang, Yang, Majumder, and Wei}]{wang2024multilingual}
Liang Wang, Nan Yang, Xiaolong Huang, Linjun Yang, Rangan Majumder, and Furu Wei. 2024.
\newblock Multilingual e5 text embeddings: A technical report.
\newblock \emph{arXiv preprint arXiv:2402.05672}.

\bibitem[{Wang et~al.(2020)Wang, Wei, Dong, Bao, Yang, and Zhou}]{wang2020minilm}
Wenhui Wang, Furu Wei, Li~Dong, Hangbo Bao, Nan Yang, and Ming Zhou. 2020.
\newblock Minilm: Deep self-attention distillation for task-agnostic compression of pre-trained transformers.
\newblock \emph{Advances in Neural Information Processing Systems}, 33:5776--5788.

\bibitem[{Wang et~al.(2023)Wang, Macdonald, Tonellotto, and Ounis}]{wang2023colbert}
Xiao Wang, Craig Macdonald, Nicola Tonellotto, and Iadh Ounis. 2023.
\newblock Colbert-prf: Semantic pseudo-relevance feedback for dense passage and document retrieval.
\newblock \emph{ACM Transactions on the Web}, 17(1):1--39.

\bibitem[{Winata et~al.(2022)Winata, Wu, Kulkarni, Solorio, and Preo{\c{t}}iuc-Pietro}]{winata2022cross}
Genta Winata, Shijie Wu, Mayank Kulkarni, Thamar Solorio, and Daniel Preo{\c{t}}iuc-Pietro. 2022.
\newblock Cross-lingual few-shot learning on unseen languages.
\newblock In \emph{Proceedings of the 2nd Conference of the Asia-Pacific Chapter of the Association for Computational Linguistics and the 12th International Joint Conference on Natural Language Processing (Volume 1: Long Papers)}, pages 777--791.

\bibitem[{Winata et~al.(2023{\natexlab{a}})Winata, Xie, Radhakrishnan, Gao, and Preo{\c{t}}iuc-Pietro}]{winata2023efficient}
Genta Winata, Lingjue Xie, Karthik Radhakrishnan, Yifan Gao, and Daniel Preo{\c{t}}iuc-Pietro. 2023{\natexlab{a}}.
\newblock Efficient zero-shot cross-lingual inference via retrieval.
\newblock In \emph{Proceedings of the 13th International Joint Conference on Natural Language Processing and the 3rd Conference of the Asia-Pacific Chapter of the Association for Computational Linguistics (Volume 2: Short Papers)}, pages 93--104.

\bibitem[{Winata et~al.(2023{\natexlab{b}})Winata, Aji, Cahyawijaya, Mahendra, Koto, Romadhony, Kurniawan, Moeljadi, Prasojo, Fung et~al.}]{winata2023nusax}
Genta~Indra Winata, Alham~Fikri Aji, Samuel Cahyawijaya, Rahmad Mahendra, Fajri Koto, Ade Romadhony, Kemal Kurniawan, David Moeljadi, Radityo~Eko Prasojo, Pascale Fung, et~al. 2023{\natexlab{b}}.
\newblock Nusax: Multilingual parallel sentiment dataset for 10 indonesian local languages.
\newblock In \emph{Proceedings of the 17th Conference of the European Chapter of the Association for Computational Linguistics}, pages 815--834.

\bibitem[{Winata et~al.(2021{\natexlab{a}})Winata, Cahyawijaya, Liu, Lin, Madotto, and Fung}]{winata2021multilingual}
Genta~Indra Winata, Samuel Cahyawijaya, Zihan Liu, Zhaojiang Lin, Andrea Madotto, and Pascale Fung. 2021{\natexlab{a}}.
\newblock Are multilingual models effective in code-switching?
\newblock \emph{NAACL 2021}, page 142.

\bibitem[{Winata et~al.(2021{\natexlab{b}})Winata, Madotto, Lin, Liu, Yosinski, and Fung}]{winata2021language}
Genta~Indra Winata, Andrea Madotto, Zhaojiang Lin, Rosanne Liu, Jason Yosinski, and Pascale Fung. 2021{\natexlab{b}}.
\newblock Language models are few-shot multilingual learners.
\newblock In \emph{Proceedings of the 1st Workshop on Multilingual Representation Learning}, pages 1--15.

\bibitem[{Xue et~al.(2021)Xue, Constant, Roberts, Kale, Al-Rfou, Siddhant, Barua, and Raffel}]{xue2021mt5}
Linting Xue, Noah Constant, Adam Roberts, Mihir Kale, Rami Al-Rfou, Aditya Siddhant, Aditya Barua, and Colin Raffel. 2021.
\newblock mt5: A massively multilingual pre-trained text-to-text transformer.
\newblock In \emph{Proceedings of the 2021 Conference of the North American Chapter of the Association for Computational Linguistics: Human Language Technologies}, pages 483--498.

\bibitem[{Yang et~al.(2019{\natexlab{a}})Yang, Zhang, and Lin}]{yang2019simple}
Wei Yang, Haotian Zhang, and Jimmy Lin. 2019{\natexlab{a}}.
\newblock Simple applications of bert for ad hoc document retrieval.
\newblock \emph{arXiv preprint arXiv:1903.10972}.

\bibitem[{Yang et~al.(2019{\natexlab{b}})Yang, Abrego, Yuan, Guo, Shen, Cer, Sung, Strope, and Kurzweil}]{yang2019improving}
Yinfei Yang, Gustavo~Hernandez Abrego, Steve Yuan, Mandy Guo, Qinlan Shen, Daniel Cer, Yun-Hsuan Sung, Brian Strope, and Ray Kurzweil. 2019{\natexlab{b}}.
\newblock Improving multilingual sentence embedding using bi-directional dual encoder with additive margin softmax.
\newblock \emph{arXiv preprint arXiv:1902.08564}.

\bibitem[{Yong et~al.(2023)Yong, Schoelkopf, Muennighoff, Aji, Adelani, Almubarak, Bari, Sutawika, Kasai, Baruwa et~al.}]{yong2023bloom+}
Zheng~Xin Yong, Hailey Schoelkopf, Niklas Muennighoff, Alham~Fikri Aji, David~Ifeoluwa Adelani, Khalid Almubarak, M~Saiful Bari, Lintang Sutawika, Jungo Kasai, Ahmed Baruwa, et~al. 2023.
\newblock Bloom+ 1: Adding language support to bloom for zero-shot prompting.
\newblock In \emph{Proceedings of the 61st Annual Meeting of the Association for Computational Linguistics (Volume 1: Long Papers)}, pages 11682--11703.

\bibitem[{Zhang et~al.(2023)Zhang, Cahyawijaya, Cruz, Winata, and Aji}]{zhang2023multilingual}
Ruochen Zhang, Samuel Cahyawijaya, Jan Christian~Blaise Cruz, Genta Winata, and Alham Aji. 2023.
\newblock Multilingual large language models are not (yet) code-switchers.
\newblock In \emph{Proceedings of the 2023 Conference on Empirical Methods in Natural Language Processing}, pages 12567--12582.

\bibitem[{Zweigenbaum et~al.(2017)Zweigenbaum, Sharoff, and Rapp}]{zweigenbaum2017overview}
Pierre Zweigenbaum, Serge Sharoff, and Reinhard Rapp. 2017.
\newblock Overview of the second bucc shared task: Spotting parallel sentences in comparable corpora.
\newblock In \emph{Proceedings of the 10th Workshop on Building and Using Comparable Corpora}, pages 60--67.

\bibitem[{Zweigenbaum et~al.(2018)Zweigenbaum, Sharoff, and Rapp}]{zweigenbaum2018overview}
Pierre Zweigenbaum, Serge Sharoff, and Reinhard Rapp. 2018.
\newblock Overview of the third bucc shared task: Spotting parallel sentences in comparable corpora.
\newblock In \emph{Proceedings of 11th workshop on building and using comparable corpora}, pages 39--42.

\end{thebibliography}
\bibliographystyle{acl_natbib}

\appendix      

\section{Experimental Details}
\subsection{Baselines}
For the task-specific evaluation, we include the following baseline models for comparison:

\paragraph{SOTA} We report the state-of-the-art (SOTA) from the existing literature as follows:
\begin{itemize}
    \item \textbf{Bitext Retrieval:} BUCC~\cite{wang2024multilingual} and Tatoeba~\cite{wang2024multilingual}.
    \item \textbf{Classification:} MASSIVE~\cite{fitzgerald2023massive}, NollySenti~\cite{shode2023nollysenti}, NusaX~\citep[monolingual]{winata2023nusax}~\citep[cross-lingual]{winata2023efficient}, and SIB-200~\cite{adelani2023sib}. We use the validation split on Accuracy for LinCE SA, but to the best of our knowledge, there is no comparable result in the literature. We make a small modification to FIRE 2020 labels, thus there are no comparable results in the literature.
\end{itemize}

\paragraph{Classification Baselines}
We report the following baselines for classification tasks:
\begin{itemize}
    \item \textbf{Random:} In this baseline, prediction labels are sampled randomly from a uniform distribution. This approach ensures that each label has an equal probability of being selected, regardless of its true distribution within the dataset. It serves as a baseline to compare the effectiveness of more sophisticated methods.
    \item \textbf{Majority:} In this baseline, prediction labels are selected by taking the majority class for all instances. By always predicting the most frequent class observed in the training data, this method provides a simple yet effective baseline, especially in datasets with class imbalance. It helps to highlight the performance of models in recognizing and classifying less frequent classes.
    \item \textbf{Fine-tune (XLM-R$_\text{BASE}$):} We fine-tune a XLM-R$_\text{BASE}$ model using the training split of the dataset. After fine-tuning, the model is evaluated on the test data split of the same dataset to assess its performance.
\end{itemize}

\begin{figure}[!t]
    \centering
    \begin{subfigure}[t]{0.235\textwidth}
        \centering
        \includegraphics[width=\linewidth]{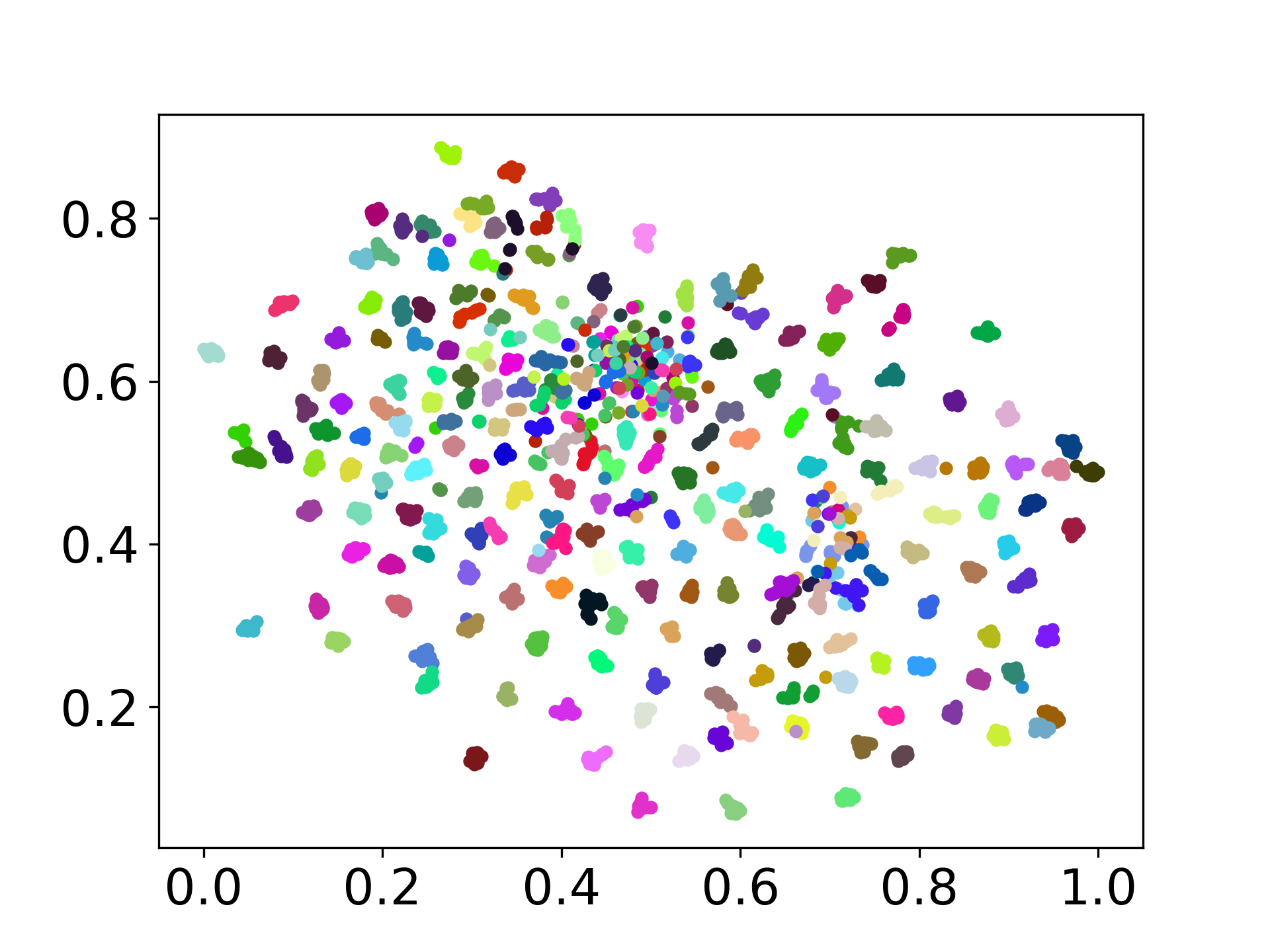} 
        \caption{LaBSE (sample ID)}
    \end{subfigure}
    \begin{subfigure}[t]{0.235\textwidth}
        \centering
        \includegraphics[width=\linewidth]{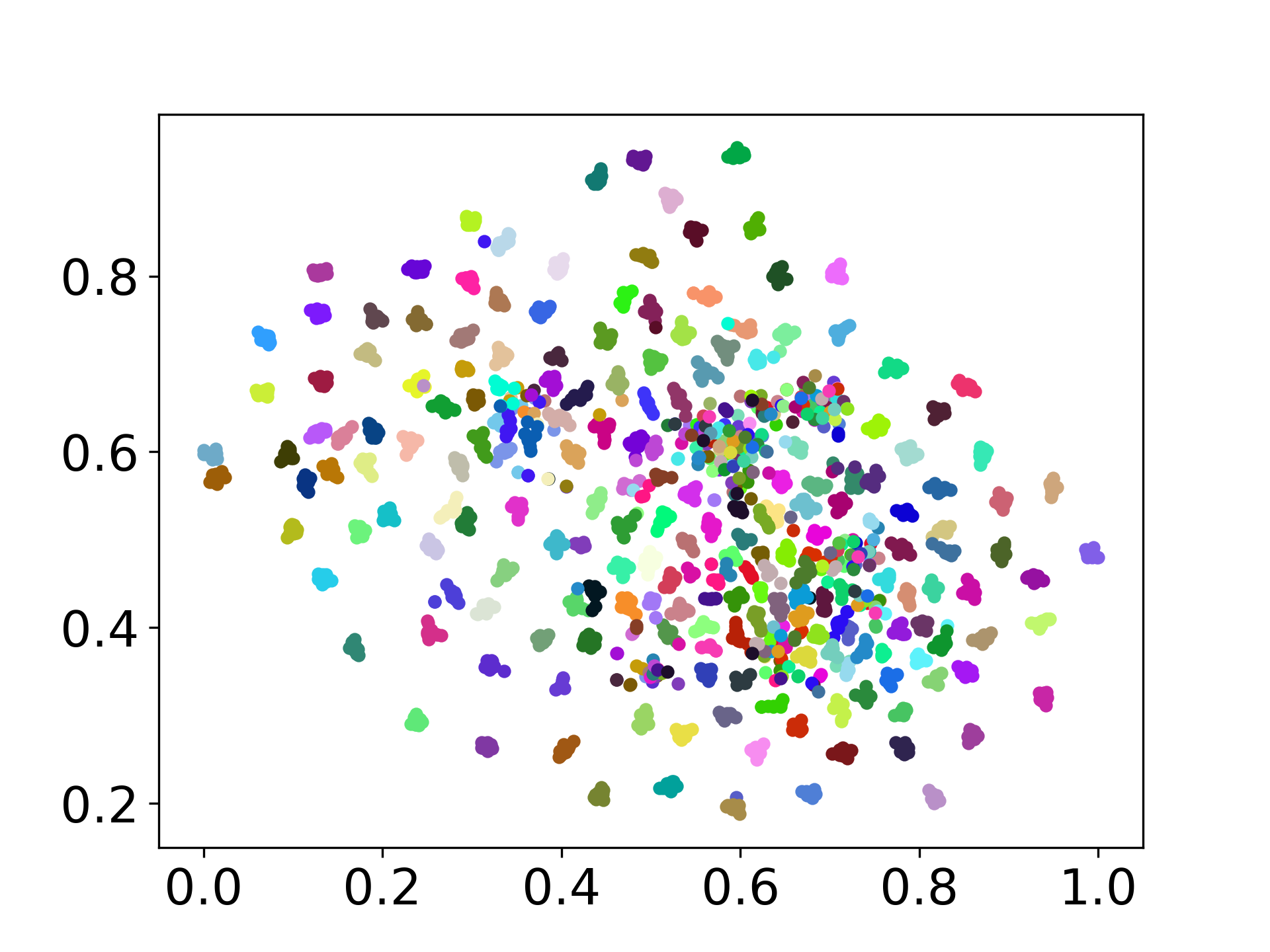} 
        \caption{Cohere-Embedv3 (sample ID)}
    \end{subfigure}
    \begin{subfigure}[t]{0.235\textwidth}
        \centering
        \includegraphics[width=\linewidth]{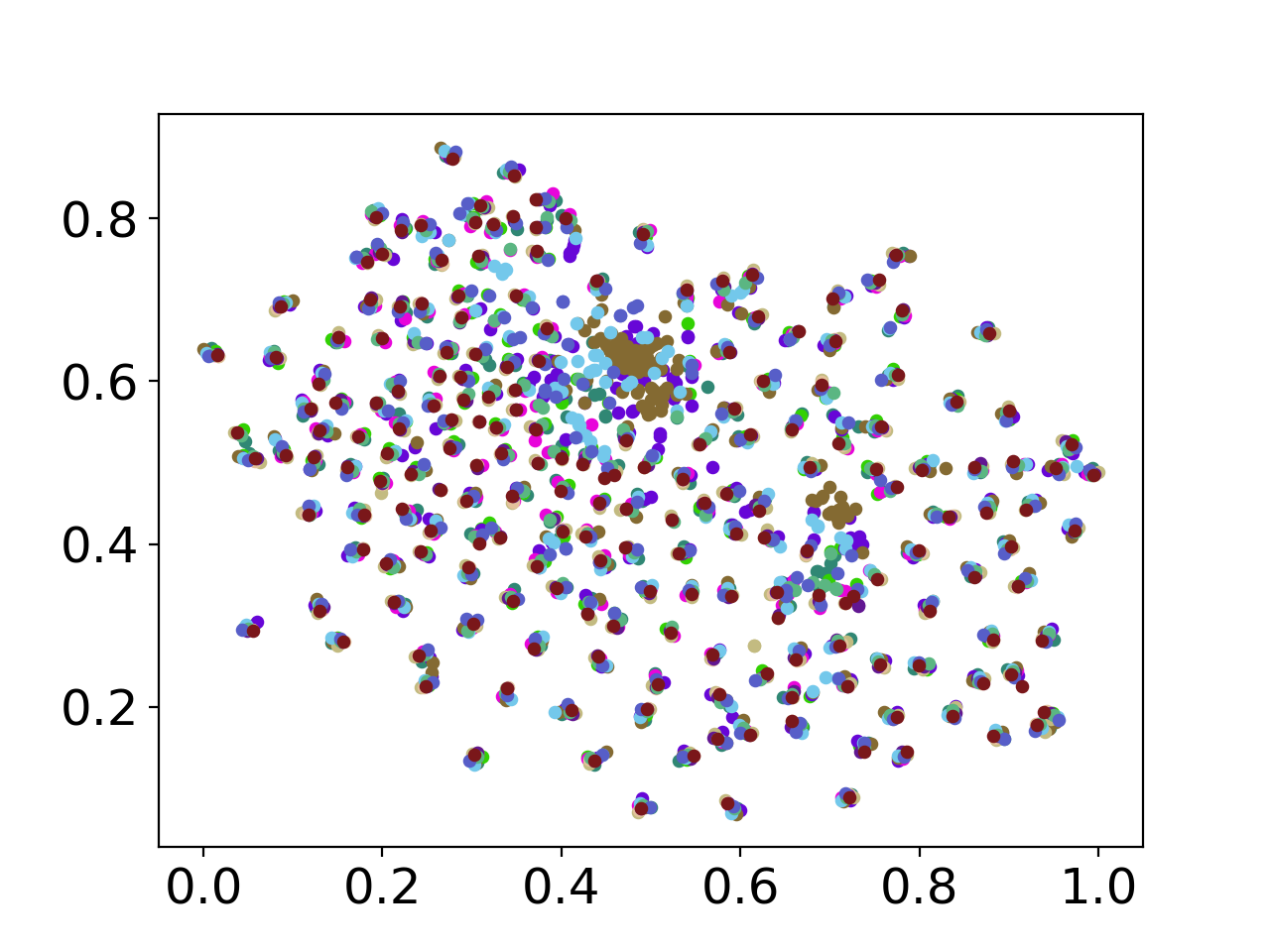} 
        \caption{LaBSE (language)}
    \end{subfigure}
    \begin{subfigure}[t]{0.235\textwidth}
        \centering
        \includegraphics[width=\linewidth]{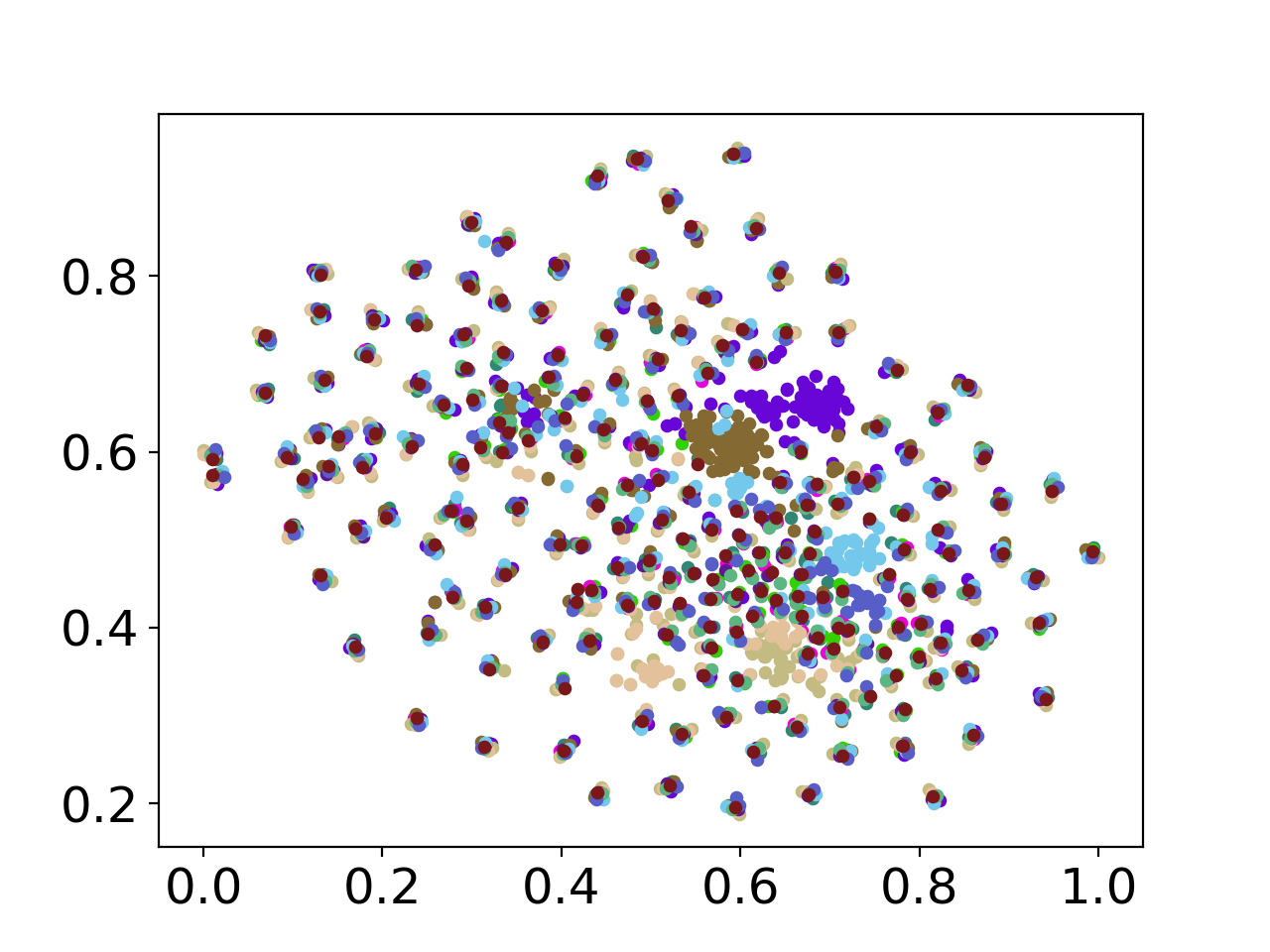} 
        \caption{Cohere-Embedv3 (language)}
    \end{subfigure}
    \begin{subfigure}[t]{0.235\textwidth}
        \centering
        \includegraphics[width=\linewidth]{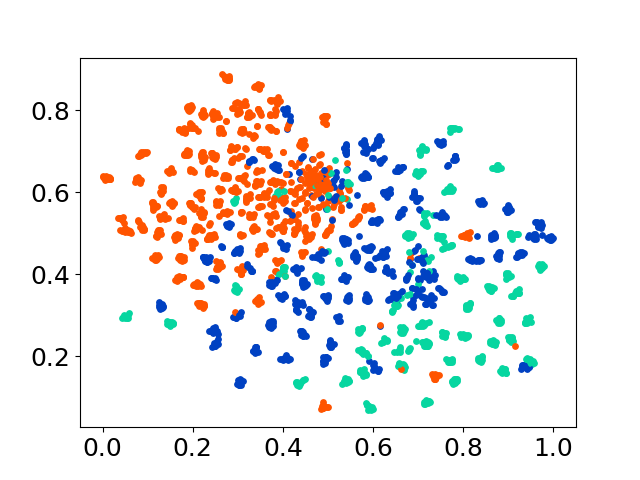} 
        \caption{LaBSE (class label)}
    \end{subfigure}
    \begin{subfigure}[t]{0.235\textwidth}
        \centering
        \includegraphics[width=\linewidth]{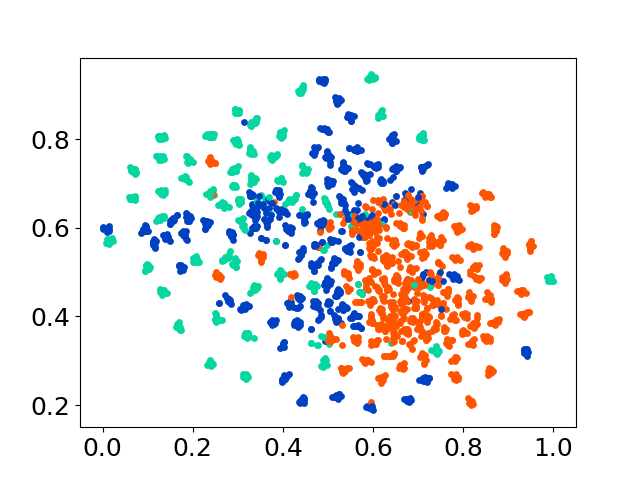} 
        \caption{Cohere-Embedv3 (class label)}
    \end{subfigure}
    \caption{t-SNE representation of 200 random samples from the NusaX dataset. The color on the figures show the sample ID for \textbf{(a)} and \textbf{(b)}, language for \textbf{(c)} and \textbf{(d)}, and class for \textbf{(e)} and \textbf{(f)}.}
    \label{tsne-e5-cohere}
\end{figure}

\subsection{Datasets}
\subsubsection{Preprocessing}
\label{sec:dataset-preprocessing}
To enhance the data clarity for LMs and improve their predictive performance, we apply preprocessing steps to the following two datasets:
\begin{itemize}
    \item \textbf{FIRE 2020}: We modify several non-standard labels to a single label for sentiment analysis. We map \texttt{``Mixed\_feeling"} into \texttt{``Mixed"}, and map \texttt{``not-malayalam"}, \texttt{``non-tamil"}, and \texttt{``unknown\_state"} into \texttt{``Unknown"}.
    \item \textbf{MASSIVE}: We replace the underscore character with a space character from the labels.
\end{itemize}
\subsubsection{Statistics}
Table~\ref{dataset-statistics} displays the dataset statistics for each dataset split. In the case of NollySenti in bitext-retrieval, the English data predominates over other languages, prompting us to consider an equal number of data points for all languages. As for LinCE SA, since we did not utilize the test set, the statistics for this particular dataset are not reported.

\begin{table*}[!t]
\centering
\resizebox{\textwidth}{!}{
    \begin{tabular}{lllllll}
    \toprule
    \textbf{Dataset} & \textbf{lang} & \textbf{\# Train} & \textbf{\# Valid} & \textbf{\# Test} & \textbf{Source} & \textbf{License} \\ \midrule 
    BUCC & all & N/A & N/A & 35k & \url{https://huggingface.co/datasets/mteb/bucc-bitext-mining} & CC-BY-SA \\
    MASSIVE & all & 587k & 104k & 152k & \url{https://huggingface.co/datasets/AmazonScience/massive} & CC-BY 4.0 \\
    NollySenti & en & 1,302 & 100  & 500 & \url{https://github.com/IyanuSh/NollySenti/tree/main} & CC-BY 4.0\\
    & yo & 900 & 100 & 500 \\
    & ha/ig/pcm & 410 & 100 & 500 \\
    NusaX & each lang. & 500 & 100 & 400 & \url{https://huggingface.co/datasets/indonlp/NusaX-senti/viewer/eng/train} & CC-BY-SA 4.0 \\
    NusaT & btk/bew/jav/ & 6.6k  & 849 & 2k & \url{https://huggingface.co/datasets/indonlp/nusatranslation_mt} & Apache 2.0 \\
    & mad/mak/min/sun & 6.6k  & 849 & 2k  \\ 
    & abs/bhp/mui/rej  & 1k & 174 & 400 \\ \midrule
    \multicolumn{5}{l}{Code-switching} \\ \midrule
    FIRE 2020 & malayalam & 4,851 & 541 & 1,348 & \url{https://dravidian-codemix.github.io/2020/} & N/A \\
    & tamil & 11,335 & 1,260 & 3,149 \\
    LinCE MT & eng-hinglish & 8,060 & 942 & N/A & \url{https://ritual.uh.edu/lince/} & Research Only \\
    LinCE SA & eng-spa & 12,002 & 2,998 & N/A & \url{https://huggingface.co/datasets/lince-benchmark/lince} & Research Only \\
    PHINC & N/A & N/A & 27,477 & & \url{https://huggingface.co/datasets/veezbo/phinc} & CC-BY 4.0
    \\ \bottomrule
    \end{tabular}
}
\caption{Dataset statistics.} 
\label{dataset-statistics}
\end{table*}

\subsection{Languages Under Study}
Table~\ref{cross-lingual-settings} presents a comprehensive list of source and target language pairs used in our cross-lingual experiments. The datasets apply different language code standards. To ensure consistency and uphold the integrity of the original datasets, we have reported the language codes exactly as they appear in the respective sources.

\begin{table*}[!ht]
\centering
\resizebox{\textwidth}{!}{
    \begin{tabular}{lll}
    \toprule
    \textbf{Dataset} & \textbf{Source Language} & \textbf{Target Language(s)}
    \\ \midrule
    BUCC & en & de, fr, zh\\ \midrule
    FIRE 2020 & tamil & malayalam \\ \midrule
    MASSIVE & en & af, am, ar, az, bn, cy, da, de, el, es, fa, fi, fr, he, hi, hu, hy, id, is, it, ja, jv, ka, km, kn, ko, lv, ml, mn, ms, my, nb, nl,\\
    & & pl, pt, ro, ru, sl, sq, sv, sw, ta, te, th, tl, tr, ur, vi, zh-CN, zh-TW\\ \midrule
    NollySenti & en & ha, ig, pcm, yo\\ \midrule
    NusaX & eng & ace, ban, bbc, bjn, bug, ind, jav, mad, min, nij, sun \\ \midrule
    SIB-200 & eng\_Latn & ace\_Arab, ace\_Latn, acm\_Arab, acq\_Arab, aeb\_Arab, afr\_Latn, ajp\_Arab, aka\_Latn, als\_Latn, amh\_Ethi, \\ 
    & & apc\_Arab, arb\_Arab, arb\_Latn, ars\_Arab, ary\_Arab, arz\_Arab, asm\_Beng, ast\_Latn, awa\_Deva, ayr\_Latn, \\ 
    & & azb\_Arab, azj\_Latn, bak\_Cyrl, bam\_Latn, ban\_Latn, bel\_Cyrl, bem\_Latn, ben\_Beng, bho\_Deva, bjn\_Arab, \\
    & & bjn\_Latn, bod\_Tibt, bos\_Latn, bug\_Latn, bul\_Cyrl, cat\_Latn, ceb\_Latn, ces\_Latn, cjk\_Latn, ckb\_Arab, crh\_Latn, \\
    & & cym\_Latn, dan\_Latn, deu\_Latn, dik\_Latn, dyu\_Latn, 
    dzo\_Tibt, ell\_Grek, epo\_Latn, est\_Latn, eus\_Latn, ewe\_Latn, \\
    & & fao\_Latn, fij\_Latn, fin\_Latn, fon\_Latn, 
    fra\_Latn, fur\_Latn,
    fuv\_Latn, gaz\_Latn, gla\_Latn, gle\_Latn, glg\_Latn, \\
    & & grn\_Latn, guj\_Gujr, hat\_Latn, hau\_Latn, heb\_Hebr, hin\_Deva, hne\_Deva, hrv\_Latn, hun\_Latn, hye\_Armn, \\
    & & ibo\_Latn, ilo\_Latn, ind\_Latn, isl\_Latn, ita\_Latn, jav\_Latn, jpn\_Jpan, kab\_Latn, kac\_Latn, kam\_Latn, kan\_Knda, \\
    & & kas\_Arab, kas\_Deva, kat\_Geor, kaz\_Cyrl, kbp\_Latn, kea\_Latn, khk\_Cyrl, khm\_Khmr, 
    kik\_Latn, kin\_Latn, kir\_Cyrl, \\
    & & kmb\_Latn, kmr\_Latn, knc\_Arab, knc\_Latn, kon\_Latn, kor\_Hang, lao\_Laoo, lij\_Latn, lim\_Latn, lin\_Latn, lit\_Latn, \\
    & & lmo\_Latn, ltg\_Latn, ltz\_Latn, lua\_Latn, lug\_Latn, luo\_Latn, 
    lus\_Latn, lvs\_Latn, mag\_Deva, mai\_Deva, mal\_Mlym, \\
    & & mar\_Deva, min\_Arab, min\_Latn, mkd\_Cyrl, mlt\_Latn, mni\_Beng, mos\_Latn, mri\_Latn, 
    mya\_Mymr, nld\_Latn, \\
    & & nno\_Latn, nob\_Latn, npi\_Deva, nqo\_Nkoo, nso\_Latn, nus\_Latn, nya\_Latn, oci\_Latn, ory\_Orya, pag\_Latn, \\
    & & pan\_Guru, pap\_Latn, pbt\_Arab, pes\_Arab, plt\_Latn, pol\_Latn, por\_Latn, prs\_Arab, quy\_Latn, ron\_Latn, \\
    & & run\_Latn, rus\_Cyrl, sag\_Latn, san\_Deva, sat\_Olck, scn\_Latn, shn\_Mymr, sin\_Sinh, slk\_Latn, slv\_Latn, \\
    & & smo\_Latn, sna\_Latn, snd\_Arab, som\_Latn, sot\_Latn, spa\_Latn, srd\_Latn, srp\_Cyrl, ssw\_Latn, sun\_Latn, \\
    & & swe\_Latn, swh\_Latn, szl\_Latn, tam\_Taml, taq\_Latn, taq\_Tfng, tat\_Cyrl, tel\_Telu, tgk\_Cyrl, tgl\_Latn, tha\_Thai, \\
    & & tir\_Ethi, tpi\_Latn, tsn\_Latn, tso\_Latn, tuk\_Latn, tum\_Latn, tur\_Latn, twi\_Latn, tzm\_Tfng, uig\_Arab, \\
    & & ukr\_Cyrl, umb\_Latn, urd\_Arab, uzn\_Latn, vec\_Latn, vie\_Latn, war\_Latn, wol\_Latn, xho\_Latn, ydd\_Hebr, \\
    & & yor\_Latn, yue\_Hant, zho\_Hans, zho\_Hant, zsm\_Latn, zul\_Latn \\ \midrule
    Tatoeba & eng & afr, amh, ang, ara, arq, arz, ast, awa, aze, bel, ben, ber, bos, bre, bul, cat, cbk, ceb, ces, cha, cmn, cor, csb, cym, dan, \\
    & & deu, dsb, dtp, ell, epo, est, eus, fao, fin, fra, fry, gla, gle, glg, gsw, heb, hin, hrv, hsb, hun, hye, ido, ile, ina, ind, isl, ita, \\
    & & jav, jpn, kab, kat, kaz, khm, kor, kur, kzj, lat, lfn, lit, lvs, mal, mar, max, mhr, mkd, mon, nds, nld, nno, nob, nov, oci, \\
    & & orv, pam, pes, pms, pol, por, ron, rus, slk, slv, spa, sqi, srp, swe, swg, swh, tam, tat, tel, tgl, tha, tuk, tur, tzl, uig, ukr, \\
    & & urd, uzb, vie, war, wuu, xho, yid, yue, zsm
    \\ \bottomrule
    \end{tabular}
}
\caption{List of source and target languages for all datasets in the cross-lingual setting. Each dataset employs a different language code standard, and we have reported them as used.} 
\label{cross-lingual-settings}
\end{table*}

\begin{table}[!ht]
\centering
\resizebox{0.49\textwidth}{!}{
    \begin{tabular}{ll}
    \toprule
    \textbf{Model} & \textbf{Hugging Face Model}
    \\ \midrule
    LaBSE & sentence-transformers/LaBSE \\
    CMLM & sentence-transformers/use-cmlm-multilingual\\
    E5$_\text{BASE}$ & intfloat/multilingual-e5-base \\
    E5$_\text{LARGE}$ & intfloat/multilingual-e5-large \\
    MPNet$_\text{BASE}$v2 & sentence-transformers/paraphrase-multilingual-mpnet-base-v2 \\
    MiniLM$_\text{L12-E384}$ & microsoft/Multilingual-MiniLM-L12-H384 \\
    Glot-500 & cis-lmu/glot500-base \\
    XLM-R$_\text{BASE}$ & FacebookAI/xlm-roberta-base\\
    XLM-R$_\text{LARGE}$ & FacebookAI/xlm-roberta-large \\
    Aya-23 8B & CohereForAI/aya-23-8B \\
    Llama 3 8B Instruct & meta-llama/Meta-Llama-3-8B-Instruct \\
    Llama 3.1 8B Instruct & meta-llama/Meta-Llama-3.1-8B-Instruct \\
    BLOOMZ 560m & bigscience/bloomz-560m \\
    BLOOMZ 1.7B & bigscience/bloomz-17b\\
    BLOOMZ 3B & bigscience/bloomz-3b \\
    mT0 3B & bigscience/mt0-xl \\
    XGLM 564m & facebook/xglm-564M\\
    XGLM 2.9B & facebook/xglm-2.9B
    \\ \bottomrule
    \end{tabular}
}
\caption{Hugging Face models.} 
\label{hf-models}
\end{table}

\begin{table*}[!t]
\centering
\resizebox{.95\textwidth}{!}{
    \begin{tabular}{lcccccc}
    \toprule
    \textbf{Parameter} & \textbf{NusaX} & \textbf{SIB-200} & \textbf{MASSIVE} & \textbf{LinCE SA} & \textbf{NollySenti} & \textbf{FIRE 2020} \\ \midrule 
    batch size & 32 & 8 & 32 & 16 & 16 & 16 \\
    learning rate & 1e-5 & 1e-5 & 1e-5 & 5e-5 & 5e-5 & 1e-5 \\
    max epoch & 100 & 100 & 100 & 20 & 20 & 100 \\
    early stopping & 3 & 5 & 3 & 5 & 5 & 5 \\
    adam beta 1 & 0.9 & 0.9 & 0.9 & 0.9 & 0.9 & 0.9 \\
    adam beta 2 & 0.999& 0.999& 0.999& 0.999& 0.999& 0.999\\
    adam epsilon & 1e-8& 1e-8& 1e-8& 1e-8& 1e-8& 1e-8
    \\ \bottomrule
    \end{tabular}
}
\caption{Hyper-parameters for fine-tuning baselines.} 
\label{hyper-parameters-finetuning}
\end{table*}

\subsection{LM Sources}
We extensively utilize a range of open-source encoder and generative LMs from the Hugging Face repository to ensure our evaluations are comprehensive and transparent. The models we employ are detailed in Table~\ref{hf-models}, showcasing the diversity in architectures and training objectives. These open-source models provide a solid foundation for our evaluations, allowing us to benchmark against widely accepted standards in the NLP community. For commercial models, we leverage state-of-the-art APIs to access robust and high-performance LMs. Specifically, we use the OpenAI API to retrieve generation responses from GPT-3.5 Turbo and GPT-4. Additionally, we utilize Cohere's Embed API to incorporate the Cohere-Embedv3 model.

\subsection{LM Inference}
\label{lm-inference}
We run our model inference on an A100 40G GPU, utilizing 8-bit quantization~\cite{dettmers2022gpt3} to optimize memory usage and speed up inference. Our experiments investigate the impact of varying the number of retrieved samples $k \in [1, 5, 10]$ to understand how retrieval quality and classification performance change with the number of instances. These samples are used for both bitext retrieval and retrieval-based classification tasks. For the ICL classification task, we evaluate our model in both zero-shot and one-shot scenarios using two methods: (1) predicting the label distribution by computing the next token probability, and (2) generating the response directly. For BLOOMZ, Aya, and XGLM models, we use the first method since we have access to the next token prediction logits. For Llama 3, Gemma, and mT0 models, obtaining these logits is less straightforward. Specifically, the presence of numerous special tokens in Llama 3 complicates logit calculation, so we opt for the second method, which leverages the model's strong capability to generate exact labels by following instructions. Similarly, for GPT-3.5 Turbo and GPT-4o models, we adopt the second method because we do not have direct access to the logits for all possible classes. These models excel in instruction following, making direct response generation a practical and effective approach.

\subsection{Hyper-parameters}
To ensure fair and consistent evaluations across our models, we employ a set of specific hyper-parameters during the inference stage, as detailed in Table~\ref{hyper-parameters-api}. These hyper-parameters have been carefully chosen to standardize the evaluation process and ensure that our comparisons are both meaningful and reliable. For our fine-tuning baselines, we adopt a different set of hyper-parameters, which are listed comprehensively in Table~\ref{hyper-parameters-finetuning}. These parameters are optimized to enhance the model's performance during the fine-tuning phase. Moreover, to streamline the fine-tuning process, we have decided not to incorporate any warmup steps. The linear scheduler has been chosen for its simplicity and effectiveness.

\section{Extended Analysis}

\subsection{LM Representation Visualization}
\label{sec:more-visualization}
In Figure~\ref{tsne-e5-cohere}, we present the t-SNE 2D visualization of a subset of 200 randomly selected samples from the NusaX dataset. The visualization showcases how the LaBSE and Cohere-Embedv3 models effectively align samples originating from various languages in a meaningful and interpretable manner. Notably, both models exhibit a high level of proficiency in grouping the samples based on their class labels, indicating robust performance in semantic alignment tasks. This finding is consistent with the behavior observed in models that have been trained using contrastive learning methods, such as the E5 models. The ability of these models to accurately capture semantic relationships across multilingual data highlights their effectiveness in handling diverse linguistic contexts and tasks.

\begin{figure*}[!th]
    \centering
    \begin{subfigure}[t]{0.37\textwidth}
        \centering
        \caption{$k=1$}
        \includegraphics[width=\linewidth]{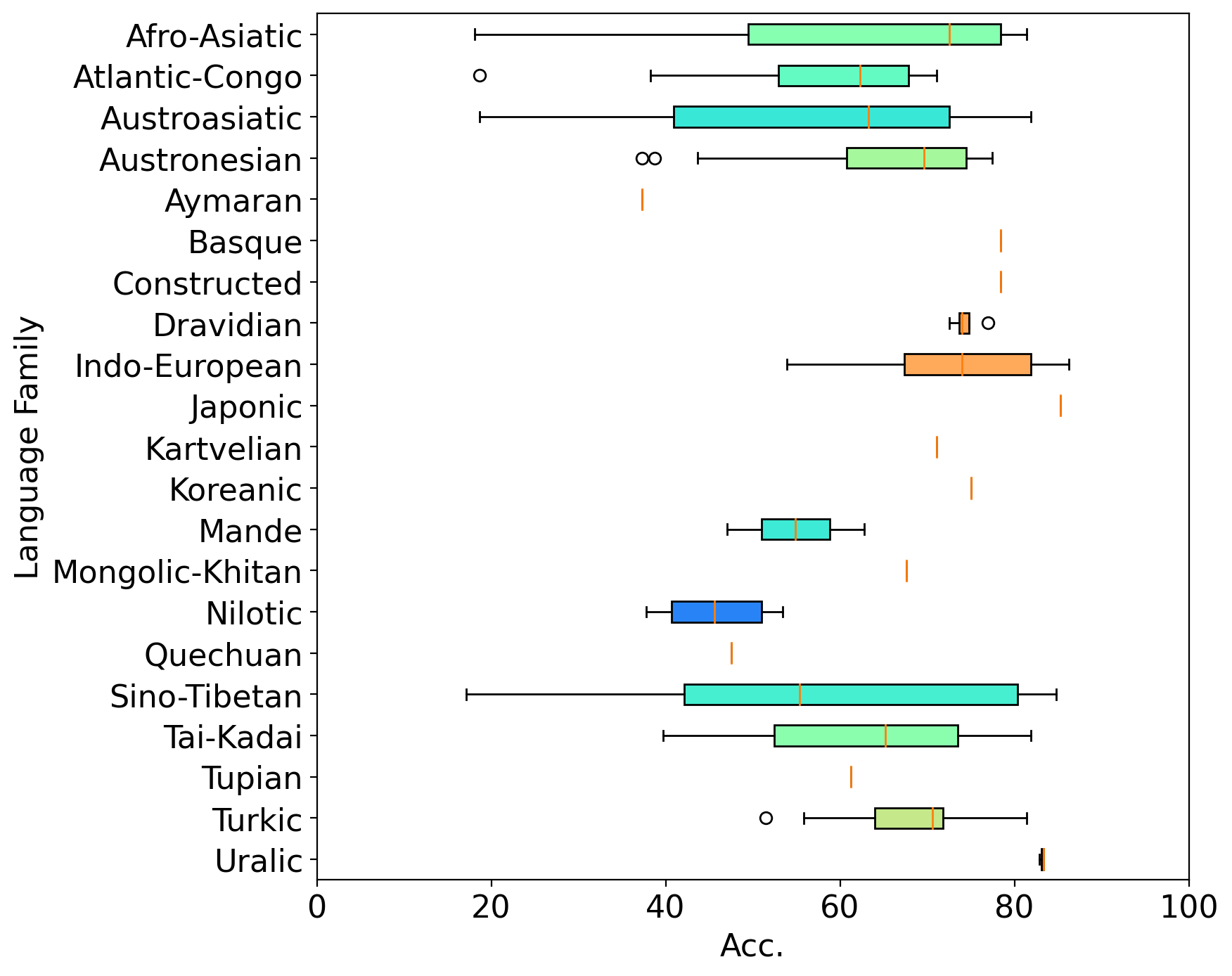} 
        
    \end{subfigure}
    \begin{subfigure}[t]{0.283\textwidth}
        \centering
        \caption{$k=5$}
        \includegraphics[width=1.0\linewidth]{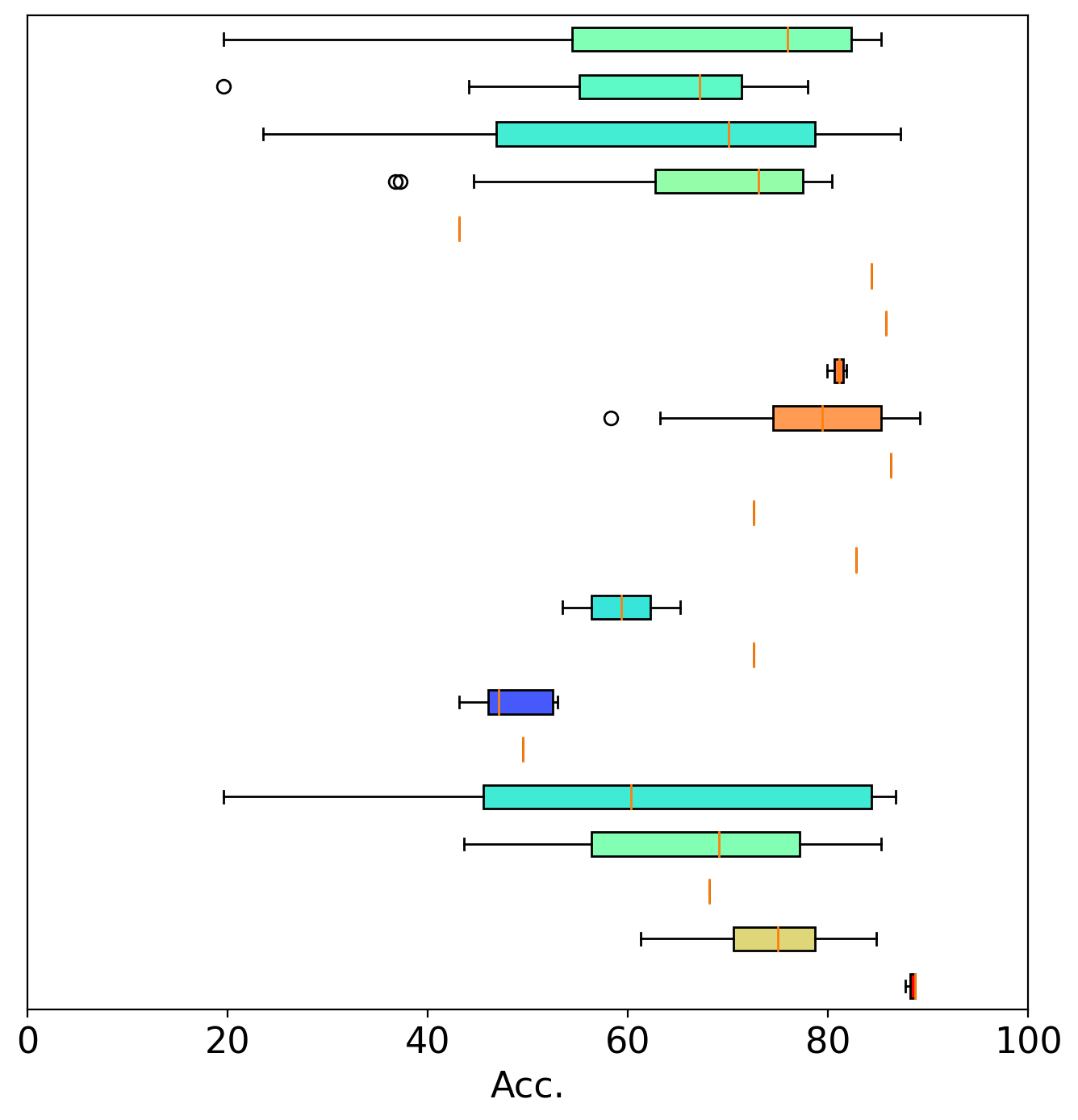} 
        
    \end{subfigure}
    \begin{subfigure}[t]
    {0.283\textwidth}
        \centering
        \caption{$k=10$}
        \includegraphics[width=1.0\linewidth]{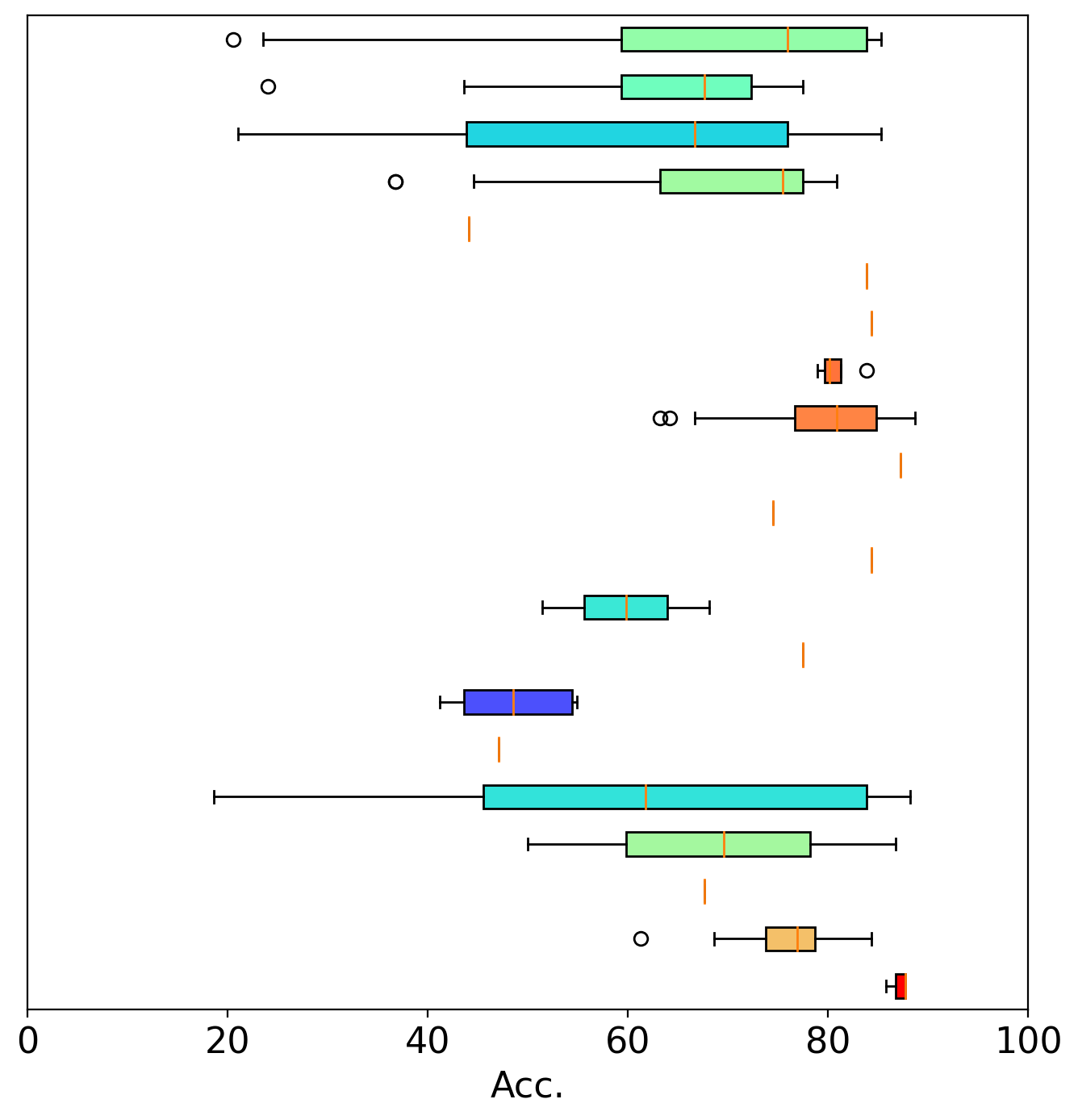} 
        
    \end{subfigure}
    \caption{Results for the retrieval-based classification task on the SIB-200 dataset, using $k$ values of [1, 5, 10], across various language families.}
    \label{classification-sib200-results}
\end{figure*} 

\begin{figure*}[!th]
    \centering
    \begin{subfigure}[t]{0.357\textwidth}
        \centering
        \caption{$k=1$}
        \includegraphics[width=\linewidth]{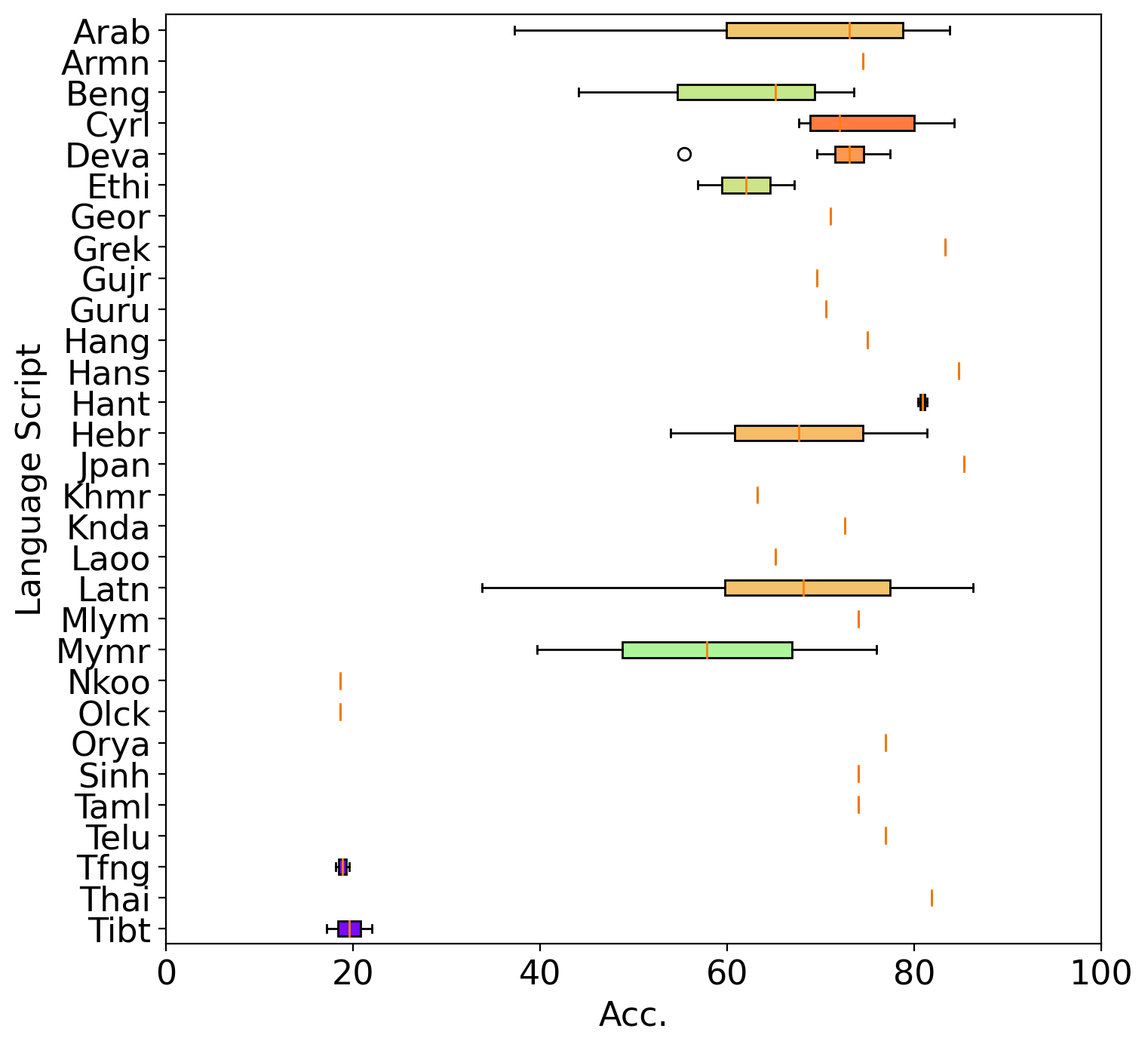} 
        
    \end{subfigure}
    \begin{subfigure}[t]{0.313\textwidth}
        \centering
        \caption{$k=5$}
        \includegraphics[width=1.0\linewidth]{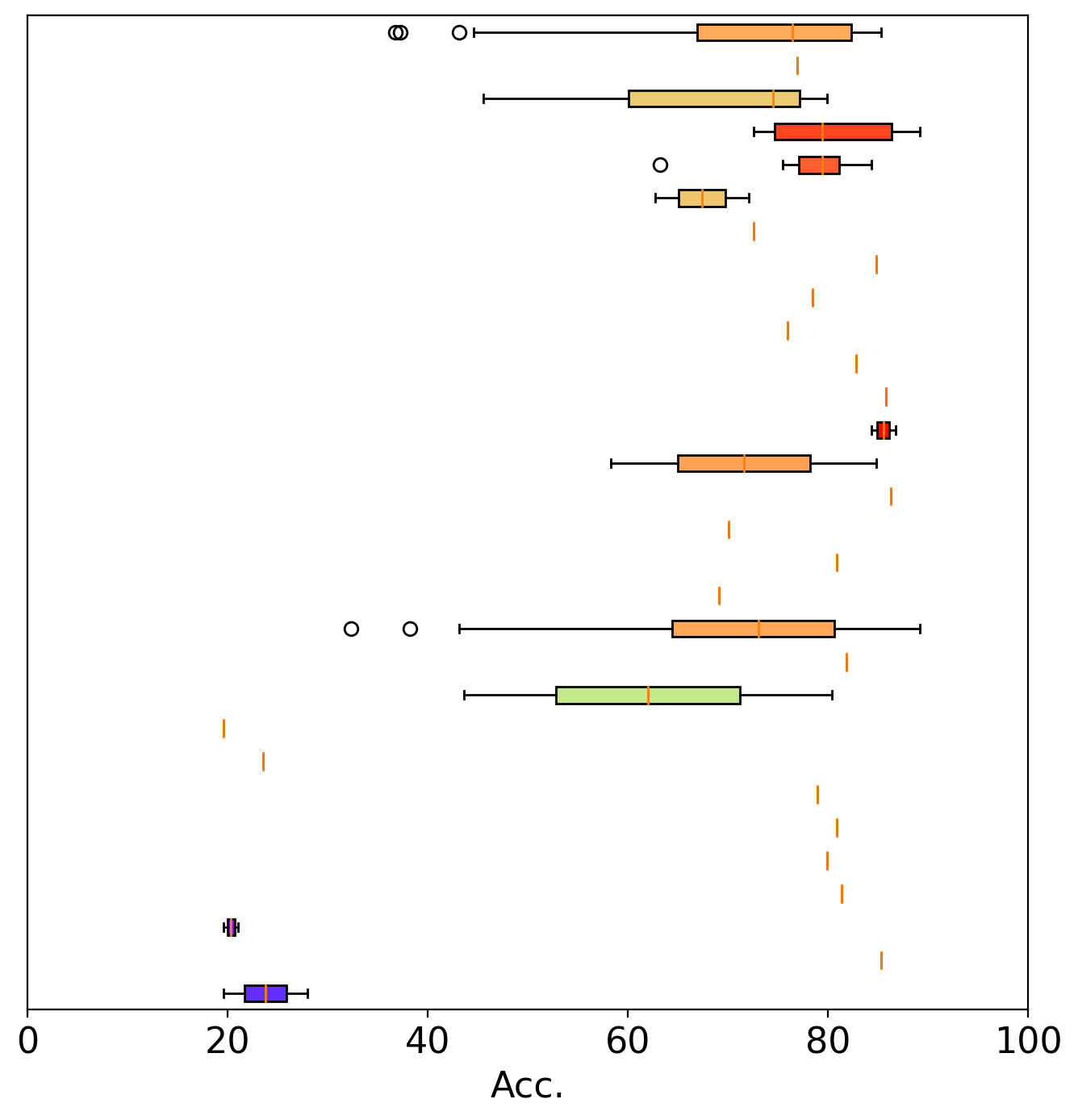} 
        
    \end{subfigure}
    \begin{subfigure}[t]
    {0.313\textwidth}
        \centering
        \caption{$k=10$}
        \includegraphics[width=1.0\linewidth]{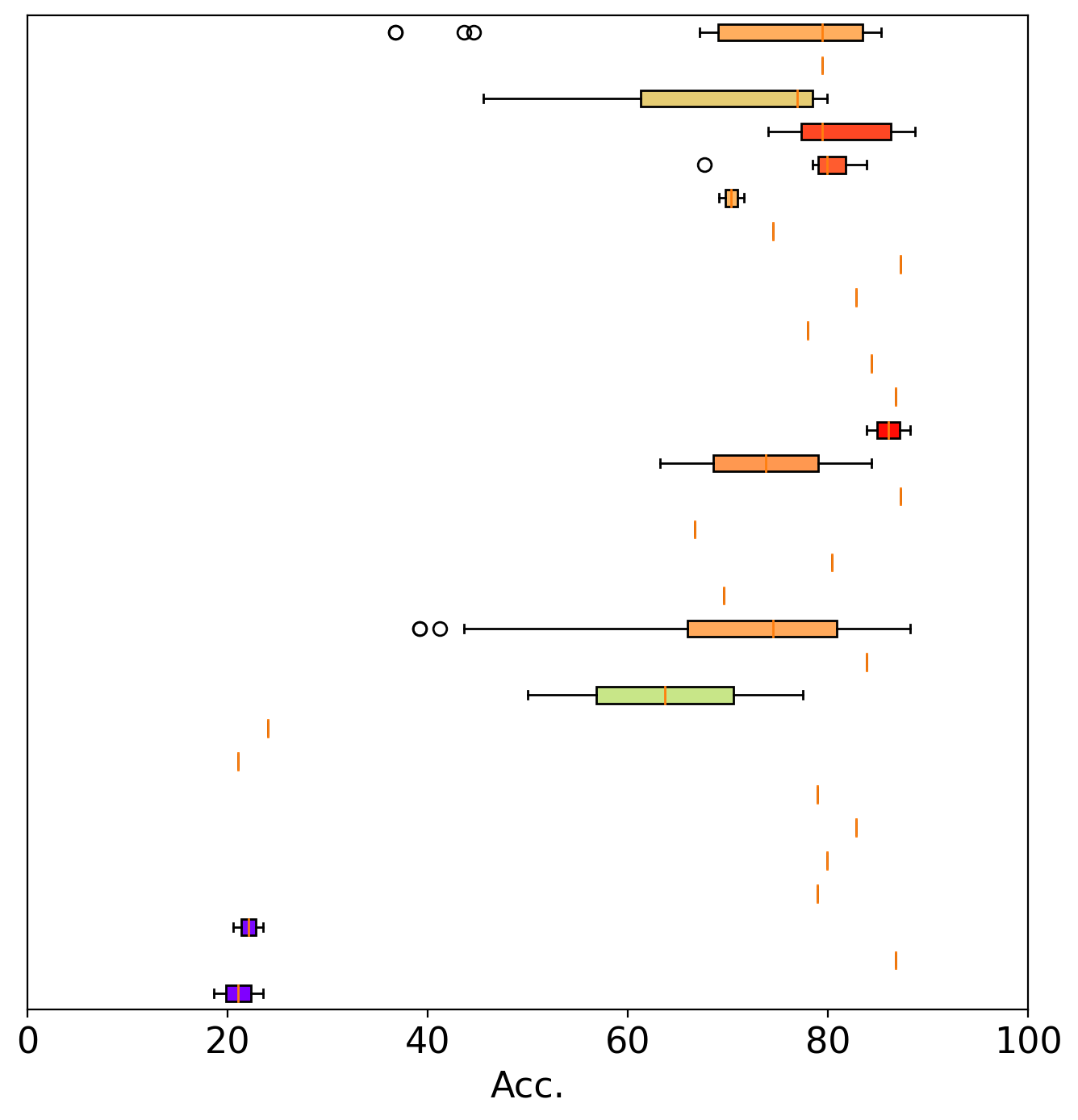} 
        
    \end{subfigure}
    \caption{Results for the retrieval-based classification task on the SIB-200 dataset, using $k$ values of [1, 5, 10], across various language scripts.}
    \label{classification-lang-scripts-sib200-results}
\end{figure*} 

\begin{table}[!t]
\centering
\resizebox{.41\textwidth}{!}{
    \begin{tabular}{lll}
    \toprule
    \textbf{Parameter} & \textbf{HF models} & \textbf{APIs} \\ \midrule 
    top-p & 1 &  1 \\
    seed & - & 42 \\
    temperature & 0.2 & 0 \\
    max new tokens & 10 & 64
    \\ \bottomrule
    \end{tabular}
}
\caption{Hyper-parameters for model inference using Hugging Face models, such as BLOOMZ, mT0, XGLM, Aya-23, Aya-101, Gemma 1.1 7B Instruct, and Llama 8B Instruct and APIs, including Command-R, GPT-3.5 Turbo and GPT-4o.} 
\label{hyper-parameters-api}
\end{table}

\subsection{Retrieved Samples}
We conduct a detailed comparison of the retrieved samples to assess their quality in terms of semantic relevance to the query. Table~\ref{qualitative-analysis-retrieval} presents a comparative analysis between the retrieved samples from E5$_\text{LARGE}$ and XLM-R$_\text{BASE}$. Moreover, Table~\ref{qualitative-analysis-retrieval-labse} showcases the retrieved samples from LaBSE. Our evaluation reveals that the samples retrieved from E5$_\text{LARGE}$ and LaBSE predominantly contain correct labels, with four out of five labels being accurate. In contrast, the samples retrieved by XLM-R$_\text{BASE}$ exhibit a lower accuracy rate, with only two out of five labels being correct. This analysis underscores the varying performance in sample quality and label accuracy across the different models, emphasizing the significance of retrieval quality in downstream tasks.

\section{Detailed Results}

\subsection{Bitext Retrieval Results}

Table~\ref{full-results-bitext-retrieval} presents the complete empirical results for each dataset and model in the bitext retrieval task. Generally, there is a positive trend in model performance as the number of $k$ samples increases.

\subsection{Retrieval-based Classification Results}
Table~\ref{retrieval-based-classification} presents the complete results for the retrieval-based classification task in both Mono and CS settings. Table~\ref{retrieval-based-classification-cross-lingual} provides the full results for the XS and XS CS settings. Figure~\ref{classification-sib200-results} presents the performance results across various language families on the SIB-200 dataset for different values of $k$. Notably, Indo-European languages consistently achieve the highest accuracies. In contrast, Afro-Asiatic, Austroasiatic, and Sino-Tibetan language families exhibit the greatest standard deviations in their results. Figure~\ref{classification-lang-scripts-sib200-results} shows the performance results across various language scripts on the SIB-200 dataset for different values of $k$. It is evident that the Latin script generally achieves the highest performance, albeit with the highest standard deviation. Conversely, the scripts Nkoo, Olck, Tibt, and Tfng exhibit the lowest performance.

\subsection{ICL Classification Results}
Table~\ref{results-icl-full} presents the complete results for ICL classification task in Mono, XS, CS, and XS CS settings.

\section{Prompt Examples}
Prompt examples used for ICL classification are provided in Tables~\ref{icl-prompt-templates-labse} and~\ref{icl-prompt-templates-e5}. Specifically, we use two different templates: for direct prediction, label options are added to the prompt; for prediction by calculating label probabilities, label options are omitted, resulting in shorter prompts.

\section{DistFuse}
We conduct a simplified hyper-parameter tuning process to determine the optimal weights for each model. Due to time constraints, we explore only a few weight combinations. For DistFuse (2), we evaluate two combinations: (1) [$\alpha$ = 1 and $\beta$ = 1], and (2) [$\alpha$ = 1 and $\beta$ = 3]. For Dist (3), we assess three combinations: (1) [$\alpha$ = 1, $\beta$ = 1, $\gamma$ = 1], (2) [$\alpha$ = 1, $\beta$ = 1, $\gamma$ = 3], and (3) [$\alpha$ = 1, $\beta$ = 2, $\gamma$ = 3].

\begin{table*}[!ht]
\centering
\resizebox{\textwidth}{!}{
    \begin{tabular}{lll|lll}
    \toprule  \multicolumn{3}{c|}{\textbf{E5}$_\textbf{LARGE}$} & \multicolumn{3}{c}{\textbf{XLM-R}$_\textbf{BASE}$} 
    \\ 
\multicolumn{1}{c}{\textbf{sample}} & \textbf{label} & \multicolumn{1}{c|}{\textbf{dist}} & \multicolumn{1}{c}{\textbf{sample}} & \textbf{label} & \multicolumn{1}{c}{\textbf{dist}} \\ \midrule
    \multicolumn{6}{l}{\textbf{Query:} Cepak saka hotelku nginep, namung digawa mlaku, ing kene akeh tenan pilian panganane, panggonane sing amba, lan nyenengake} \\ 
    \multicolumn{6}{l}{\textbf{Translation (in English):} Near the hotel I stayed in, reachable by foot, so many food choice here, the place is huge, and fun} \\ 
    \textbf{Label:} positive \\
    \midrule
    Miturutku mangan ana ing kene porsine akeh lan   & positive & 0.436 & Panggonan iki nyediakake pirang-pirang panganan, & positive & 0.923 \\
    regane murah, banjur panganane cepet tekane & & & nanging sing aku jajal mesthi wae batagore, panggonane .  & &  \\
    maneh lan panggonane uga resik lan amba & & & uga resik & & \\ 
    & & & & \\
    \textbf{Translation (in English): } In my opinion, eating & & & \textbf{Translation (in English):} This place served several & \\ 
    here will grant you large portions for a cheap & & & food, but of course the one I tried was the batagor, .\\ 
    price, add to the fact that it's served quickly, & & & place was clean too \\
    too, and the place being clean and wide. & & & \\ \midrule
    Aku seneng banget mangan ning restoran iki,   & positive & 0.452 & Timku bakal nganakake mangan mbengi tema ing burgundy.   & negative &  0.972 \\ 
    menu masakane rena-rena, rasane enak, regane. & & &  Katimbang ilang, aku lan carikku njajal ngecek panggonane  \\ 
     ora tek larang & & & ndhisik sadina sakdurunge. Sisan uga tes panganane. Dalane \\
    & & & adoh banget lan munggah medun bukit. Luwih nemen   \\ 
    \textbf{Translation (in English):} I really love eating & & & maneh pas dhewe mara mrana kuwi dina minggu sore  \\
    in this restaurant. Varied menu, awesome & & & dadi macet. Tekane ing kana sih pemandangane oke. \\ 
    flavours, and not really that expensive.& && Nanging model restorane biasa wae. \\ 
    & & & \\ 
    & & & \textbf{Translation (in English):} My team will be having a theme \\ 
    & & &  dinner in burgundy. Instead of getting lost, my secretary\\ 
    & & & and I tried to check the place first the day before. Also\\
    & & & test the food. The journey is very long and goes up and  \\
    & & & down hills. What made it worse was that when we went \\
    & & & there it was a Sunday afternoon so there was traffic jam.\\
    & & & When we got there, the view was okay. But the restaurant\\
    & & & layout is ordinary\\ \midrule
    Ing restoran iki panganan kang disediakake akeh & positive & 0.452 & Pithik gorenge enak ing kene. Cocok kanggo sing lebare   & positive & 0.974 \\
    banget lan regane cukup kajangkau, kahanane sek   & & & perjalanan adoh. Aku marang kene mulih saka njaba   \\ 
    enak lan nyaman & & & kutha, dadi mangane pas ngelih ngono deh. Marakake  \\
    & & & weteng wareg, panganane enak banjur pelayanane mantap.  \\ 
    \textbf{Translation (in English):} In this restaurant& & & Kasire ayu ayu \\ 
     there is a lot of food provided and the prices are & & & \\ 
    quite affordable, the atmosphere is delicious and & & & \textbf{Translation (in English):} The fried chicken is amazing \\ 
    comfortable & & & here. Perfect after a long trip. I came here after returning \\ 
    & & & out of town, so I was absolutely starving. My stomach \\
    & & & was filled right back up. The food was good and servers \\ 
    & & & were great. Not to mention, the cashiers were beautiful \\ \midrule
    Panggonan iki nyediakake pirang-pirang panganan,  & positive & 0.467 & Bingung arep mangan nandi sing panggone asik, panganane  & positive & 1.009 \\ 
    nanging sing aku jajal mesthi wae batagore, . & & & enak lan regane terjangkau? Mrene ae. Aku lan bojoku  \\
    panggonane uga resik & & & nikmati banget. Sing mara akeh-akeh cah enom dadi   \\
    & & & melu-melu ngrasa enom maneh. \\
    \textbf{Translation (in English):} This place served & & & \\ 
    several food, but of course the one I tried was  & & & \textbf{Translation (in English):} Don't know where to have \\ 
    the batagor, place was clean too. & & &  a nice and affordable place to grab a bite? Just come\\ 
    & & & right here. Me and my partner are really enjoying it.\\ 
    & & & Most of the customers are young people, making \\
    & & & us feel just as young again. \\ \midrule
    Kuota dadi entek resik kanggo ndelok foto-foto sing   & negative & 0.477 & Panganane lumayan, nanging ana pelayan sing lumayan  & negative & 1.018\\
    mung gawe aku srei, panganan enak-enak sing marai  & & & kemproh war dadi kurang nyaman. Kanggo panganan rada  \\
    ngiler & & & cepet yo ben ora kangelihen konsumene. Isih akeh  \\
    & & & sing kudu ditingkatake.
    \\ 
    \textbf{Translation (in English):} My quota is drained & & & \\ 
    dry just to see photos that make me jelly, and  &&& \textbf{Translation (in English):} The food was okay, but there  \\ 
    delicious food that makes my mouth water. &&& was this one server who was kinda dirty, making it a \\
    && & little less comfortable. Please serve the food quicker\\
    &&& so the customers won't get hungry. There are many things\\
    &&& to improve. \\\bottomrule
    \end{tabular}
}
\caption{Retrieved samples from E5$_\text{LARGE}$ and XLM-R$_\text{BASE}$.}
\label{qualitative-analysis-retrieval}
\end{table*}

\begin{table*}[!ht]
\centering
\resizebox{0.65\textwidth}{!}{
    \begin{tabular}{lcc}
    \toprule  \multicolumn{3}{c}{\textbf{LaBSE}} \\ 
\multicolumn{1}{c}{\textbf{sample}} & \textbf{label} & \multicolumn{1}{c}{\textbf{dist}} \\ \midrule
    \multicolumn{3}{l}{\textbf{Query:} Cepak saka hotelku nginep, namung digawa mlaku, ing kene} \\
    \multicolumn{3}{l}{akeh tenan pilian panganane, panggonane sing amba, lan nyenengake} \\ 
    \multicolumn{3}{l}{\textbf{Translation (in English):} Near the hotel I stayed in, reachable} \\
    \multicolumn{3}{l}{by foot, so many food choice here, the place is huge, and fun} \\ 
    \textbf{Label:} positive \\ \midrule
    Ing restoran iki panganan kang disediakake   & positive & 0.896 \\
    akeh banget lan regane cukup kajangkau, & &  \\
    kahanane sek enak lan nyaman & \\ \\
    \textbf{Translation (in English):} In this restaurant \\
    there is a lot of food provided and the prices are\\
    quite affordable, the atmosphere is delicious and\\
    comfortable\\
    \midrule
    Wektu pengen mangan variasi panganan, & positive & 0.934 \\
    piliane mesthi Hanamasa. Lokasi panggonane\\
    cukup enak. Pilian panganane akeh, saka awit \\
    camilan, bakar-bakaran, godhokan nganti \\
    panganan panutup. Ora nggelakne banget.  \\ \\
    \textbf{Translation (in English):} When you wanna \\
    enjoy a variety of food, the first choice has \\
    to be Hanamasa. The location's pretty great. \\
    Lots of food you can choose from, ranging from \\
    snacks, barbeques, boiled food, all the way to \\
    desserts. Not bad at all!\\ \midrule
Mangan abreng karo dulur-dulur wedok kala  & positive & 0.938 \\
wingi, panggon nyaman, enak
kanggo nongkrong, \\
pelayanane apik. Wis ping bolak-balik mangan \\
ning kene.  \\ \\
\textbf{Translation (in English):} Dined togetha with\\
da sistahs a lil' bit ago, cosy place, nice to hang out, \\
good service. Have gone ta this place multiple times.
\\ \midrule
Aku seneng banget mangan ning restoran iki, & positive & 0.940 \\
menu masakane rena-rena, rasane enak, regane \\
ora tek larang.  \\ 
\\
\textbf{Translation (in English):} I really love \\
eating in this restaurant. Varied menu, awesome flavours,\\
and not really that expensive.\\ \midrule
Pithik gorenge enak ing kene. Cocok kanggo & positive & 0.946\\
sing lebare perjalanan adoh. Aku marang kene\\
mulih saka njaba kutha, dadi mangane pas \\
ngelih ngono deh. Marakake weteng wareg, \\
panganane enak banjur pelayanane mantap. \\
Kasire ayu ayu \\ \\
\textbf{Translation (in English):} The fried chicken \\
is amazing here. Perfect after a long trip. I came\\
here after returning out of town, so I was \\
absolutely starving. My stomach was filled right\\
back up. The food was good and servers were great. \\
Not to mention, the cashiers were beautiful \\ \bottomrule
    \end{tabular}
}
\caption{Retrieved samples from LaBSE.}
\label{qualitative-analysis-retrieval-labse}
\end{table*}

\begin{table*}[!ht]
\centering
\resizebox{\textwidth}{!}{
    \begin{tabular}{lcccccc|ccc|c|c}
    \toprule
    \textbf{Model} & \multicolumn{6}{c}{\textbf{Cross-lingual (XL)}} & \multicolumn{3}{|c|}{\textbf{Code-Switching (CS)}}
    & Micro & Macro\\
     & BUCC & NollySenti & NusaX & NusaT & Tatoeba & avg. & LinCE MT & PHINC & avg. & avg. & avg.\\ 
     \midrule 
    metric & F1 & F1 & F1 & F1 & F1 & & F1 & F1 & & &\\ \midrule
    Fine-tune (SOTA) & 99.00 & N/A & N/A & N/A & 83.80 & N/A & N/A & N/A & N/A & N/A & N/A \\ \midrule
    $k=1$ \\ \midrule
    LaBSE & 98.77 & 80.52 & 77.89 & 81.17 & 81.14 & 83.90 & 34.36 & 69.70 & 52.03 & 74.79 & 67.97 \\
    CMLM & 98.64 & 58.06 & 55.64 & 63.08 & 78.43 & 70.77 & 29.34 & 55.91 & 42.62 & 62.73 & 56.70 \\
    E5$_\text{BASE}$ & 98.33 & 63.40 & 68.01 & 63.52 & 68.06 & 72.26 & 29.17 & 57.41 & 43.29 & 63.99 & 57.78 \\
    E5$_\text{LARGE}$ & 98.66 & 67.50 & 72.67 & 67.20 & 75.73 & 76.35 & 34.32 & 65.63 & 49.97 & 68.82 & 63.16 \\
    MPNet$_\text{BASE}$v2 & 98.05 & 22.64 & 38.46 & 40.52 & 61.60 &52.25 & 14.04 & 37.70 &25.87 & 44.72 & 39.06 \\ 
    MiniLM$_\text{L12-E384}$ & 57.98 & 8.26 & 7.52 & 19.66 & 30.70 &24.82& 4.85 & 14.95 &9.90& 20.56 & 17.36 \\ 
    Glot-500 & 17.90 & 11.49 & 5.78 & 27.65 & 10.58 &14.68& 5.76 & 27.52 &16.64& 15.24 & 15.66  \\ 
    XLM-R$_\text{BASE}$ & 39.70 & 7.59 & 8.05 & 20.97 & 12.62 &17.79& 4.15 & 17.06 &10.61& 15.73 & 14.20\\
    XLM-R$_\text{LARGE}$ &  26.51 & 5.53 & 5.03 & 18.60 & 6.57 &12.45& 2.20 & 9.88 &6.04& 10.62 & 9.25 \\ \midrule
    Cohere-Embedv3 & 98.76 & 62.91 & 76.51 & 69.13 & 74.66 &76.39& 34.44 & 72.07 &53.25& 69.78 & 64.82 \\ 
    OpenAI-Embedv3-large &  98.98 & 35.84 & 70.94 & 73.74 & 65.61 &69.02& 46.60 & 90.87 &68.73& 68.94 & 68.88 \\ \midrule
    DistFuse (2) & 98.90  &      80.29  &  80.96  &            80.28   &   83.19 & 84.72 & 37.97 &   74.97& 56.47& 76.65 &70.60\\
    DistFuse (3) 
    & 98.90     &     77.02   & 81.25      &        77.61  &    81.64& 83.28& 37.23  &  76.42 & 56.83 &  75.72 & 70.06\\ \midrule
    $k=5$ \\ \midrule
    LaBSE & 99.12 & 89.94 & 84.98 & 88.57 & 89.15 & 90.35 & 56.69  &  79.17 & 67.93 & 83.95 & 79.14\\
CMLM & 99.15 & 71.54 & 65.91 & 74.48 & 87.38 & 79.69 & 50.96    & 66.91 & 58.94  & 73.76& 69.32\\
E5$_\text{BASE}$ & 99.05 & 77.58 & 79.47 & 74.68 & 80.42& 82.24 & 51.48  &  69.79 & 60.64 & 76.07&  71.44\\
E5$_\text{LARGE}$ & 99.19 & 79.70 & 82.60 & 78.76 & 86.27 & 85.30 & 56.80  &   75.96 & 66.38  & 79.90 & 75.84\\ 
MPNet$_\text{BASE}$v2 & 99.01 & 29.95 & 50.24 & 51.55 & 70.38 & 60.23 & 26.26 & 47.28 & 36.77 & 53.52&  48.50 \\ 
MiniLM$_\text{L12-E384}$ & 72.82 & 14.72 & 14.07 & 29.01 & 45.39 & 35.20 & 10.47  &   20.33 & 15.40  & 29.54&  25.30\\ 
Glot-500 & 32.26 & 20.92 & 14.39 & 38.51 & 21.28 & 25.47 & 11.30    & 35.22 & 23.26  &24.84 & 24.37\\
XLM-R$_\text{BASE}$ & 57.23 & 15.00 & 15.74 & 29.42 & 20.87 & 27.65 & 9.08  &  22.27 & 15.68  &  24.23 & 21.67\\
XLM-R$_\text{LARGE}$ & 41.06 & 10.66 & 10.73 & 25.65 & 13.61 & 20.34 &  4.24   & 11.68 &  7.96  & 16.80 & 14.15\\ \midrule
Cohere-Embedv3 & 99.29 & 76.19 & 85.08 & 80.72 & 84.98 & 85.25 &57.28   & 82.05 &  69.66  &   80.80 & 77.46\\
OpenAI-Embedv3-large & \underline{99.43} & 43.39 & 79.37 & 83.05 & 76.77 & 76.40 & 74.92  & \underline{94.64} &  \underline{84.78} & 78.80 & 80.59\\ \midrule
DistFuse (2) & 99.21 & 90.66 & 88.08 & 89.13 & 91.02 & 91.62 &  61.70  &  84.21 &  72.95 &  86.29 & 82.29 \\
DistFuse (3) & 99.26 & 87.65 & 88.24 & 87.36 & 90.07 & 90.52 &  61.27  &  85.22 & 73.25 & 85.58 &  81.89\\ \midrule
$k=10$\\ \midrule
LaBSE & 99.17 & \underline{92.62} & 87.67 & 90.21 & 91.02 & 92.14 & 61.54 & 82.15  & 71.84 &  86.34 &  81.99 \\
CMLM & 99.17 & 77.54 & 70.81 & 78.35 & 89.53 & 83.08  & 56.44  & 70.72  & 63.58  & 77.51  & 73.33\\
E5$_\text{BASE}$ & 99.18 & 83.07 & 83.76 & 78.52 & 83.89 & 85.68 &  56.85 &  74.34 &  65.59 &  79.94 &  75.64\\
E5$_\text{LARGE}$ & 99.31 & 83.78 & 86.19 & 82.35 & 88.80 & 88.09  & 61.6 &  79.57  & 70.58  & 83.09  & 79.34\\ 
MPNet$_\text{BASE}$v2 & 99.13 & 32.75 & 55.52 & 55.61 & 73.33 & 63.27 &  30.61  & 50.74 &  40.67 &  56.81  & 51.97\\ 
MiniLM$_\text{L12-E384}$ & 78.08 & 20.25 & 19.84 & 33.34 & 51.89 & 40.68 &  13.52 &23.29 &  18.41 &  34.32  & 29.55 \\ 
Glot-500 & 39.08 & 26.72 & 20.60 & 43.06 & 27.69 & 31.43 & 14.01 &  38.30  & 26.15 &  29.92  & 28.79\\
XLM-R$_\text{BASE}$ & 63.56 & 20.01 & 20.66 & 33.19 & 25.61 & 32.61  & 11.50 &  24.83 &  18.16 &  28.48  & 25.39\\
XLM-R$_\text{LARGE}$ & 47.24 & 14.03 & 14.34 & 28.31 & 17.79 & 24.34  & 5.11 &  12.70  & 8.90 & 19.93 &  16.62\\ \midrule
Cohere-Embedv3 & 99.39 & 81.16 & 88.01 & 84.24 & 87.56 & 88.07 & 62.23  & 84.82 &  73.52 &  83.92 & 80.80\\
OpenAI-Embedv3-large & \textbf{99.50} & 47.01 & 82.55 & 85.61 & 80.20 & 78.97  & 79.75  & \textbf{95.36} & \textbf{87.56} &  81.43 & 83.27\\ \midrule
DistFuse (2) & 99.29 & \textbf{93.21} & \textbf{90.48} & \textbf{91.12} & \textbf{92.85} & \textbf{93.39} &  \textbf{66.20} &  86.75  & 76.47  & \textbf{88.56} &  \textbf{84.93}\\
DistFuse (3) & 99.39 & 90.45 & \underline{90.40} & \underline{89.83} & \underline{91.87} & \underline{92.39} &  \underline{65.94} &  87.54  & 76.74 &  \underline{87.92}  & \underline{84.57} \\ \bottomrule
    \end{tabular}
}
\caption{Results on bitext retrieval. \textbf{Bold} and \underline{underlined} numbers present the best and second-best models.} 
\label{full-results-bitext-retrieval}
\end{table*}            

\begin{table*}[!ht]
\centering
\resizebox{\textwidth}{!}{
    \begin{tabular}{lccccc|ccc|c|c}
    \toprule
    \textbf{Model} & \multicolumn{5}{c}{\textbf{Monolingual (Mono)}} & \multicolumn{3}{|c|}{\textbf{Code-Switching (CS)}}& Micro & Macro\\
    & MASSIVE & NollySenti & NusaX & SIB-200 & avg. & FIRE 2020 & LinCE SA & avg. & avg. & avg. \\ \midrule
    metric & Acc. & Acc. & F1 & Acc. & & Acc. & Acc. & & & \\
    \midrule
    Random & 1.67 & 33.33 & 33.33 & 14.29 & 20.66 & 25.00 & 33.33 & 29.00 & 23.49 & 24.83 \\
    Majority & 7.03 & 50.00 & 18.44 & 25.00 & 25.12 & 53.90 &  55.78 & 54.84 & 35.03 & 39.98 \\
    Fine-tune (SOTA) & \textbf{86.10} & \underline{88.80} & \textbf{80.00} & \textbf{75.90} & \textbf{82.70} & N/A$^\ddagger$ & N/A$^\ddagger$ & N/A & N/A & N/A \\
    Fine-tune (XLM-R$_\text{BASE}$) & \underline{85.04} & 87.16 & \underline{75.43} & 70.55 & \underline{79.55} & \textbf{68.78} & 55.78 & \underline{62.28} & \textbf{73.79} & \textbf{70.92} \\ \midrule
    $k=1$ \\ \midrule
    LaBSE & 76.55 & 80.04 & 62.23 & 61.14 & 69.99 & 56.56 & 49.92 & 53.24 & 64.41 & 61.62 \\
    CMLM & 76.24 & 79.48 & 63.40 & 60.42 & 69.89 & 54.83 & 48.63 & 51.73 & 63.83 & 60.81\\
E5$_\text{BASE}$ & 74.82 & 82.96 & 65.59 & 62.23 & 71.40 & 57.14 & 50.03 & 53.59 & 65.46 & 62.49\\
E5$_\text{LARGE}$ & 76.67 & 85.24 & 67.14 & 66.64 & 73.92 & 58.25 & 51.00 & 54.63 & 67.49 & 64.27\\
MPNet$_\text{BASE}$v2 & 69.41 & 75.24 & 53.29 & 56.24 & 63.55 & 51.21 & 49.70 & 50.46 & 59.18 & 57.00\\
MiniLM$_\text{L12-E384}$ & 63.32 & 72.28 & 58.35 & 39.77 & 58.43 & 51.49 & 49.00 & 50.25 & 55.70 & 54.34\\
Glot-500 & 64.01 & 75.52 & 57.00 & 51.76 & 62.07 & 53.48 & 48.84 & 51.16 & 58.44 & 56.62 \\
XLM-R$_\text{BASE}$ & 61.93 & 74.56 & 58.29 & 43.66 & 59.61 & 53.57 & 47.44 & 50.51 & 56.57 & 55.06\\
XLM-R$_\text{LARGE}$ & 60.39 & 73.36 & 57.62 & 40.66 & 58.01 & 52.17 & 47.18 & 49.68 & 55.23 & 53.84\\\midrule
Cohere-Embedv3 & 77.78 & 86.80 & 68.54 & 71.08 & 76.05 & 59.30 & 51.43 & 55.37 & 69.16 & 65.71\\
OpenAI-Embedv3-large & 74.97 & 79.56 & 63.61 & 67.44 & 71.40 & 61.19 & 51.37 & 56.28 & 66.36 & 63.84 \\\midrule
DistFuse (2) & 78.18 & 84.72 & 66.65 & 68.32 & 74.47 & 58.89 & 50.73 & 54.81 & 67.92 & 64.64\\
DistFuse (3) & 78.59 & 86.24 & 67.44 & 70.76 & 75.76 & 59.15 & 50.94 & 55.05 & 68.85 & 65.40\\\midrule
$k=5$\\\midrule
LaBSE & 78.62 & 82.08 & 66.90 & 64.67 & 73.07 & 63.65 & 53.85 & 58.75 & 68.30 & 65.91\\
CMLM & 78.38 & 80.60 & 67.07 & 64.62 & 72.67 & 61.73 & 54.87 & 58.30 & 67.88 & 65.48\\
E5$_\text{BASE}$ & 77.13 & 85.96 & 69.16 & 66.82 & 74.77 & 63.38 & 55.51 & 59.45 & 69.66 & 67.11\\
E5$_\text{LARGE}$ & 79.10 & 87.20 & 71.72 & 71.05 & 77.27 & 64.14 & 57.40 & 60.77 & 71.77 & 69.02\\
MPNet$_\text{BASE}$v2 & 71.24 & 79.12 & 54.76 & 59.20 & 66.08 & 56.48 & 54.76 & 55.62 & 62.59 & 60.85\\
MiniLM$_\text{L12-E384}$ & 65.16 & 76.28 & 63.84 & 44.56 & 62.46 & 57.96 & 52.23 & 55.10 & 60.01 & 58.78\\
Glot-500 & 65.72 & 78.60 & 60.08 & 57.49 & 65.47 & 59.65 & 51.37 & 55.51 & 62.15 & 60.49\\
XLM-R$_\text{BASE}$ & 63.54 & 76.24 & 61.32 & 48.22 & 62.33 & 60.35 & 53.42 & 56.89 & 60.52 & 59.61\\
XLM-R$_\text{LARGE}$ & 62.08 & 76.20 & 60.57 & 45.44 & 61.07 & 59.11 & 52.56 & 55.84 & 59.33 & 58.45\\\midrule
Cohere-Embedv3 & 80.15 & 88.12 & 71.00 & 74.73 & 78.50 & 65.12 & 57.56 & 61.34 & 72.78 & 69.92\\
OpenAI-Embedv3-large & 77.32 & 80.64 & 67.77 & 69.88 & 73.90 & 66.19 & 56.27 & 61.23 & 69.68 & 67.57\\\midrule
DistFuse (2) & 80.42 & 87.00 & 71.90 & 72.13 & 77.86 & 64.21 & 56.16 & 60.19 & 71.97 & 69.02\\
DistFuse (3) & 80.92 & 88.48 & 71.70 & 74.63 & 78.93 & 64.69 & 57.13 & 60.91 & 72.93 & 69.92\\\midrule
$k=10$\\\midrule
LaBSE & 78.47 & 82.48 & 67.39 & 65.50 & 73.46 & 64.73 & 56.54 & 60.64 & 69.19 & 67.05\\
CMLM & 78.21 & 82.04 & 67.11 & 64.84 & 73.05 & 62.96 & 55.57 & 59.27 & 68.46 & 66.16\\
E5$_\text{BASE}$ & 77.18 & 86.36 & 69.07 & 67.72 & 75.08 & 64.71 & 57.61 & 61.16 & 70.44 & 68.12\\
E5$_\text{LARGE}$ & 79.02 & 88.00 & 71.15 & 71.91 & 77.52 & 65.30 & 58.53 & 61.92 & 72.32 & 69.72\\
MPNet$_\text{BASE}$v2 & 70.75 & 80.40 & 53.85 & 59.67 & 66.17 & 59.26 & 57.40 & 58.33 & 63.56 & 62.25\\
MiniLM$_\text{L12-E384}$ & 64.47 & 77.12 & 64.27 & 46.87 & 63.18 & 60.61 & 53.95 & 57.28 & 61.22 & 60.23\\
Glot-500 & 65.14 & 79.36 & 58.69 & 59.47 & 65.67 & 62.04 & 54.17 & 58.11 & 63.15 & 61.89\\
XLM-R$_\text{BASE}$ & 62.98 & 78.40 & 62.72 & 50.39 & 63.62 & 62.06 & 54.44 & 58.25 & 61.83 & 60.94\\
XLM-R$_\text{LARGE}$ & 61.58 & 77.56 & 60.62 & 47.29 & 61.76 & 60.92 & 53.68 & 57.30 & 60.28 & 59.53\\\midrule
Cohere-Embedv3 & 80.15 & 88.64 & 69.87 & \underline{75.57} & 78.56 & 65.88 & 58.36 & 62.12 & 73.08 & 70.34\\
OpenAI-Embedv3-large & 77.27 & 82.28 & 66.80 & 69.54 & 73.97 & \underline{67.33} & 58.20 & \textbf{62.77} & 70.24 & 68.37\\\midrule
DistFuse (2) & 80.38 & 88.28 & 71.83 & 72.88 & 78.34 & 65.73 & \textbf{58.53} & 62.13 & 72.94 & 70.24\\
DistFuse (3) & 80.79 & \textbf{88.96} & 70.99 & 75.32 & 79.02 & 65.97 & \underline{58.42} & 62.20 & \underline{73.41} & \underline{70.61}
    \\ \bottomrule
    \end{tabular}
}
\caption{Results on retrieval-based classification. \textbf{Bold} and \underline{underlined} numbers present the best and second-best models. $^\ddagger$For FIRE 2020, we modify the labels, thus there are no comparable results in the literature. For LinCE SA, we evaluate on the development split and we could not find any comparable result in the literature.}
\label{retrieval-based-classification}
\end{table*}

\begin{table*}[!ht]
\centering
\resizebox{\textwidth}{!}{
    \begin{tabular}{lccccc|c|c|c}
    \toprule
    \textbf{Model} & \multicolumn{5}{c|}{\textbf{Cross-lingual (XL)}} & \multicolumn{1}{c|}{\textbf{Code-Switching (CS)}} & Micro & Macro \\
    & MASSIVE & NollySenti & NusaX & SIB-200 & avg. & \multicolumn{1}{|c|}{FIRE 2020} & avg. & avg. \\ \midrule    
    source lang. & eng & en & eng & eng\_Latn & & tamil\\ 
    metric & Acc. & Acc. & F1 & Acc. & & Acc. \\ \midrule
    Random & 1.67 & 33.33 & 33.33 & 14.29 & 20.66 & 25.00 & 21.52 & 21.09 \\
Majority & 7.03 & 50.00 & 18.44 & 25.00 & 25.12 & 41.91 & 28.48 & 26.80\\
Fine-tune (SOTA) & 70.60 & N/A & 52.08 & \underline{69.10} & N/A & N/A$^\dagger$ & N/A & N/A \\
Fine-tune (XLM-R$_\text{BASE}$) & 68.94 & 74.95 & 56.71 & 63.10 & 65.92 & 34.64 & 59.67 & 62.79\\\midrule
$k=1$\\\midrule
LaBSE & 73.96 & 79.80 & 63.65 & 60.18 & 69.40 & 32.94 & 62.11 & 65.75\\
CMLM & 73.08 & 74.00 & 58.98 & 57.51 & 65.89 & 34.87 & 59.69 & 62.79\\
E5$_\text{BASE}$ & 63.43 & 74.30 & 34.08 & 63.11 & 58.73 & 35.53 & 54.09 & 56.41\\
E5$_\text{LARGE}$ & 69.38 & 79.85 & 40.73 & 67.63 & 64.40 & 35.91 & 58.70 & 61.55\\
MPNet$_\text{BASE}$v2 & 46.05 & 61.60 & 48.44 & 55.95 & 53.01 & 32.12 & 48.83 & 50.92\\
MiniLM$_\text{L12-E384}$ & 35.72 & 62.20 & 41.15 & 30.50 & 42.39 & 31.60 & 40.23 & 41.31\\
Glot-500 & 24.66 & 66.70 & 44.45 & 40.08 & 43.97 & 33.16 & 41.81 & 42.89\\
XLM-R$_\text{BASE}$ & 27.49 & 64.85 & 36.41 & 33.98 & 40.68 & 32.42 & 39.03 & 39.86\\
XLM-R$_\text{LARGE}$ & 20.38 & 66.50 & 34.19 & 28.04 & 37.28 & 31.75 & 36.17 & 36.72\\\midrule
Cohere-Embedv3 & 70.87 & \underline{81.30} & 65.29 & \textbf{69.67} & 71.78 & 35.68 & 64.56 & 68.17\\
OpenAI-Embedv3-large & 61.09 & 67.85 & 65.45 & 67.36 & 65.44 & 31.90 & 58.73 & 62.08
\\\midrule
$k=5$\\\midrule
LaBSE & 75.80 & \textbf{81.80} & 68.25 & 63.75 & 72.40 & 38.58 & 65.64 & 69.02\\
CMLM & 75.48 & 78.70 & 64.89 & 58.89 & 69.49 & 38.72 & 63.34 & 66.41\\
E5$_\text{BASE}$ & 66.83 & 73.45 & 51.82 & 67.43 & 64.88 & 40.28 & 59.96 & 62.42\\
E5$_\text{LARGE}$ & 72.48 & 78.60 & 60.99 & 71.53 & 70.90 & 40.28 & 64.78 & 67.84\\
MPNet$_\text{BASE}$v2 & 50.83 & 64.00 & 53.98 & 58.73 & 56.89 & 38.58 & 53.22 & 55.05\\
MiniLM$_\text{L12-E384}$ & 40.19 & 65.55 & 52.79 & 34.81 & 48.34 & 36.80 & 46.03 & 47.18\\
Glot-500 & 28.67 & 73.50 & 49.37 & 47.01 & 49.64 & 37.61 & 47.23 & 48.43\\
XLM-R$_\text{BASE}$ & 31.27 & 69.15 & 39.52 & 39.89 & 44.96 & 38.58 & 43.68 & 44.32\\
XLM-R$_\text{LARGE}$ & 24.74 & 69.20 & 36.13 & 34.19 & 41.07 & 37.69 & 40.39 & 40.73\\\midrule
Cohere-Embedv3 & 74.18 & 78.60 & 64.59 & 74.62 & \textbf{73.00} & 40.28 & \underline{66.45} & \textbf{69.73}\\
OpenAI-Embedv3-large & 63.62 & 66.15 & \underline{69.22} & 69.09 & 67.02 & 38.43 & 61.30 & 64.16\\\midrule
DistFuse (2) & 77.53 & 79.25 & 63.74 & 65.03 & 71.39 & 39.24 & 64.96 & 68.17\\
DistFuse (3) & 77.27 & 78.25 & 61.67 & 66.00 & 70.80 & 38.65 & 64.37 & 67.58\\\midrule
$k=10$\\\midrule
LaBSE & 75.89 & 81.20 & 68.54 & 65.29 & 72.73 & 41.10 & 66.40 & 69.57\\
CMLM & 75.77 & \underline{81.30} & 66.06 & 58.11 & 70.31 & 40.88 & 64.42 & 67.37\\
E5$_\text{BASE}$ & 67.60 & 74.20 & 51.54 & 68.71 & 65.51 & \textbf{42.73} & 60.96 & 63.23\\
E5$_\text{LARGE}$ & 73.09 & 77.50 & 61.40 & 72.33 & 71.08 & 41.99 & 65.26 & 68.17\\
MPNet$_\text{BASE}$v2 & 56.45 & 64.80 & 57.88 & 59.61 & 59.69 & 41.25 & 56.00 & 57.84\\
MiniLM$_\text{L12-E384}$ & 42.07 & 66.55 & 58.66 & 37.34 & 51.16 & 39.61 & 48.85 & 50.00\\
Glot-500 & 30.73 & 74.10 & 51.50 & 50.67 & 51.75 & 40.06 & 49.41 & 50.58\\
XLM-R$_\text{BASE}$ & 32.96 & 70.45 & 45.11 & 41.83 & 47.59 & 41.02 & 46.27 & 46.93\\
XLM-R$_\text{LARGE}$ & 27.18 & 69.50 & 39.62 & 39.20 & 43.88 & 39.47 & 42.99 & 43.43\\\midrule
Cohere-Embedv3 & 74.98 & 77.95 & 61.69 & 76.06 & \underline{72.67} & \underline{42.36} & \textbf{66.61} & \underline{69.64}\\
OpenAI-Embedv3-large & 64.43 & 65.15 & \textbf{69.88} & 69.07 & 67.13 & 40.50 & 61.81 & 64.47\\\midrule
DistFuse (2) & \textbf{77.75} & 78.30 & 62.72 & 64.71 & 70.87 & 40.73 & 64.84 & 67.86\\
DistFuse (3) & \underline{77.67} & 76.70 & 58.94 & 67.43 & 70.19 & 41.77 & 64.50 & 67.34\\ \bottomrule
    \end{tabular}
}
\caption{Results on retrieval-based classification in the cross-lingual setting. The source language is English for all datasets except FIRE 2020, where the source language is Tamil. \textbf{Bold} and \underline{underlined} numbers present the best and second-best models. $^\dagger$We preprocess the dataset differently from the original dataset. Thus, there are no comparable results in the literature.} 
\label{retrieval-based-classification-cross-lingual}
\end{table*}

\begin{table*}[!ht]
\centering
\resizebox{\textwidth}{!}{
    \begin{tabular}{lcccc|cccc|ccc|c|c|c}
    \toprule
    \textbf{Model} & \multicolumn{4}{c|}{\textbf{Mono}} & \multicolumn{4}{c}{\textbf{XL}} & \multicolumn{3}{|c|}{\textbf{CS}}& \multicolumn{1}{|c|}{\textbf{XL CS}} & Micro & Macro \\
    & NollySenti & NusaX & SIB-200 & avg. & NollySenti & NusaX & SIB-200 & avg. & FIRE 2020 & LinCE SA & avg. & \multicolumn{1}{|c|}{FIRE 2020} & avg. & avg. \\ \midrule
    metric & Acc. & F1 & Acc. & & Acc. & F1 & Acc. &  & Acc. & Acc. &  & Acc. \\ \midrule
    \textbf{BLOOMZ 560m} \\ \midrule
    $k=0$ & 70.68 & 29.01 & 37.94 & 45.88 &  65.40 & 37.87 & 26.82 & 43.36 & 16.25  &  55.41 & 35.83 & 12.09 & 39.05 & 34.29 \\ 
    $k=1$  & & && && && && &&&\\ 
    $\quad$LaBSE & 80.20 & 62.79 & 60.19 &67.73 & 81.60    & 63.57  &   58.54  & 67.90
& 55.82 &  51.10 &53.46
&  33.61 & 60.82 & 55.68 \\
    $\quad$E5$_\text{LARGE}$ & 82.64 & 66.94 & 67.54& 72.37 & 82.00    &   66.41   &  67.54  & 71.98 & 57.94  & 50.56 &54.25 & 36.35 & 64.21 & 58.74\\ 
    $\quad$Cohere-Embedv3 & 83.40 & 66.44 & 69.43 & 73.09 & 82.05  &  65.02   &  67.70   &71.59& 58.12   &  52.18 & 55.15& 35.98 & 64.48 & 58.95\\ 
    \midrule
    \textbf{BLOOMZ 1.7B} \\ \midrule
    $k=0$ & 82.28 &        47.03 &   33.00 & 54.10 & 79.25   & 46.34   & 33.00  & 52.86 & 17.55  &  53.85 & 35.70 & 11.80 & 44.90 & 38.62 \\ 
    $k=1$  & & && && && && &&&\\ 
    $\quad$LaBSE & 84.60 & 54.27 & 62.00 & 66.96 & 81.75  &  55.81  &   60.50  &  66.02 & 57.19 & 56.75& 56.97 & 35.39 & 60.92 & 56.33 \\
    $\quad$E5$_\text{LARGE}$ & 86.48  &  58.14  &   69.51 &71.38 &  82.50   &  60.16  &   69.28   &70.65& \underline{59.05}   &    57.02 & \underline{58.04} & 38.50 & 64.52 & 59.64 \\ 
    $\quad$Cohere-Embedv3 & 86.48   & 58.80     & 71.31 & 72.20 &82.50  &   57.48  &   69.36   & 69.78 & \textbf{59.27}  &   57.07 &\textbf{58.17}& 37.69 & 64.44 & 59.46 \\
    \midrule
    \textbf{BLOOMZ 3B} \\ \midrule
    $k=0$ & 79.48         & 45.99  &  34.12 & 53.20 & 76.25   &      45.07   & 34.02 &51.78& 14.16 &  58.47 & 36.32 & 9.50 & 44.12 & 37.70 \\ 
    $k=1$ & & && && && && &&&\\
    $\quad$LaBSE & 85.68 & 64.98 & 62.37 &71.01 & 82.95  &  62.08     & 61.65 &  68.89  & 57.99   &    55.14 &56.57
& 37.46 & 63.37 & 58.48 \\
    $\quad$E5$_\text{LARGE}$ & \underline{86.52} & 66.68&  69.05 &74.08&  \underline{83.70}  &   65.69  &   70.17   &73.19&  59.41   &    55.46 &57.44& 39.09 & 66.20 & 60.95 \\ 
    $\quad$Cohere-Embedv3 & \textbf{86.88} & 66.80 &   70.59  &74.76& 82.40     & 61.05   &  69.72   &71.06 & 59.19   &    56.11 &57.65& 38.58 & 65.70 & 60.51 \\ \midrule
    \textbf{mT0 3B}\\\midrule
$k=0$ & 83.96 & 28.18 & 47.74 & 53.29 & 83.35 & 30.09 & 47.48 & 53.64 & 54.18 & 26.04 & 40.11 & 42.51 & 49.28 & 47.39\\
$k=1$ &  &  &  &  &  &  &  &  &  &  &  &  &  & \\
$\quad$E5$_\text{LARGE}$ & 85.12 & 39.34 & 52.60 & 59.02 & 81.55 & 36.87 & 55.16 & 57.86 & 54.17 & 39.16 & 46.67 & 42.36 & 54.04 & 51.48\\
\midrule 
\textbf{XGLM 564m}\\\midrule
$k=0$ & 60.80 & 32.11 & 24.84 & 39.25 & 55.50 & 31.24 & 24.83 & 37.19 & 11.91 & 47.93 & 29.92 & 10.46 & 33.29 & 29.21\\
$k=1$ &  &  &  &  &  &  &  &  &  &  &  &  &  & \\
$\quad$E5$_\text{LARGE}$ & 23.76 & 35.05 & 52.97 & 37.26 & 33.25 & 36.50 & 50.60 & 40.12 & 21.61 & 23.67 & 22.64 & 12.83 & 32.25 & 28.21\\
\midrule
\textbf{XGLM 2.9B}\\\midrule
$k=0$ & 63.84 & 38.84 & 24.56 & 42.41 & 58.20 & 37.76 & 24.53 & 40.16 & 11.97 & 57.45 & 34.71 & 10.39 & 36.39 & 31.92\\
$k=1$ &  &  &  &  &  &  &  &  &  &  &  &  &  & \\
$\quad$E5$_\text{LARGE}$ & 39.72 & 32.56 & 55.43 & 42.57 & 52.60 & 36.60 & 57.09 & 48.76 & 14.61 & 40.29 & 27.45 & 10.39 & 37.70 & 32.29\\
\midrule
\textbf{Aya-23 8B}\\\midrule
$k=0$ & 61.12 & 39.59 & 18.94 & 39.88 & 54.45 & 37.26 & 18.94 & 36.88 & 54.44 & 52.99 & 53.72 & 43.18 & 42.32 & 43.42\\
$k=1$ &  &  &  &  &  &  &  &  &  &  &  &  &  & \\
$\quad$LaBSE & 56.24 & 68.24 & 63.67 & 62.72 & 55.35 & 67.81 & 58.76 & 60.64 & 56.66 & 47.71 & 52.19 & 36.42 & 56.76 & 52.99\\
$\quad$E5$_\text{LARGE}$ & 54.52 & 67.57 & 68.90 & 63.66 & 56.15 & 67.54 & 66.89 & 63.53 & 58.85 & 47.39 & 53.12 & 38.50 & 58.48 & 54.70\\
$\quad$Cohere-Embedv3 & 54.16 & 67.66 & 69.61 & 63.81 & 54.05 & 68.51 & 64.72 & 62.43 & 58.77 & 46.53 & 52.65 & 37.17 & 57.91 & 54.01\\\midrule
\multicolumn{10}{l}{\textbf{Aya-101 13B}}\\\midrule
$k=0$ & 84.40      &    77.78   & 73.78  & 78.65 & 82.35     &    76.98 &   73.83     & 77.72 &  35.25  &  49.33 & 42.29 & 26.26 & 64.44 & 56.23 \\
$k=1$ &  &  &  &  &  &  &  &  &  &  &  &  &  & \\
$\quad$E5$_\text{LARGE}$ & 86.40     &     \underline{79.19} &    77.42 & \underline{81.00} &\textbf{85.80} &          79.24  &  75.56     & 80.20 &     48.59  &  53.20& 50.90 & 36.20 & 69.07 & 62.08 \\\midrule
\multicolumn{10}{l}{\textbf{Gemma 1.1 7B Instruct}}\\\midrule
$k=0$ & 71.20   &       52.68   & 42.64    & 55.51 & 67.05    &     50.21  &  42.82      &53.36 & 47.47  &  55.78 & 51.62 & 37.24 & 51.90 & 49.43 \\
$k=1$ &  &  &  &  &  &  &  &  &  &  &  &  &  & \\
$\quad$E5$_\text{LARGE}$ & 76.00      &    56.20 &     65.26  &  65.82  & 74.85     &    52.90   &  65.71  & 64.49   &  48.14   & 58.10 & 53.12 & 35.68  & 59.20 & 54.78 \\\midrule
\textbf{Llama 3 8B Instruct}\\\midrule
$k=0$ & 71.60 & 57.77 & 57.82 & 62.40 & 66.95 & 56.46 & 57.82 & 60.41 & 46.81 & 58.63 & 52.72 & 36.05 & 56.66 & 52.90\\
$k=1$ &  &  &  &  &  &  &  &  &  &  &  &  &  & \\
$\quad$LaBSE & 83.16 & 64.86 & 67.48 & 71.83 & 78.15 & 63.34 & 62.50 & 68.00 & 48.57 & 59.01 & 53.79 & 34.94 & 62.45 & 57.14\\
$\quad$E5$_\text{LARGE}$ & 85.04 & 66.59 & 72.92 & 74.85 & 77.65 & 64.34 & 66.83 & 69.61 & 49.82 & 58.42 & 54.12 & 35.68 & 64.14 & 58.57 \\
$\quad$Cohere-Embedv3 & 85.76 & 66.79 & 73.64 & 75.40 & 74.70 & 62.31 & 67.21 & 68.07 & 49.64 & 58.74 & 54.19 & 37.02 & 63.98 & 58.67\\\midrule
\textbf{Llama 3.1 8B Instruct}\\\midrule
$k=0$ &  74.88 & 49.85 &   57.04    &  60.59 & 70.85    &     48.66  &  57.07 & 58.86  & 37.45 &  58.53 & 47.99 & 26.56 & 53.43 & 48.50 \\
$k=1$ &  &  &  &  &  &  &  &  &  &  &  &  &  & \\
$\quad$E5$_\text{LARGE}$ & 86.36   &      58.70 &    72.99 & 72.68 & 78.45    &     32.49    & 66.05 & 59.00 &  49.37  & 58.85 & 54.11 & 35.16 & 59.82 & 55.24 \\\midrule
\textbf{Command-R}\\\midrule
$k=0$ & 65.16 & 35.27 & 43.50 & 47.98 & 59.25 & 35.42 & 43.39 & 46.02 & 50.72 & 58.96 & 54.84 & 44.44 & 48.46 & 48.32\\
$k=1$ &  &  &  &  &  &  &  &  &  &  &  &  &  & \\
$\quad$E5$_\text{LARGE}$ & 67.96 & 39.21 & 67.91 & 58.36 & 62.30 & 41.45 & 66.92 & 56.89 & 55.10 & 58.58 & 56.84 & 41.99 & 55.71 & 53.52\\\midrule
    \textbf{GPT-3.5 Turbo} \\ \midrule
    $k=0$ & 68.80 & 63.96    & 68.53   & 67.10 & 63.30      &    63.64 &   68.46     & 65.13 &  50.65  &  57.99  & 54.32 & 45.18 & 61.17 & 57.93 \\
    $k=1$ & & &&&&& &&&&&&\\ 
    $\quad$LaBSE & 77.12   & 62.53  &   71.43 &70.36& 75.25 &   65.65  & 72.03 &70.98 &  53.58    &   \textbf{60.95} & 57.27
& 42.14 & 64.52 & 60.19
\\
    $\quad$E5$_\text{LARGE}$ &  77.24 &   63.30    &  72.48 & 71.01 &75.25 &   65.97 & 73.47 &71.56& 53.84 &      \underline{60.41} & 57.13 & 42.73 & 64.97 & 60.61 \\ 
    $\quad$Cohere-Embedv3 & 77.16  &  63.07   &  72.23&70.82&74.20    & 66.52 & 73.27 &71.33& 52.90       & 60.14 & 56.52& 41.84 & 64.59 & 60.13 \\ \midrule
    \textbf{GPT-4o} \\ \midrule
    $k=0$ & 83.16  &  77.08    & \underline{79.53}& 79.92 &81.55     &    \underline{76.42}   & \underline{79.47}    & \underline{79.15} & 49.89  &  57.07 & 53.48 &\textbf{53.04} & \underline{70.80} & \underline{66.40} \\ 
    $k=1$  & & &&&&& &&&&&& \\
    $\quad$E5$_\text{LARGE}$ & 85.04      &  \textbf{79.52}  & \textbf{82.15} & \textbf{82.24} &83.20     &     \textbf{78.96}  &  \textbf{80.69}    & \textbf{80.95} & 57.25  &  57.02 & 57.14 &\underline{49.26} & \textbf{72.57} & \textbf{67.40} 
    \\ \bottomrule
    \end{tabular}
}
\caption{Results on ICL classification. \textbf{Bold} and \underline{underlined} numbers present the best and second-best models.} 
\label{results-icl-full}
\end{table*}

\begin{table*}[!ht]
\centering
\resizebox{\textwidth}{!}{
    \begin{tabular}{ll}
    \toprule
    \multicolumn{1}{l}{\textbf{Template}} & \multicolumn{1}{l}{Instruction:\texttt{<INSTRUCTION>}} \\
    \multicolumn{1}{l}{} & \multicolumn{1}{l}{Please only output the label.}\\ 
    \multicolumn{1}{l}{}& \multicolumn{1}{l}{\texttt{<FEW-SHOT SAMPLE>}}\\ 
    \\ 
    \multicolumn{1}{l}{}& \multicolumn{1}{l}{Options:\texttt{<OPTIONS>}}\\
    \multicolumn{1}{l}{}& \multicolumn{1}{l}{Input:\texttt{<QUERY>} Prediction:}
    \\ \midrule
    \multicolumn{1}{l}{\textbf{Few-shot sample}} & \multicolumn{1}{l}{Input:\texttt{<INPUT TEXT>}. Prediction:\texttt{<LABEL>}}
    \\ \midrule
    \textbf{Dataset} & \textbf{Prompt}
    \\ \midrule
    FIRE 2020 &  Instruction:Generate a sentiment label for a given input.\\
    & Please only output the label.\\
    & Input: Ikka     waiting......... Prediction:Positive\\ \\
    & Options:['Positive', 'Negative', 'Mixed', 'Unknown']\\
    & Input:mind blowing ikkaaaa.... Prediction: \\ \midrule
    LinCE SA & Instruction:Generate a sentiment label for a given input.\\
    & Please only output the label.\\ 
    & Input:@brissamayen Thanks :) ay si todavia le hablas a mi chikiya in the future te invitamos\\
    & a la boda ;) lol \u2665 Prediction:positive\\ \\ 
    & Options:['negative', 'neutral', 'positive']\\
    & Input:@brissamayen @sanluispotoyees estopp I blashhh lol jk but aww :) thanks haha ( x Prediction:\\ \midrule
    NollySenti & Instruction:Generate a sentiment label for a given input.\\
    & Please only output the label.\\
    & Input:Enjoy! Very nice... very nice indeed. Prediction:positive \\ \\
    & Options:['negative', 'neutral', 'positive']\\
    & Input:Damn....so interesting Prediction:  \\ \midrule
    NusaX & Instruction:Generate a sentiment label for a given input.\\
    & Please only output the label.\\
    & Input:Kawan ulun bagawi di gojek Prediction:neutral \\ \\
    & Options:['negative', 'neutral', 'positive']\\
    & Input:Macet di mana-mana pasl agi peraian Prediction: \\ \midrule
    SIB200 & Instruction:Generate a topic label for a given input.\\
    & Please only output the label.\\
    & Input:Batu kabidi bateka mikalu bua njila ya makasa ni ya makalu. Prediction:travel\\ \\ 
    & Options:['geography', 'science/technology', 'entertainment', 'politics', 'health', 'travel', 'sports']\\
    & Input:Anu kaniemesha uvua mutapika bibi ku mutu. Prediction:
    \\ \bottomrule
    \end{tabular}
}
\caption{Prompt examples. $k=1$ with LaBSE.} 
\label{icl-prompt-templates-labse}
\end{table*} 

\begin{table*}[!ht]
\centering
\resizebox{\textwidth}{!}{
    \begin{tabular}{ll}
    \toprule
    \multicolumn{1}{l}{\textbf{Template}} & \multicolumn{1}{l}{Instruction:\texttt{<INSTRUCTION>}} \\
    \multicolumn{1}{l}{} & \multicolumn{1}{l}{Please only output the label.}\\ 
    \multicolumn{1}{l}{}& \multicolumn{1}{l}{\texttt{<FEW-SHOT SAMPLE>}}\\ 
    \\ 
    \multicolumn{1}{l}{}& \multicolumn{1}{l}{Options:\texttt{<OPTIONS>}}\\
    \multicolumn{1}{l}{}& \multicolumn{1}{l}{Input:\texttt{<QUERY>} Prediction:}
    \\ \midrule
    \multicolumn{1}{l}{\textbf{Few-shot sample}} & \multicolumn{1}{l}{Input:\texttt{<INPUT TEXT>}. Prediction:\texttt{<LABEL>}}
    \\ \midrule
    \textbf{Dataset} & \textbf{Prompt}
    \\ \midrule
    FIRE 2020 &  Instruction:Generate a sentiment label for a given input.\\
    & Please only output the label.\\
    & Input: Njan mathram aano sunny chechiyee kaanan vannath  Sunny chechi uyir Prediction:Positive\\ \\
    & Options:['Positive', 'Negative', 'Mixed', 'Unknown']\\
    & Input:Sunny chechiye kaanan vannathu njan maathram aano Prediction: \\ \midrule
    LinCE SA & Instruction:Generate a sentiment label for a given input.\\
    & Please only output the label.\\ 
    & Input:hablar de los planes de spring break y mis 18 me pone bien hyper ! :D\\
    & Prediction:positive\\ \\ 
    & Options:['negative', 'neutral', 'positive']\\
    & Input: Prediction:\\ \midrule
    NollySenti & Instruction:Generate a sentiment label for a given input.\\
    & Please only output the label.\\
    & Input:Amazing Film. . . . Indeed the most anticipated film from Nollywood 2019 didn't disappoint. \\
    & Loved it all. Well done to Genevieve and Team. Prediction:positive \\ \\
    & Options:['negative', 'neutral', 'positive']\\
    & Input:This is the nollywood evolution. . . . This is arguably my best Nigeria movie for year 2019. \\
    & I cannot find any misplaced in this movie, perfectly executed, simple and so informative about our \\
    & society n thought provoking on career part for our children Prediction:  \\ \midrule
    NusaX & Instruction:Generate a sentiment label for a given input.\\
    & Please only output the label.\\
    & Input:Tempatnya nyaman banget, makanannya enak, kopinya enak. \\
    & Pas buat nongkrong bareng teman-teman atau makan malam. Prediction:positive \\ \\
    & Options:['negative', 'neutral', 'positive']\\
    & Input:Tempat yang bagus kalau dinikmati malam hari. Cukup nyaman. Harga cukup terjangkau. 
    \\
    & Favorit saya steak tenderloinnya. Cukup enak. Prediction: \\ \midrule
    SIB200 & Instruction:Generate a topic label for a given input.\\
    & Please only output the label.\\
    & Kel sirvisu ta uzadu txeu pa transporti, inkluindu artizanatu di lazer, y tanb\u00ea ispidisonz ki ten \\
    & nisisidadi di dadus y v\u00f3s a dist\u00e1nsia. Prediction:science/technology\\ \\ 
    & Options:['geography', 'science/technology', 'entertainment', 'politics', 'health', 'travel', 'sports']\\
    & Input:Sist\u00e9ma di IA gosi ta uzadu kuazi txeu na \u00e1rias di ikonumia, midisina, injinharia y \\
    & militar, sima ten stadu ta podu na txeu konputador di kaza y software di v\u00eddio geimi. Prediction:
    \\ \bottomrule
    \end{tabular}
}
\caption{Prompt examples. $k=1$ with E5$_\text{LARGE}$.} 
\label{icl-prompt-templates-e5}
\end{table*}

\end{document}